\documentclass{article} 
\usepackage[dvipsnames]{xcolor} 

\definecolor{mydarkred}{rgb}{0.6,0,0}
\definecolor{myblue}{HTML}{268BD2}
\definecolor{mygreen}{HTML}{658354}
\definecolor{orangeinplot}{HTML}{e29c7a}
\definecolor{purpleinplot}{HTML}{7676a4}
\definecolor{greeninplot}{HTML}{288308}

\usepackage{iclr2025_conference,times}
\usepackage{amsfonts}
\usepackage{amsmath}
\usepackage{bm}
\usepackage{natbib}
\usepackage{algorithm}
\usepackage{algpseudocode}
\usepackage{verbatim}
\usepackage{float}
\usepackage{multirow}
\usepackage{xspace}
\usepackage{pifont}
\usepackage{wrapfig}
\usepackage{tikz}
\usepackage{ctable}
\usepackage{graphicx}
\usepackage{caption}
\usepackage{subcaption}
\usepackage{amsthm}
\usepackage[most]{tcolorbox}
\usepackage{url}

\usepackage[colorlinks,
linkcolor=mydarkred,
urlcolor=RoyalBlue,
citecolor=myblue]{hyperref}
\usepackage{cleveref}

\definecolor{propbg}{HTML}{F2F2F9}
\definecolor{propfr}{HTML}{00007B}
\newtcolorbox{proposition}{
  enhanced,
  boxrule=0pt,frame hidden,
  borderline west = {2pt}{0pt}{propfr},
  colback=propbg,
  sharp corners
}

\definecolor{darkred}{rgb}{0.9, 0, 0}
\newcommand{\redcircle}[1]{%
  \tikz[baseline=(char.base)]\node[shape=circle,fill=darkred,text=white,inner sep=0.05pt] (char) {\small #1};%
}

\title{GOLD: Graph Out-of-Distribution Detection via Implicit Adversarial Latent Generation}

\author{Danny Wang, Ruihong Qiu, Guangdong Bai, Zi Huang\\
The University of Queensland\\
\texttt{\{danny.wang,r.qiu,g.bai,helen.huang\}@uq.edu.au}
}

\iclrfinalcopy 
\begin{document}

\maketitle

\begin{abstract}
Despite graph neural networks' (GNNs) great success in modelling graph-structured data, out-of-distribution (OOD) test instances still pose a great challenge for current GNNs. One of the most effective techniques to detect OOD nodes is to expose the detector model with an additional OOD node-set, yet the extra OOD instances are often difficult to obtain in practice. Recent methods for image data address this problem using OOD data synthesis, typically relying on pre-trained generative models like Stable Diffusion. However, these approaches require vast amounts of additional data, as well as one-for-all pre-trained generative models, which are not available for graph data.
Therefore, we propose the GOLD framework for graph OOD detection, an implicit adversarial learning pipeline with synthetic OOD exposure without pre-trained models.
The implicit adversarial training process employs a novel alternating optimisation framework by training: (1) a latent generative model to regularly imitate the in-distribution (ID) embeddings from an evolving GNN, and (2) a GNN encoder and an OOD detector to accurately classify ID data while increasing the energy divergence between the ID embeddings and the generative model's synthetic embeddings. This novel approach implicitly transforms the synthetic embeddings into pseudo-OOD instances relative to the ID data, effectively simulating exposure to OOD scenarios without auxiliary data.
Extensive OOD detection experiments are conducted on five benchmark graph datasets, verifying the superior performance of GOLD without using real OOD data compared with the state-of-the-art OOD exposure and non-exposure baselines.
\footnote{Code is available at \url{https://github.com/DannyW618/GOLD}.}
\end{abstract}

\section{Introduction}
The proliferation of Graph Neural Networks (GNNs) across diverse domains and real-world applications has underscored the importance of robust and reliable predictive systems~\citep {GCN,GraphSage}. 
Their performance relies crucially on the assumption that the testing data follows the same distribution as the training data~\citep{OOD-GNN, GCN, LiSA, GraphSage}. This assumption is frequently violated in practice, as real-world graph data is generally filled with out-of-distribution (OOD) instances~\citep{GOOD-cert, CIGA, GDSBenchmark, OODLink, GOODSurvey, MoleOOD}. Consequently, inaccurate predictions will inevitably be made by the deployed models, which can be detrimental in critical areas like medical diagnosis and drug discovery~\citep{medicalDiagnoisis,GraphMedicalDiagnosis, HealthOOD, HealthForecast, DrugOOD}. Thus, it is necessary to develop OOD detection methods to identify out-of-distribution instances that deviate from the training distribution~\citep{OODD, LRatio_OODD, NLPOODD, GOODD-uncertainty}.

Recent work has made significant strides in developing OOD detection techniques tailored for graph-structured data, primarily in three categories~\citep{OODGAT, LMN, GNNSafe, GPN}. (1) General OOD detection methods train the detector only with in-distribution (ID) data from the training set~\citep{SGOOD, grasp, GOOD-D, GOODAT}. This process involves fine-tuning a classifier and learning graph representations to improve the model's OOD detection performance using various scoring metrics. (2) A more effective method for OOD detection is OOD exposure, which takes advantage of exposing the detector with additional OOD samples during training~\citep{SLW, GNNSafe, OE, GenOE}. These methods generally require an extra dataset containing OOD samples and the detector is trained to discriminate the ID training data with these OOD data. (3) More recently, OOD synthesis methods have been proposed for image data, mainly leveraging pre-trained generative models, e.g., Stable Diffusion~\citep{sd}, to create OOD samples that lie on the boundary of ID data~\citep{NPOS, Dream-OOD, DFDD, ATOL}.


Despite the effectiveness of OOD exposure-based methods over general OOD detection methods, two challenges remain: (1) For the OOD exposure approaches using a real and additional OOD dataset, acquiring these extra OOD samples is often infeasible during model training in the real world. Furthermore, relying on the additional OOD dataset to guide the detector in distinguishing the ID and OOD data could lead to an inaccurate decision boundary. This is because the training logic assumes that the exposed OOD data can represent the distribution of OOD data from test scenarios, which has no guarantee in real-world~\citep{VOS, manifold}. (2) Although OOD synthesis-based approaches have been proposed to resolve the lack of unknown data, these methods typically rely on pre-trained models built upon substantial amounts of auxiliary data~\citep{regOOD, NPOS, Dream-OOD, MOL_DIF}. 
Moreover, the lack of a one-for-all pre-trained generative model for graph data hinders the synthesis of OOD data using simple plug-and-play models~\citep{GFM}. Thus, this presents the key motivation:
\begin{center}
    \textbf{\textit{How to enhance graph OOD detection by exposing to OOD scenarios without auxiliary data?}}
\end{center}

\begin{wrapfigure}{R}{0.5\textwidth}
    \vspace{-0.5cm}
    \centering
    \begin{subfigure}{0.24\textwidth}
        \centering
        \includegraphics[width=1\textwidth]{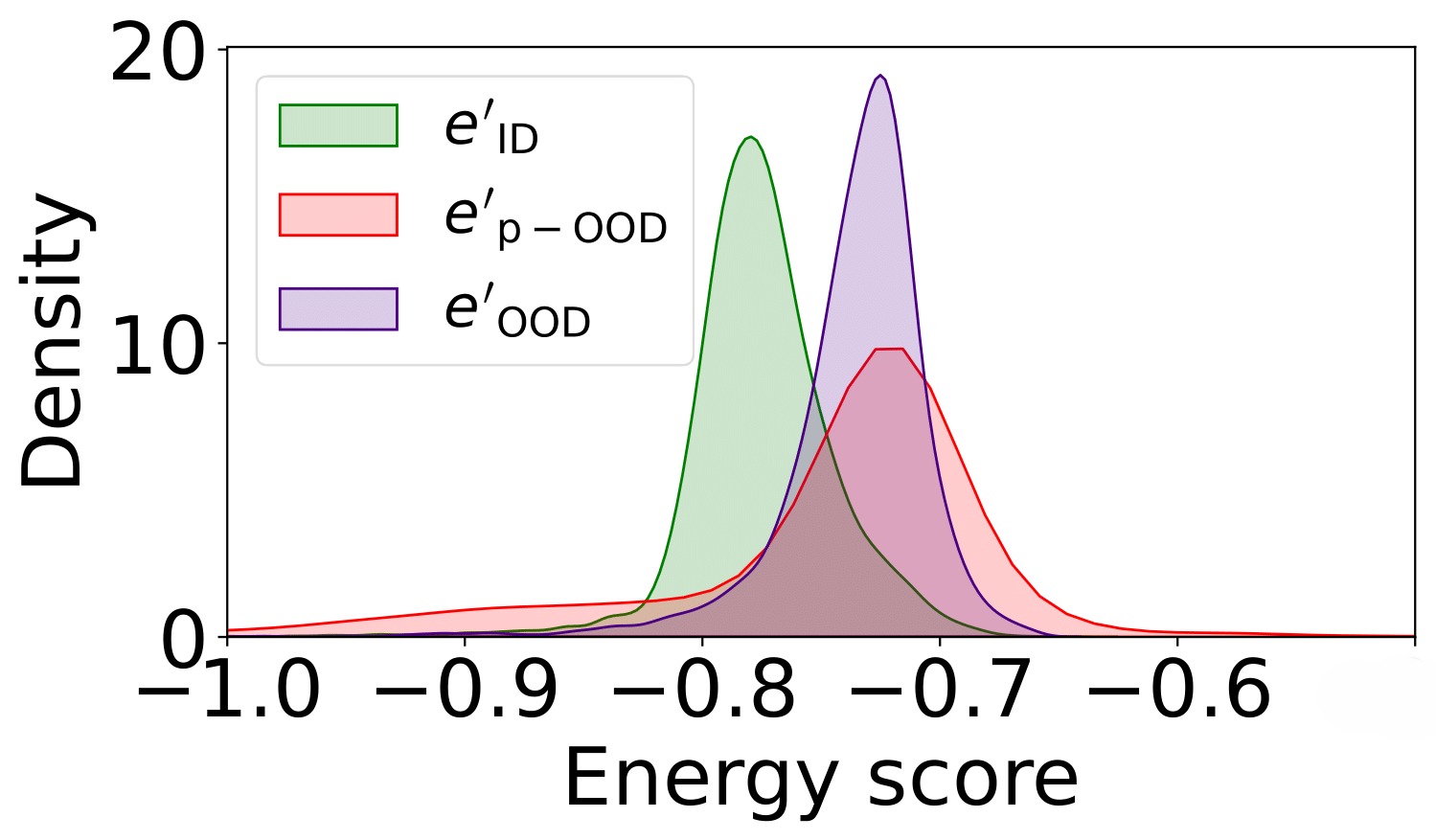}
        \vspace{-0.6cm}
        \caption*{(a) Initial energy.}
    \end{subfigure}
    \begin{subfigure}{0.24\textwidth}
        \centering
        \includegraphics[width=\linewidth]{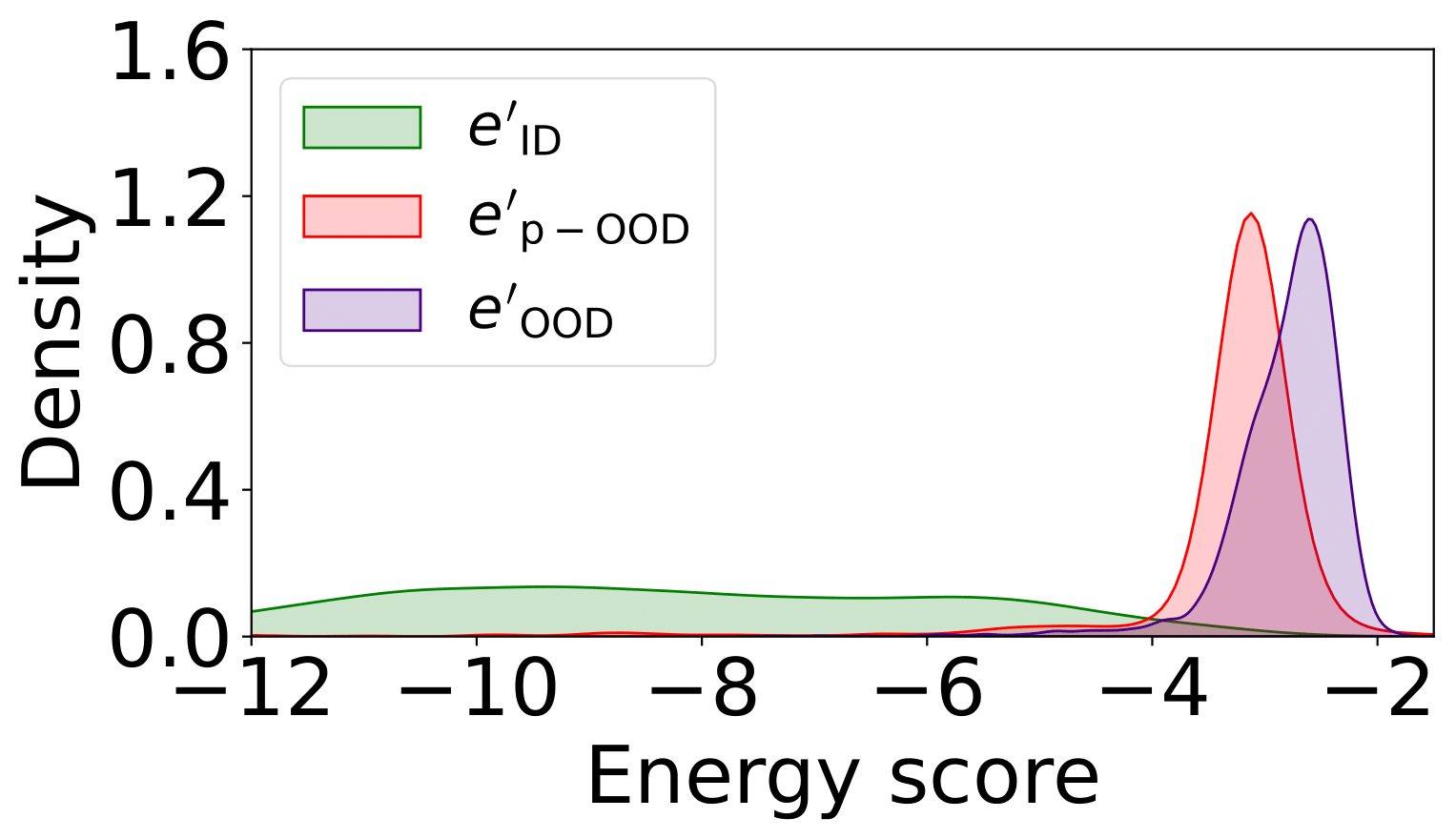}
         \vspace{-0.6cm}
        \caption*{(b) Post-training energy.}
    \end{subfigure}
    
    \begin{subfigure}{0.25\textwidth}
        \centering
        \includegraphics[width=\linewidth]{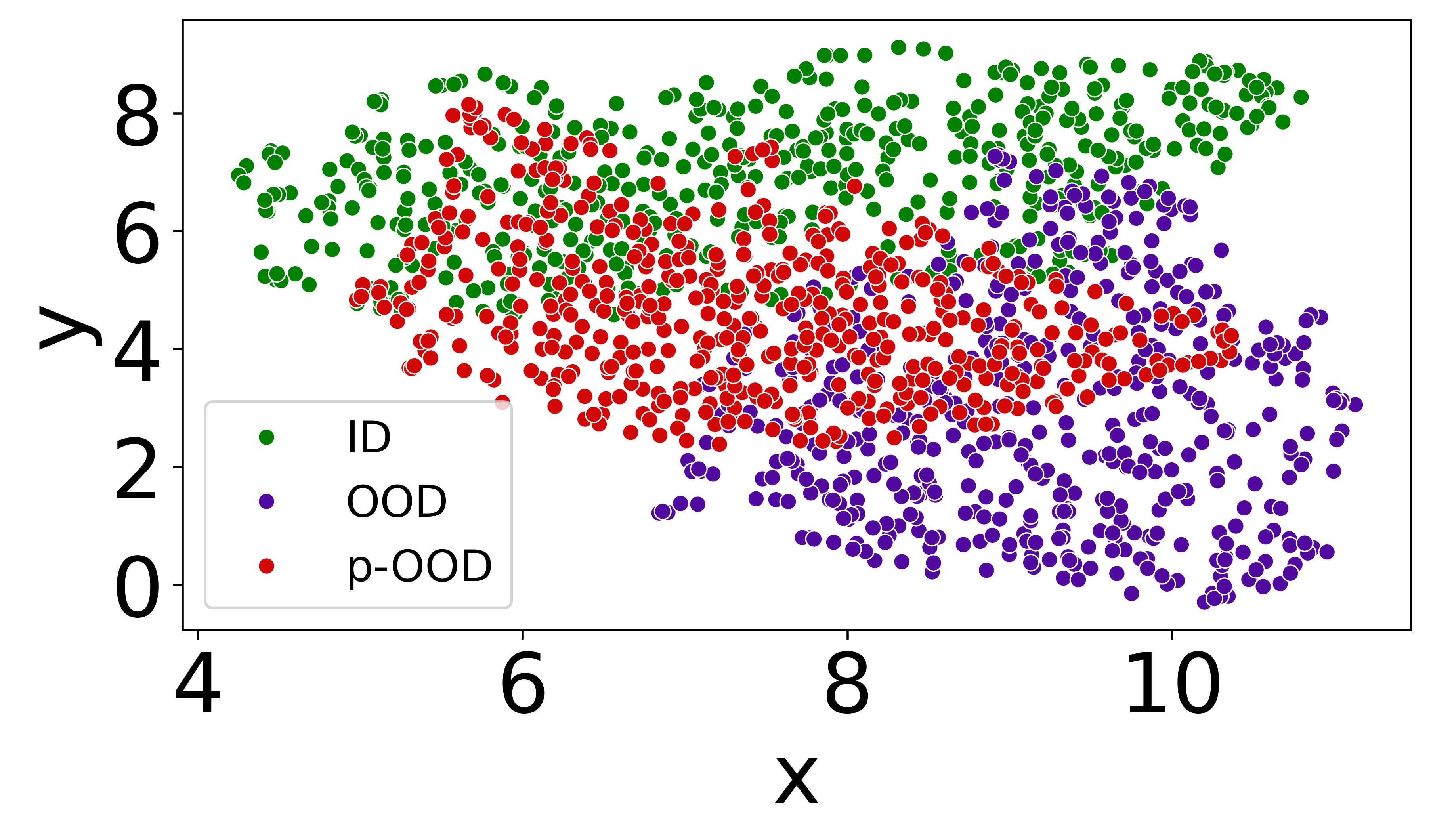}
         \vspace{-0.6cm}
        \caption*{(c) Initial embeds.}
    \end{subfigure}%
    \begin{subfigure}{0.26\textwidth}
        \centering
        \includegraphics[width=\linewidth]{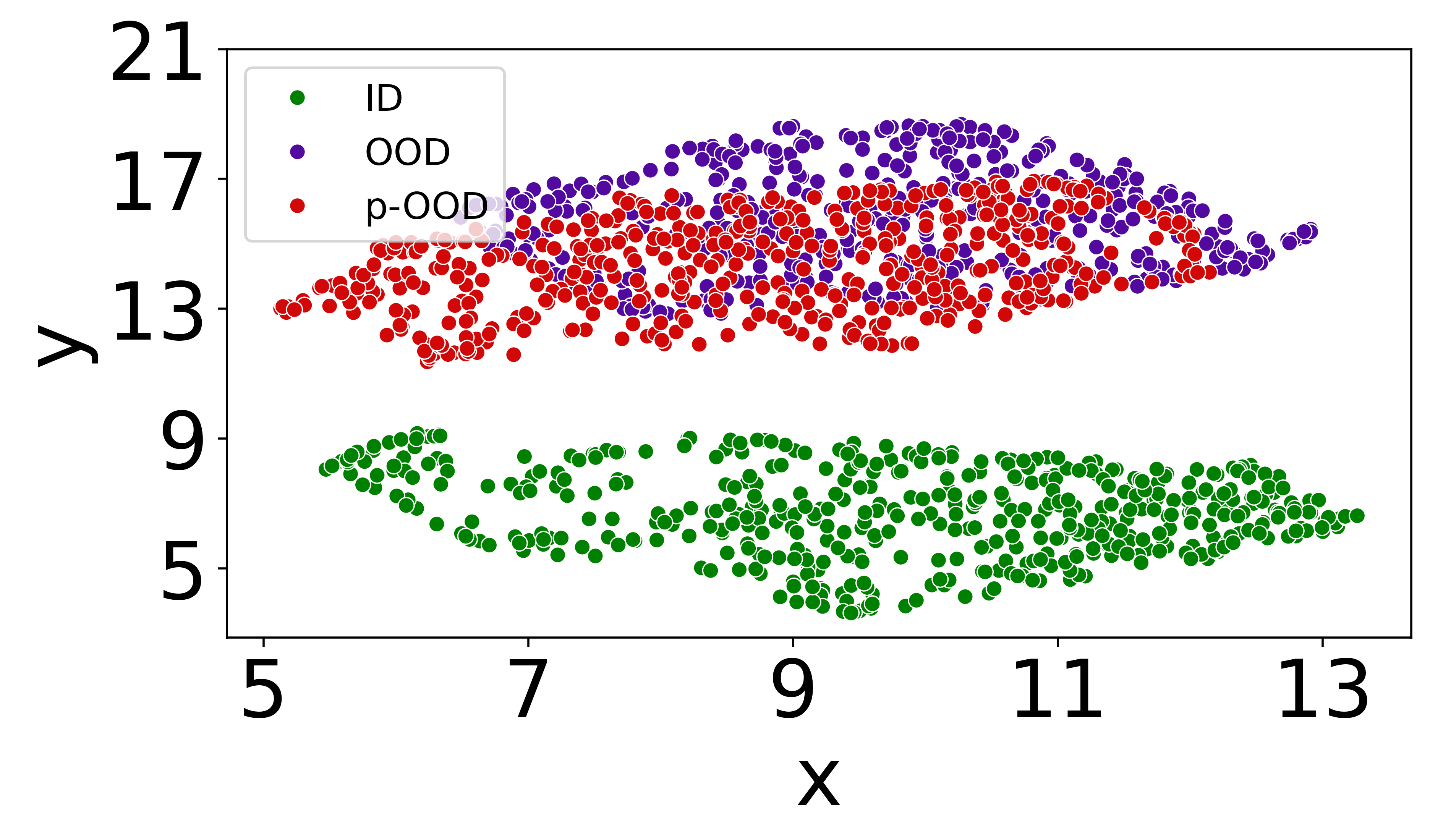}
         \vspace{-0.6cm}
        \caption*{(d) Post-training embeds.}
    \end{subfigure}%
\caption{Motivation of GOLD: The initially close energy distributions (a) after training the latent generative model, become separated  after training GOLD (b), where the initial pseudo-OOD (p-OOD) embeddings (embeds.,) (c) implicitly diverges from the ID data and resembles real OOD instances (d).}
\vspace{-0.5cm}
\label{fig:GOLD Motivation}
\end{wrapfigure}

In light of the above challenges, the intuition of this work is to generate and expose pseudo-OOD samples solely based on the ID training data to ensure effective OOD detection. To achieve this, we propose an implicit adversarial training framework with a novel alternating optimisation schema by training: (1) a latent generative model (LGM) to \textbf{regularly generate embeddings similar to the in-distribution (ID) embeddings from an evolving GNN}, and (2) a GNN encoder and an OOD detector to accurately classify ID data while \textbf{increasing the energy divergence between these generated embeddings and the ID embeddings}. This novel approach implicitly transforms synthetic embeddings into pseudo-OOD instances relative to the ID data, effectively simulating OOD exposure without auxiliary data. Evident in Figure \ref{fig:GOLD Motivation}, the initially similar energy distributions after LGM training diverge post-training, which implicitly separates the embedding distributions, ensuring the pseudo-OOD data resemble close to the real OOD instances. The main contributions of this paper are summarised as follows:
\begin{itemize}
  \item We propose GOLD, a novel non-OOD exposed synthesis-based framework for graph OOD detection. GOLD includes a unique implicit adversarial training paradigm for effective pseudo-OOD synthesis, which is achieved by a latent generative model and a novel detector.
  \item We conducted extensive experiments on five benchmark datasets. Without auxiliary OOD data, GOLD achieves state-of-the-art performance compared with non-OOD and OOD exposure methods, with the best improvement of FPR95 reduced from 33.57\% to 1.78\%.
\end{itemize}




\section{Preliminary}
Generally, for a node classification problem, a graph is denoted as $ \mathcal{G} = (\mathbf{X},\mathbf{A})$, where $\mathbf{X}\in\mathbb{R}^{n\times d}$ is the node feature matrix with $n$ nodes and feature dimension $d$, and $\mathbf{A}\in\mathbb{R}^{n\times n}$ is an adjacency matrix indicating the connection among nodes. Each node is associated with a label $y \in \{1, 2, ..., C\}$ indicating a total of $C$ classes. For out-of-distribution detection, there are generally two main tasks:

\paragraph{Task I: In-distribution classification.} To formulate the node classification problem for in-distribution data, given test nodes from the same distribution as training nodes, $ P_{train}(\mathbf{X, A}) = P_{test}(\mathbf{X, A})$ and the conditional distribution $P_{train}(\mathbf{y}|\mathbf{X, A}) = P_{test}(\mathbf{y}|\mathbf{X, A})$, the task is to develop an $L$-layer GNN classifier to predict the label $\mathbf{y}\in\mathbb{R}^n$ for the testing nodes with trainable parameters in the GNN classifier (see Appendix~\ref{Appendix:GNN} for details of GNN):
\begin{equation}
    \mathbf{y}=\text{Softmax}(\text{GNN}(\mathbf{X},\mathbf{A})).
\end{equation}
\paragraph{Task II: Out-of-distribution detection.} To detect the testing nodes coming from a different distribution from the training data, where $ P_{train}(\mathbf{X, A}) \neq P_{test}(\mathbf{X, A})$ and the conditional distribution $P_{train}(\mathbf{y}|\mathbf{X, A}) \neq P_{test}(\mathbf{y}|\mathbf{X, A})$, the task is to require an OOD detector $F$ to output a binary prediction for the testing nodes. $F$ is usually built upon the output from the classifier GNN with $F(\mathbf{x},\mathbf{A};\text{GNN})=1$ for data from in-distribution and $F(\mathbf{x},\mathbf{A};\text{GNN})=0$ for data from out-of-distribution~\citep{GNNSafe, energy, FS-OOD}.

\paragraph{Energy Score-based Detector.} Recent work indicated that using the energy score from logits in the classifier can benefit OOD detection~\citep{energy, GNNSafe, EBM-NN}. The energy score $e$ for a node $i$ is defined as:
\begin{equation}
    e_i = - \log \sum\nolimits_{c=0}^{C-1}{\exp(\mathbf{z}_{i,c})},
    \label{eq: energy_gcn}
\end{equation}
where $e_i\in\mathbb{R}$ is the energy score of node $i$, $\mathbf{z}_i\in\mathbb{R}^C$ is the logits for node $i$ output from the classifier $\mathbf{Z}=\text{GNN}(\mathbf{X},\mathbf{A})\in\mathbb{R}^{n\times C}$, and $c$ is to select the logit of the $c$-th element of $\mathbf{z}_i$. Therefore, the energy score-based OOD detector for a node $i$ is instantiated with a threshold $\tau$ as:
\begin{equation}
    F\left(\mathbf{x}_i, \mathbf{A}; \text{GNN}\right)= \begin{cases}0, & \text { if } \quad e_i \ge \tau, \\ 1, & \text { if } \quad e_i < \tau.\end{cases}
    \label{eq:ood_criteria}
\end{equation}
The training of this energy score-based OOD detector is generally based on an energy regulariser~\citep{energy}, which maximises the difference between the energy scores from in-distribution data ($P_\text{ID}$) and out-of-distribution data ($P_\text{OOD}$) with two scalar thresholds, $t_\text{ID}$ and $t_\text{OOD}$:
\begin{equation}
    \begin{split}
    \max_\text{GNN}\mathcal{L}_\text{EReg},    \text{ where }\mathcal{L}_\text{EReg} = 
    \mathbb{E}_{i\sim P_\text{ID}}\left[\operatorname{max}\left(0, t_{\text {ID}}-e_i\right)\right]^2
    + \mathbb{E}_{j\sim P_\text{OOD}}\left[\operatorname{max}\left(0, e_j-t_{\text{OOD}}\right)\right]^2.
    \end{split} 
    \label{eq: ereg}
\end{equation}
\paragraph{Energy Propagation for OOD Detector.} To facilitate the energy score for graph data, \textsc{GNNSafe}~\citep{GNNSafe} proposes an energy propagation schema that emulates label propagation for effective OOD detection. This propagated energy is then fed into the objective in Eq.~\ref{eq: ereg}:
\begin{equation}
\mathbf{e}^{(k)}=\alpha \mathbf{e}^{(k-1)}+(1-\alpha) \mathbf{D}^{-1} \mathbf{A} \mathbf{e}^{(k-1)},
\label{eq:eprop}
\end{equation}

where $\mathbf{e}^{(k)}\in\mathbb{R}^{n\times 1}$ is the energy scores for $n$ nodes after $k$-th energy propagation with $\alpha \in [0,1]$ controlling the concentration of energy. $\mathbf{D}$ is the degree matrix of graph $\mathcal{G}$. In the following, the energy scores in our framework will be the propagated energy scores and will be used interchangeably.

\section{GOLD} \label{sec:GOLD_Method}
\begin{figure}[!t]
\includegraphics[width=0.8\linewidth]{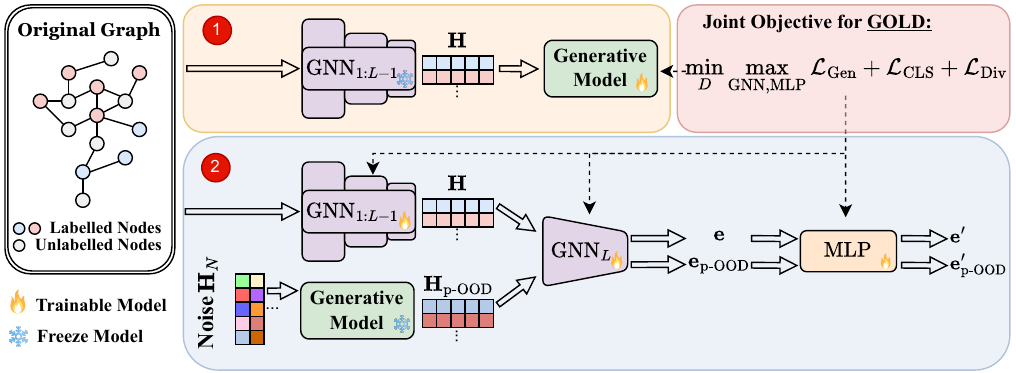}
\centering
\caption{Overview of GOLD. Given an input graph, GOLD consists of two components: \textbf{Step} \protect\redcircle{1} trains a latent generative model using hidden representation $\mathbf{H}$ from a frozen GNN.
\textbf{Step} \protect\redcircle{2} trains a GNN classifier and an OOD detector based on the ID data $\mathbf{H}$ and the latent generator generated pseudo data $\mathbf{H}_\text{p-OOD}$. The overall training is in an adversarial manner.
}
\vspace{-0.3cm}
\label{fig:pipline}
\end{figure}

In this section, the GOLD framework for graph OOD detection is described with illustration in Figure~\ref{fig:pipline}. 
In summary, GOLD is trained with a novel implicit adversarial objective that optimises a latent generative model (LGM), a GNN classifier, and an OOD detector. The LGM aims to generate embeddings akin to ID data, while the implicit adversarial objective encourages divergence between the ID and OOD energy scores derived from GNN and detector. This process implicitly transforms the synthetic embeddings into pseudo-OOD, effectively facilitating synthetic OOD exposure. 

The GNN classifier is trained to maximise the log probability of the ground truth classification label:
\begin{equation}\max_\text{GNN}\mathcal{L}_\text{CLS},    \text{ where }\mathcal{L}_\text{CLS} = \log p(y| \mathbf{x}, \mathcal{G}_{\mathbf{x}}).
\label{eq:cls}
\end{equation}
In the following, the detector and the latent generator are both built upon this GNN in GOLD.




\subsection{Latent Generative Model as Pseudo-OOD Generator} \label{sec:latent diffusion model}

In light of the key motivation of exposing the model to OOD scenarios with generated data, an LGM is employed for pseudo-OOD synthesis. The model would take input from the encoded node representations $\mathbf{H} \in \mathbb{R}^{n\times d'}$, where $d'$ is the hidden dimension, of the GNN module after the $(L-1)$-th layer, which captures both the global and local information~\citep{GCN}:
\begin{equation}
    \mathbf{H}=\text{GNN}_{1:L-1}(\mathbf{X},\mathbf{A}),\quad\mathbf{Z}=\text{GNN}_{L}(\mathbf{H},\mathbf{A})=\text{GNN}(\mathbf{X},\mathbf{A}). \label{eq:GCN_emb_logits}
\end{equation}
The LGM aims to mimic and generate latent embeddings close to the ID representations. This is typically achieved by minimising a reconstruction loss or distance between a target and predicted embedding, i.e., if a latent diffusion model (LDM)~\citep{DDPM} or a variational autoencoder (VAE)~\citep{VAE} is used as the pseudo-OOD generator, the objective is given by:

\vspace{-0.5cm}
\begin{equation}
\min_D\mathcal{L}_{\mathrm{Gen}},
\label{eq:latent}
\end{equation}
\begin{equation}
\notag
\text{ where }\mathcal{L}_{\mathrm{Gen}} = 
\begin{cases}
\mathbb{E}_{\mathbf{h}_0, \bm{\epsilon}, t}\left[\left\|\bm{\epsilon} -D\left(\sqrt{\bar{a}_t} \mathbf{h}_0 +\sqrt{\left(1-\bar{a}_t\right)} \bm{\epsilon}, t\right)\right\|_2^2\right],& \text{if LDM}\\
 \mathbb{E}_{q_{D}(\mathbf{h}_\text{p-OOD}|\mathbf{h})} \left[ \log p(\mathbf{h} |\mathbf{h}_\text{p-OOD}) \right] - \text{KL} \left[ {q_{D}}(\mathbf{\mathbf{h}_\text{p-OOD}} | \mathbf{h}) \| p(\mathbf{h}_\text{p-OOD}) \right] ,& \text{if VAE}
\end{cases}
\end{equation}
with latent vectors $\mathbf{h}$ and decoder $D$. For LDM, $D$ is a denoising network that predicts and progressively removes noise during the backward denoising step. Contrary, VAE minimises a reconstruction and embedding distance loss, based on the input latent embeddings and the embeddings generated by the decoder $D$. The pseudo-OOD latent embeddings $\mathbf{h}_\text{p-OOD}$ can thus be generated using the decoder network $D$ with noise vector sampled from a normal distribution $\mathbf{h}_N\sim\mathcal{N}(0,\mathbf{I})$ via:
\begin{equation}
\label{eq:p-ood-dist}
    \mathbf{h}_\text{p-OOD}\sim P_\text{p-OOD},\text{ and }P_\text{p-OOD}=P_D(\mathbf{h}_{\text{p-OOD}}|\mathbf{h}_N)
\end{equation}


For comparison, VAE presents competitive performance and faster training time due to model design, while LDM remains efficient and performs better among (non-) OOD exposure methods when used in GOLD. Moreover, GOLD achieves the same inference time as SOTA baselines with any LGMs. This is because the generative model is not involved during inference. A detailed description of the two generative methods and their corresponding objective is provided in Appendix \ref{Appendix:latent_generative_model}, and further results about effectiveness and efficiency will be discussed in Section \ref{sec:overall_performance} and \ref{sec:computational_cost}. 

Note that at this stage, the synthetic embeddings generated by the LGM still imitate the ID data. In the following subsections, with a novel detector, an implicit adversarial training process will be introduced, which separates the synthetic embeddings from the ID representations, transforming it into pseudo-OOD instances.

\subsection{OOD Detector for ID and Pseuro-OOD Separation}\label{sec:ood detector}
Given that a trained latent generator can synthesise latent representations akin to the ID embeddings, the OOD detector is designed to pull apart the energy scores of ID instances from those of the generated data. This ensures a clear separation between the distributions, and through gradient flow to the trainable GNN encoder, it implicitly separates the synthetic embeddings from the ID embeddings. Hence, the synthetic data is effectively transformed into pseudo-OOD instances relative to ID data, as shown in Figure~\ref{fig:GOLD Motivation}, allowing the model to be exposed to OOD scenarios without the need for real OOD data.
For clarity, in the following, the OOD exposure in previous methods will be replaced with the pseudo-OOD data generated by the LGM from Eq.~\ref{eq:p-ood-dist}.

In the general design of an energy-based OOD detector as in Eq.~\ref{eq: energy_gcn}, the energy score is a combination of the prediction logits. To overcome the potential difficulties when the number of classes increases or when certain classes are unable to be accurately distinguished by the model, an MLP is applied to the energy and trained with an uncertainty loss as in~\cite{VOS} with $\phi$ as the softmax function:
\begin{equation}
\label{eq:uncertainty}
    \max_\text{GNN, MLP}\mathcal{L}_\text{Unc},\text{ where }\mathcal{L}_\text{Unc}=\mathbb{E}_{i\sim P_\text{ID}} \log[\phi(\text{MLP}(e_i))_{{[0]}}] + \mathbb{E}_{j\sim P_\text{p-OOD}}\log[\phi(\text{MLP}(e_j))_{{[1]}}].
\end{equation}
The subscripts indicate the label of the corresponding logit value from the MLP model after applying $\phi$ (i.e., [0] represents the ID Class 0 and [1] represents the OOD Class 1).
In addition to using this uncertainty objective, we aim to further transform the energy with the classifier output to enhance the separability of the energy:
\begin{equation}
    e'_i = - \log [ e^{\text{MLP}(e_i)_{{[0]}}} + e^{\text{MLP}(e_i)_{{[1]}}}]. \label{eq:denergy}
\end{equation}
With the transformed energy $\mathbf{e}'$, we propose a new divergence regularisation to obtain a more diverged energy score distribution than the pre-transformed energy $e$ from the GNN classifier:
\begin{equation}
    \max_\text{GNN, MLP}\mathcal{L}_\text{DReg},\text{ where }\mathcal{L}_\text{DReg} =  \mathbb{E}_{i\sim P_\text{ID}}\operatorname{max}\left( 0, e_i - e'_i\right)^2+\mathbb{E}_{j\sim P_\text{p-OOD}}\operatorname{max}\left(0, e'_j -e_j\right)^2. \label{eq:dreg}
\end{equation}
We next show that the combination of the two proposed losses with pseudo-OOD data could enable the detector to produce more distinctive energy scores between distributions to assist OOD detection.
\vspace{-0.4cm}
\begin{proposition}
    \vspace{-0.2cm}
    \textbf{Proposition 1.} \textit{The gradient descent on $\mathcal{L}_\text{Unc}$ and $\mathcal{L}_\text{DReg}$ will overall decrease (increase) the transformed energy $\mathbf{e}'$ for in-distribution (pseudo-out-of-distribution) instance, bounded by the given initial energy $\mathbf{e}\sim P_\text{ID}$ ($P_\text{p-OOD}$), respectively, for the detector MLP model.}
    \vspace{-0.2cm}
\end{proposition}


The proof is provided in Appendix~\ref{Appendix:Proof}. Intuitively, the $\mathcal{L}_\text{Unc}$ aims to train the detector to classify the ID and OOD data with high probability under binary classification, which ensures the separability of embeddings. While the $\mathcal{L}_\text{DReg}$ aims to diverge the energy of ID and OOD based on the logits from this same detector. Therefore, the logits of ID data are expected to have a larger scale than the logits of OOD data, which leads to the energy score based on the logits for this binary classification providing a greater discrepancy between ID and OOD data. The empirical visualisation is shown in Figure~\ref{fig:Logits_smf} of Appendix~\ref{Appendix:Logits vs. SFM}.

\textbf{\textit{Energy Divergence Objective}}: Replacing $P_\text{OOD}$ with $P_\text{p-OOD}$, additionally with the $\mathcal{L}_\text{EReg}$ from Eq.~\ref{eq: ereg}, the objective to diverge the energy for the OOD detector is a combination with weight $\mu, \lambda, \gamma \in \mathbb{R}$:
\begin{equation}
\label{eq:div}
    \max_\text{GNN, MLP}\mathcal{L}_\text{Div},\text{ where }\mathcal{L}_\text{Div}=\mu\mathcal{L}_\text{EReg}+\lambda\mathcal{L}_\text{Unc}+\gamma\mathcal{L}_\text{DReg}.
\end{equation}
After optimising the MLP detector and GNN classifier with the final energy divergence objective $\mathcal{L}_\text{Div}$, the embeddings generated by the fixed LGM will implicitly diverge from the ID embeddings produced by the updated classifier. This divergence occurs because the energy scores generated by the detector are separated by the optimised objective, which will further train the GNN classifier via gradient flow. As a result, the LGM would effectively function as a pseudo-OOD generator.

\subsection{Implicit Adversarial Objective}\label{sec:adversarial paradigm}
To accomplish the ID classification and the OOD detection, the overall objective of the pseudo-OOD synthesis and OOD detector can be formulated in an adversarial style, by combining Eq.~\ref{eq:latent},~\ref{eq:cls} and~\ref{eq:div}:
\begin{equation}
\min_{D}\max_{\text{GNN}, \text{MLP}}\quad \mathcal{L}_\text{Gen}+\mathcal{L}_\text{CLS}+\mathcal{L}_\text{Div}\label{eq:jointObj}
\end{equation}
The intuition of this adversarial objective stems from the contradictory optimisation purpose from the individual objectives. When fixing the GNN encoder, $\mathcal{L}_\text{Gen}$ aims to optimise the LGM to minimise the gap between the generated pseudo-OOD embeddings and ID embeddings. This ensures the LGM can generate meaningful representations that are initially close to ID data, instead of generating meaningless and far away pseudo-OOD data. When fixing the generator, $\mathcal{L}_\text{CLS}$ and $\mathcal{L}_\text{Div}$ aims to optimise the GNN encoder and the MLP detector to maximise the discrepancy in the energy score between the pseudo-OOD data and the ID data, while keeping the GNN encoding for ID data meaningful for classification. Notably, GOLD does not directly generate pseudo-embeddings from the LGM against the ID embeddings but instead, the encoder and the detector implicitly pull the ID embeddings away from the generated pseudo-OOD embeddings via the energy score divergence.


We note that adversarial training schemas were also previously developed by \cite{ConfOOD} and \cite{BadGAN} to generate OOD data or outliers without pre-training. These methods present a GAN-based model with generator and discriminator components, alongside additional density estimation methods, such as a confidence classifier and pre-trained models for OOD prediction. These methods are shown to be not comparable to the OOD exposure-based detector~\citep{Dream-OOD}. In contrast, the key adversarial aspect of our framework is that the LGM generates embeddings resembling ID data, while the detector and classifier amplify the energy score gap between these generated pseudo-OOD embeddings and ID samples, implicitly affecting the embeddings' distribution.

\subsection{Alternating Optimisation}
\begin{wrapfigure}{R}{0.55\textwidth}
\begin{minipage}{0.55\textwidth}
\vspace{-1.7cm}
\begin{algorithm}[H] 
\caption{Adversarial Optimisation of GOLD}\label{alg:GOLD}
\begin{algorithmic}[1]
\Require {ID graph $\mathcal{G} = (\mathbf{A}, \mathbf{X})$, randomly initialised GNN, MLP detector, and latent generator $D$, epoch numbers $M_1$ and $M_2$, loss coefficients $\lambda, \mu, \gamma$.}
\Ensure Optimised GNN, MLP detector, and generative model $D$
\While{$train$}
\State Obtain $\mathbf{H}$ for ID data with GNN from Eq.~\ref{eq:GCN_emb_logits}
\For{$epoch = 1, \dots, M_1$} \hfill{\textcolor{blue}{// \textbf{Step \protect\redcircle{1}}}}
    \State Train $D$ with $\mathcal{L}_\text{Gen}$ and $\mathbf{H}$ from Eq.~\ref{eq:latent}
\EndFor

\State Sample noise $\mathbf{H}_N$ from Normal distribution
\State Generate pseudo-OOD $\mathbf{H}_\text{p-OOD}$ with $D$ and $\mathbf{H}_T$ from Eq.~\ref{eq:p-ood-dist}
\For{$epoch = 1, \dots, M_2$} \hfill{\textcolor{blue}{// \textbf{Step \protect\redcircle{2}}}}
    \State Train GNN with $\mathcal{L}_\text{CLS}$ and $\mathbf{H}$ Eq.~\ref{eq:cls}
    \State Train GNN and MLP with $\mathcal{L}_\text{Div}$, $\mathbf{H}$ and $\mathbf{H}_\text{p-OOD}$ from Eq.~\ref{eq:div}
\EndFor
\EndWhile
\end{algorithmic}
\end{algorithm}
\vspace{-1cm}
\end{minipage}
\end{wrapfigure}

To facilitate effective optimisation of the implicit adversarial objective in Eq.~\ref{eq:jointObj}, we propose an alternating training schema for GOLD, shown in Algorithm \ref{alg:GOLD}. The intuition is to repetitively generate samples close to the ID embeddings of GNN, and then diverge the energy score distribution of the ID and the synthesised pseudo-OOD data. This process ensures the pseudo-OOD data do not diverge too far from ID, enabling effective pseudo-OOD exposure.

\begin{figure}[!h]
    \centering
    \begin{subfigure}[b]{0.24\textwidth}
        \centering
        \includegraphics[width=\textwidth]{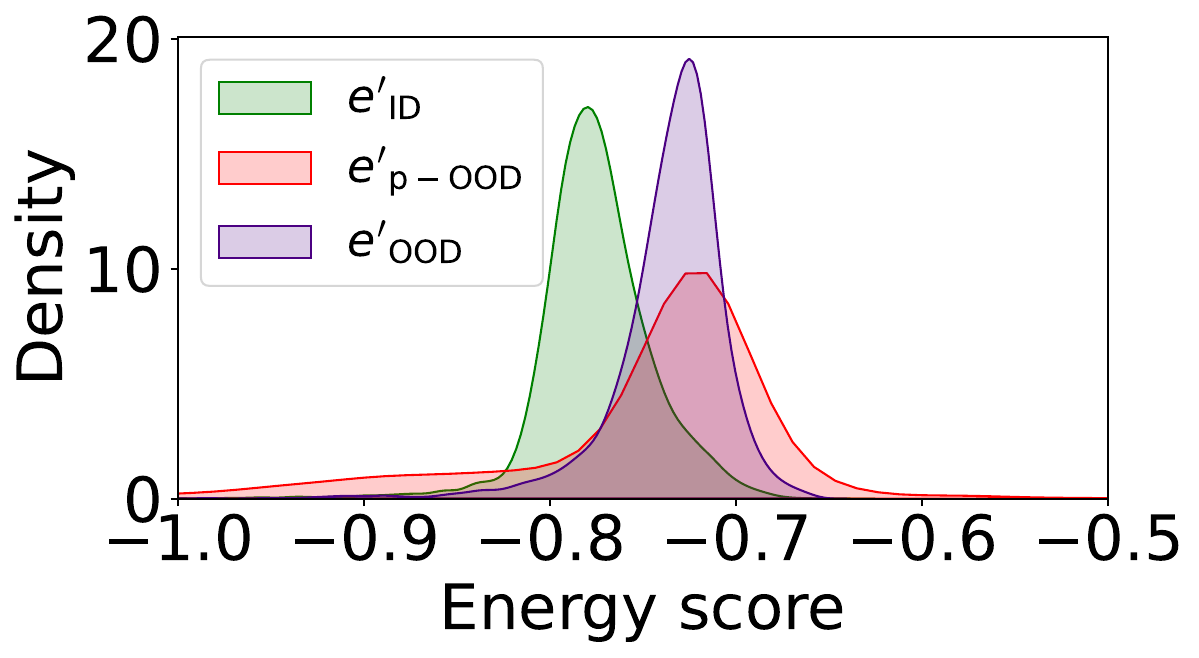}
        \caption{Ep 1, Tr. LGM.}
    \end{subfigure}
    \hfill
    \begin{subfigure}[b]{0.24\textwidth}
        \centering
        \includegraphics[width=\textwidth]{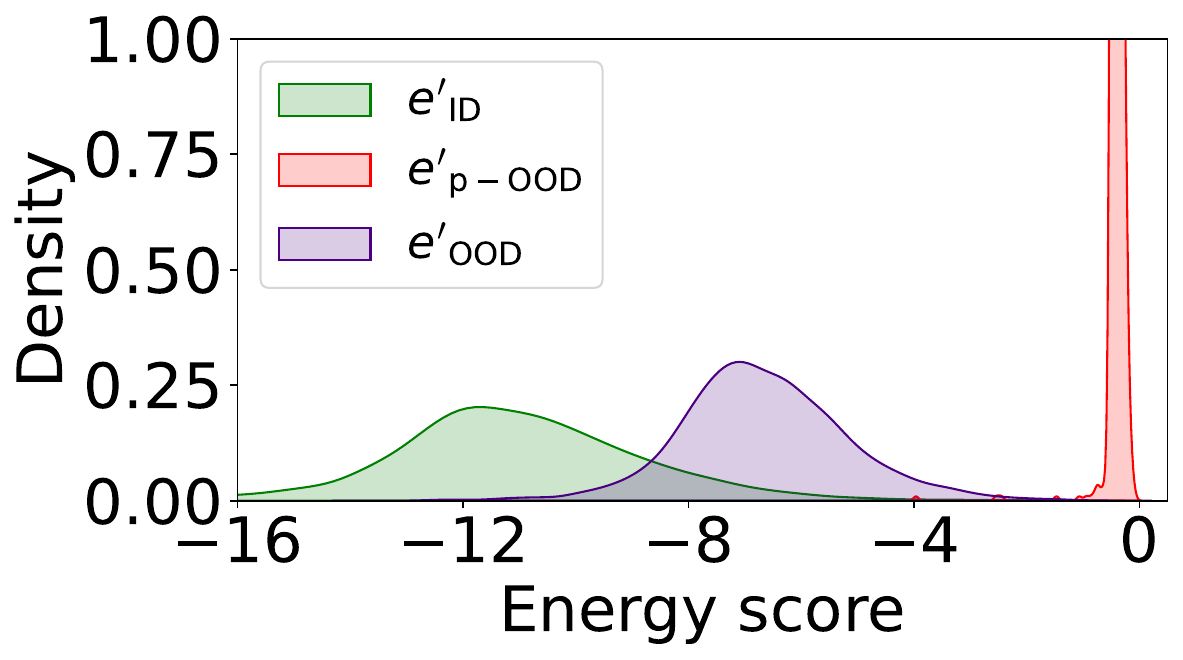}
        \caption{Ep 14, Tr. GNN \& Det.}
    \end{subfigure}
    \hfill
    \begin{subfigure}[b]{0.24\textwidth}
        \centering
        \includegraphics[width=\textwidth]{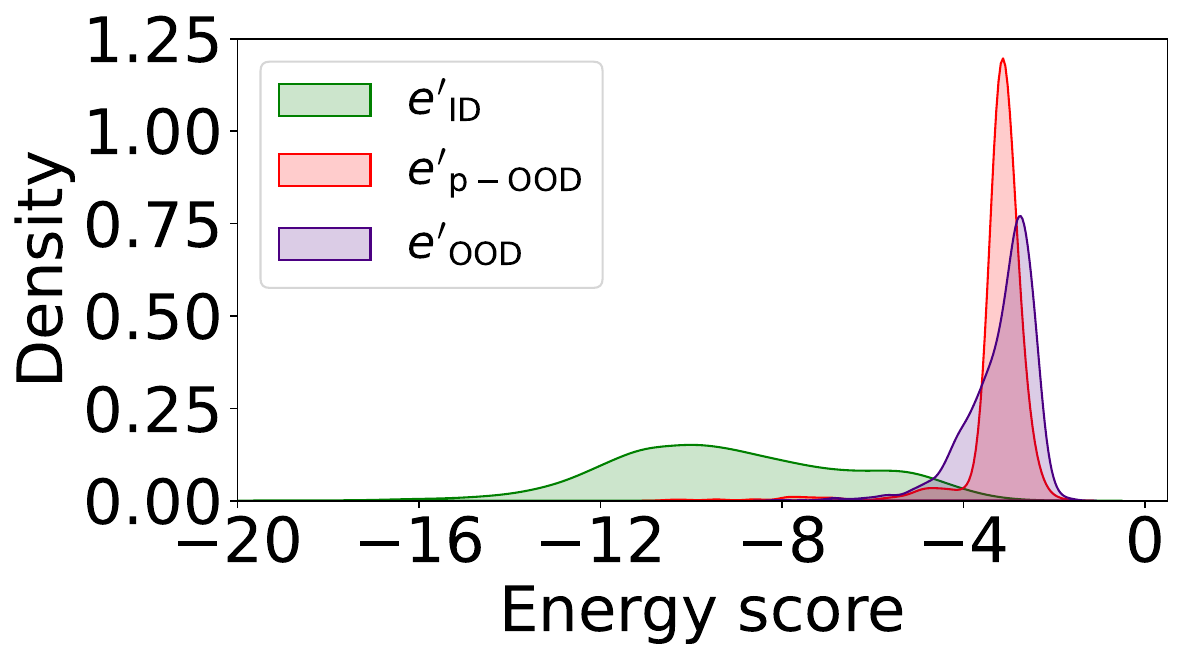}
        \caption{Ep 15, Tr. LGM.}
    \end{subfigure}
    \hfill
    \begin{subfigure}[b]{0.24\textwidth}
        \centering
        \includegraphics[width=\textwidth]{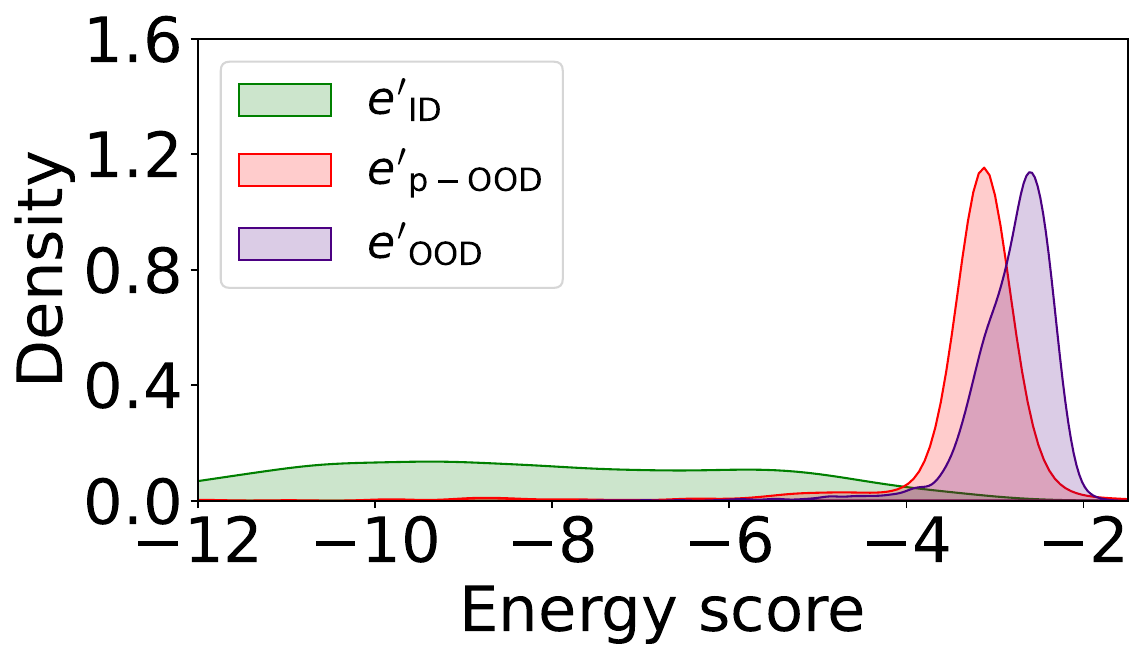}
        \caption{Ep 22, Tr. GNN \& Det.}
    \end{subfigure}
    \caption{Transformed energy $\mathbf{e}'$ distribution during adversarial training (Tr.) on the {\fontfamily{qcr}\selectfont Twitch} dataset for \textcolor{LimeGreen}{in-distribution (ID)}, \textcolor{red}{pseudo (p-)OOD}, and \textcolor{violet}{real OOD} across epochs. \textbf{(a)} shows that after LGM trains to mimic ID data, energy scores are overlapped for ID, p-OOD, and OOD in the initial stages. \textbf{(b)} indicates that after training GNN and the detector to separate the energy of ID and the p-OOD, the real OOD energy cannot be effectively separated from ID. This is a similar situation to the OOD exposure for GNNSafe as in Figure~\ref{fig:twitch_gnnsafe_id_ood}. \textbf{(c)} shows that under adversarial learning, the LGM will be updated to generate p-OOD closer to the updated ID data, preventing it from being too far away from ID data with ineffective OOD learning. \textbf{(d)} displays the final energy distribution after convergence, with real OOD and ID being well separated, while p-OOD and OOD being well aligned.}
    \vspace{-0.5cm}
    \label{fig:adv_energy_vis}
\end{figure}

Thus, GOLD consists of two alternating optimisation steps: \textbf{Step \protect\redcircle{1}}: \textbf{LGM mimicking embeddings of evolving GNN}: A latent generative model is trained to mimic the ID embeddings $\mathbf{H}$, extracted from the $(L-1)$-th layer of an evolving GNN encoder trained in Step 2, to generate pseudo-OOD embeddings. This ensures the pseudo-OOD, $\mathbf{H}_\text{p-OOD}$, is sufficiently close to the ID embeddings and avoids far away and meaningless generation before subsequent divergence (Line 2-7 in Algorithm \ref{alg:GOLD}); \textbf{Step \protect\redcircle{2}}: \textbf{Detector and GNN diverging energy of evolving generator}: The GNN encoder and OOD detector are trained to diverge the energy between the ID embeddings and the pseudo-OOD embeddings generated by the evolving LGM trained in Step 1 (Line 8-11 in Algorithm \ref{alg:GOLD}). Figure \ref{fig:adv_energy_vis} illustrates the adversarial essence of this training paradigm.


\section{Experiments} \label{sec:Experiments}

\paragraph{Datasets.} \label{sec:datasets}
Following \cite{GNNSafe}, five benchmark datasets are used for OOD detection evaluation, including four single-graph datasets: (1) {\fontfamily{qcr}\selectfont Cora}, (2) {\fontfamily{qcr}\selectfont Amazon-Photo}, (3) {\fontfamily{qcr}\selectfont Coauthor-CS}, with synthetic OOD data created via: structure manipulation, feature interpolation, and label leave-out; and (4) {\fontfamily{qcr}\selectfont ogbn-Arxiv}, OOD by year, and (5) one multi-graph scenario: {\fontfamily{qcr}\selectfont TwitchGamers-Explicit}, OOD by different graphs. Detailed splits are provided in Appendix~\ref{Appendix:dataset}.

\paragraph{Baselines.}
We compared GOLD with 12 baseline models, classed into three categories. (1) General non-OOD exposed methods: MSP~\citep{MSP}, ODIN~\citep{ODIN}, Mahalanobis (short for Maha)~\citep{Mahalanobis}, and Energy~\citep{energy}, with GNN used as backbone. 
(2) Graph-specific non-OOD exposed detection methods: GKDE~\citep{GKDE}, GPN~\citep{GPN}, \textsc{GNNSafe}~\citep{GNNSafe}, and \textsc{NODESafe}~\citep{NODESAFE}. (3) Real OOD exposed methods: adopts techniques from computer vision, such as OE~\citep{OE} and Energy FT~\citep{energy}, along with the state-of-the-arts {GNNSafe++}~\citep{GNNSafe} and \textsc{NODESafe++}~\citep{NODESAFE} for graph data. Note that OOD synthesis methods from computer vision~\citep{VOS,Dream-OOD,ConfOOD,NPOS} are not compared due to the non-trivial application from image to graph.
\vspace{-0.1cm}
\paragraph{Metrics.}
The following common practice metrics are used for evaluation: AUROC, AUPR, and FPR95 for OOD detection and Accuracy for ID classification. Metric details are in Appendix \ref{Appendix:Metrics}.
\paragraph{Implementations.}
For a fair comparison, GCN is used as the backbone across all methods, with a layer depth of $2$ and a hidden size of $64$. The propagation iteration $k$ in Eq.~\ref{eq:eprop} is set to $2$, and the controlling parameter $\alpha$ of $0.5$ is used. For LDM, the timestep $T$ is configured within \{$600$, $800$, $1000$\}, $\beta_1 = 10^{-4}$, and $\beta_T = 0.02$. The denoising network $D$ and the MLP detector model are implemented with varying layer and hidden dimension sizes within \{$2$, $3$\} and \{$128$, $256$, $512$\} respectively, subject to the dataset. Additional hyperparameter analysis and parameter details are provided in Appendix \ref{Appendix:latent_generative_model}. We use the Adam optimizer for optimisation~\citep{Adam}.

\begin{table}[t!]
    \centering
    \caption{Model performance comparison: out-of-distribution detection results are measured by \textbf{AUROC} ($\uparrow$) $/$ \textbf{AUPR} ($\uparrow$) $/$ \textbf{FPR95} ($\downarrow$) ($\%$) and in-distribution classification results are measured by accuracy \textbf{(ID ACC)} ($\uparrow$). The average performance of the OOD test sets is reported, with variance reflecting performance differences across distinct test sets. Detailed results for individual subsets are reported in Appendix \ref{Appendix:Additional_exp}. OOD detection performance was prioritised, with the detection results of our Non-OOD exposed GOLD against Non- (Real-) OOD Exposure methods highlighted by {\color{teal}{best}} and {\color{purple}{runner-up}} (\textbf{best} and {\underline{runner-up}}), respectively. Dashed line indicates unavailability.}
    \resizebox{1\linewidth}{!}{
    \begin{tabular}{c|c|cccccccc|cccc|cc}
    \specialrule{.1em}{.05em}{.05em} 
    \multirow{2}{*}{} & \multirow{2}{*}{\textbf{Metrics}} & \multicolumn{8}{c|}{\textbf{Non-OOD Exposure}} & \multicolumn{4}{c|}{\textbf{Real OOD Exposure}} & \multicolumn{2}{c}{\textbf{GOLD (Non-OOD)}}\\
    & & \textbf{MSP} & \textbf{ODIN} & \textbf{Maha} & \textbf{Energy} & \textbf{GKDE} & \textbf{GPN} & \textbf{\textsc{GNNSafe}} & \textbf{\textsc{NODESafe}} & \textbf{OE} & \textbf{Energy FT} & \textbf{\textsc{GNNSafe++}} & \textbf{\textsc{NODESafe++}} & \textbf{w/ VAE} & \textbf{w/ LDM}\\
    \midrule
    \multirow{4}{*}{\rotatebox{90}{{\fontfamily{qcr}\selectfont Twitch}}} 
        & AUROC & 33.59 & 58.16 & 55.68 & 51.24 & 46.48 & 51.73 & 66.82 & \textcolor{purple}{89.99} & 55.72 & 84.50 & 95.36 & \underline{98.50} & 99.26 & \textcolor{teal}{\textbf{99.46 $\pm$ 0.09}} \\
        & AUPR  & 49.14 & 72.12 & 66.42 & 60.81 & 62.11 & 66.36 & 70.97 & \textcolor{purple}{93.33} & 70.18 & 88.04 & 97.12 & \underline{99.18} & 98.54 & \textcolor{teal}{\textbf{99.62 $\pm$ 0.06}} \\
        & FPR95   & 97.45 & 93.96 & 90.13 & 91.61 & 95.62 & 95.51 & 76.24 & \textcolor{purple}{47.00} & 95.07 & 61.29 & 33.57 & \underline{3.43} & 3.03 & \textcolor{teal}{\textbf{1.78 $\pm$ 0.43}} \\
        & ID ACC & 68.72 & 70.79 & 70.51 & 70.40 & 67.44 & 68.09 & 70.40 & 71.79 & 70.73 & 70.52 & 70.18 & 71.85 & 68.50 & 68.49 $\pm$ 0.13 \\
    \midrule
    \multirow{4}{*}{\rotatebox{90}{{\fontfamily{qcr}\selectfont Cora}}} 
        & AUROC & 82.55 & 49.87 & 54.74 & 83.09 & 69.54 & 84.56 & 91.25 & \textcolor{purple}{94.39} & 79.76 & 85.13 & 92.98 & \underline{95.36} & 89.96 & \textcolor{teal}{\textbf{95.84 $\pm$ 0.69}} \\
        & AUPR  & 65.82 & 26.08 & 34.43 & 66.21 & 46.09 & 68.02 & 82.62 & \textcolor{purple}{86.01} & 64.93 & 67.89 & 84.93 & \underline{88.08} & 93.19 & \textcolor{teal}{\textbf{91.17 $\pm$ 2.59}}\\
        & FPR95   & 62.39 & 100.00 & 96.30 & 65.21 & 80.51 & 58.30 & 47.38 & \textcolor{purple}{26.04} & 75.22 & 51.03 & 38.44 & \underline{20.20} & 28.66 & \textcolor{teal}{\textbf{17.83 $\pm$ 3.78}} \\
        & ID ACC & 79.91 & 79.61 & 79.57 & 80.34 & 79.86 & 81.65 & 80.37 & 81.92 & 77.69 & 80.44 & 81.45 & 81.65 & 76.79 & 81.66 $\pm$ 7.94 \\
    \midrule
    \multirow{4}{*}{\rotatebox{90}{{\fontfamily{qcr}\selectfont Amazon}}}    
        & AUROC & 96.52 & 80.12 & 73.81 & 96.73 & 66.98 & 92.60 & \textcolor{purple}{98.49} & - & 97.79 & 98.04 & \textbf{98.99} & - & 98.68 & \textcolor{teal}{\underline{98.81 $\pm$ 1.40}} \\
        & AUPR  & 95.01 & 77.18 & 72.35 & 95.16 & 71.18 & 90.50 & \textcolor{purple}{98.62} & - & 97.26 & 96.96 & \underline{98.88} & - & 98.89 & \textcolor{teal}{\textbf{98.92 $\pm$ 1.31}} \\
        & FPR95   & 13.83 & 85.22 & 83.44 & 13.15 & 98.47 & 32.64 & \textcolor{purple}{2.30} & - &  7.52 &  5.98 &  \underline{2.10} & - & 5.11 & \textcolor{teal}{\textbf{2.07 $\pm$ 3.46}} \\
        & ID ACC & 93.83 & 93.88 & 93.80 & 93.85 & 87.71 & 89.54 & 93.70 & - & 93.54 & 93.38 & 93.48 & - & 89.91 & 92.99 $\pm$ 1.90 \\
    \midrule
    \multirow{4}{*}[0.22em]{\rotatebox{90}{{\fontfamily{qcr}\selectfont Coauthor}}} 
        & AUROC & 95.74 & 51.71 & 82.02 & 96.64 & 69.24 & 69.89 & \textcolor{purple}{98.82} & - & 97.65 & 98.17 & \textbf{99.28} & - & 98.78 & \textcolor{teal}{\underline{99.00 $\pm$ 1.19}} \\
        & AUPR  & 96.43 & 56.37 & 87.05 & 97.09 & 80.17 & 72.77 & \textcolor{purple}{99.44} & - & 98.04 & 98.51 & \textbf{99.73} & - & 96.40 & \textcolor{teal}{\underline{99.56 $\pm$ 0.43}} \\
        & FPR95   & 21.37 & 99.97 & 48.09 & 15.49 & 97.04 & 69.60 &  \textcolor{purple}{4.28} & - & 10.61 &  7.76 &  \underline{3.18} & - & 4.66 & \textcolor{teal}{\textbf{3.16 $\pm$ 5.46}} \\
        & ID ACC & 93.37 & 93.29 & 93.29 & 93.57 & 87.74 & 89.39 & 93.56 & - & 93.41 & 93.44 & 93.68 & - & 92.22 & 92.69 $\pm$ 1.87 \\
    \midrule
    \multirow{4}{*}{\rotatebox{90}{{\fontfamily{qcr}\selectfont Arxiv}}}    
        & AUROC & 63.91 & 55.07 & 56.92 & 64.20 & 58.32 & OOM & 71.06 & \textcolor{purple}{72.44} & 69.80 & 71.56 & 74.77 & \textbf{75.49} & 71.52 & \textcolor{teal}{\underline{73.90 $\pm$ 0.11}} \\
        & AUPR  & 75.85 & 68.85 & 69.63 & 75.78 & 72.62 & OOM & 80.44 & \textcolor{purple}{81.51} & 80.15 & 80.47 & 83.21 & \textbf{83.71} & 80.25 & \textcolor{teal}{\underline{82.52 $\pm$ 0.12}} \\
        & FPR95   & 90.59 & 100.0 & 94.24 & 90.80 & 93.84 & OOM & 87.01 & \textcolor{purple}{84.27}& 85.16 & 80.59 & 77.43 & \textbf{75.24} & 81.95 & \textcolor{teal}{\underline{80.57 $\pm$ 0.32}} \\
        & ID ACC & 53.78 & 51.39 & 51.59 & 53.36 & 50.76 & OOM & 53.39 & 51.20 & 52.39 & 53.26 & 53.50 & 52.93 & 49.70 & 50.59 $\pm$ 0.53 \\
    \specialrule{.1em}{.05em}{.05em} 
     \end{tabular}}
    \vspace{-0.3cm}
    \label{Table:overall_performance}
\end{table}

\subsection{Overall Performance} \label{sec:overall_performance}

\paragraph{Our Non-OOD exposed GOLD can outperform Non-OOD exposure methods and is competitive with Real OOD exposed methods.} As shown in Table~\ref{Table:overall_performance}, GOLD with LDM consistently surpasses the state-of-the-art non-OOD exposure methods \textsc{NODESafe} and \textsc{GNNSafe} by a large margin across all datasets, as indicated by the teal colouring. When using VAE as LGM, the OOD detection performance is very close while being more lightweight due to the model design. GOLD with VAE can achieve state-of-the-art performance especially when the datasets are challenging for general methods, like {\fontfamily{qcr}\selectfont Twitch} and {\fontfamily{qcr}\selectfont Arxiv}. Additionally, considering LDM as the generative model, GOLD can largely outperform \textsc{GNNSafe++} and achieves better performance than the SOTA \textsc{NODESafe++} in OOD detection for {\fontfamily{qcr}\selectfont Twitch} and {\fontfamily{qcr}\selectfont CORA}, as highlighted in bold. While for {\fontfamily{qcr}\selectfont Amazon}, {\fontfamily{qcr}\selectfont Coauthor}, and {\fontfamily{qcr}\selectfont Arxiv} dataset, GOLD can achieve a comparable performance with \textsc{GNNSafe++} while not significantly surpassing them. The reason can be two-fold. For {\fontfamily{qcr}\selectfont Amazon} and {\fontfamily{qcr}\selectfont Coauthor}, the classifier and the OOD detector are already in high performance, which leads to the fact that the energy from the classifier and the information given by the real OOD data have already been well utilised. The pseudo-OOD generation in GOLD cannot provide much more useful supervision signals for the detector. Nonetheless, GOLD still largely outperforms the non-OOD exposure. While for the {\fontfamily{qcr}\selectfont Arxiv} dataset, the OOD situation is defined by time, which leads to a huge boost of OOD information when exposing a real OOD dataset. In contrast, for GOLD, the pseudo-OOD generation is largely limited by the ID accuracy of the classifier at 50\%. A more detailed table with individual OOD test set performance and variance can be found in Appendix \ref{Appendix:Additional_exp}.


Since GOLD uses \textsc{GNNSafe} as the backbone, the following detailed experiments are mainly conducted based on the comparison with the base \textsc{GNNSafe/++} approach. GOLD with LDM is used as the default model without specific notation.

\subsection{Visualisation of Energy Score Gap} \label{sec:vis}
\begin{figure}[!h]
    \centering
    \begin{subfigure}[b]{0.3\textwidth}
        \centering
        \includegraphics[width=\textwidth]{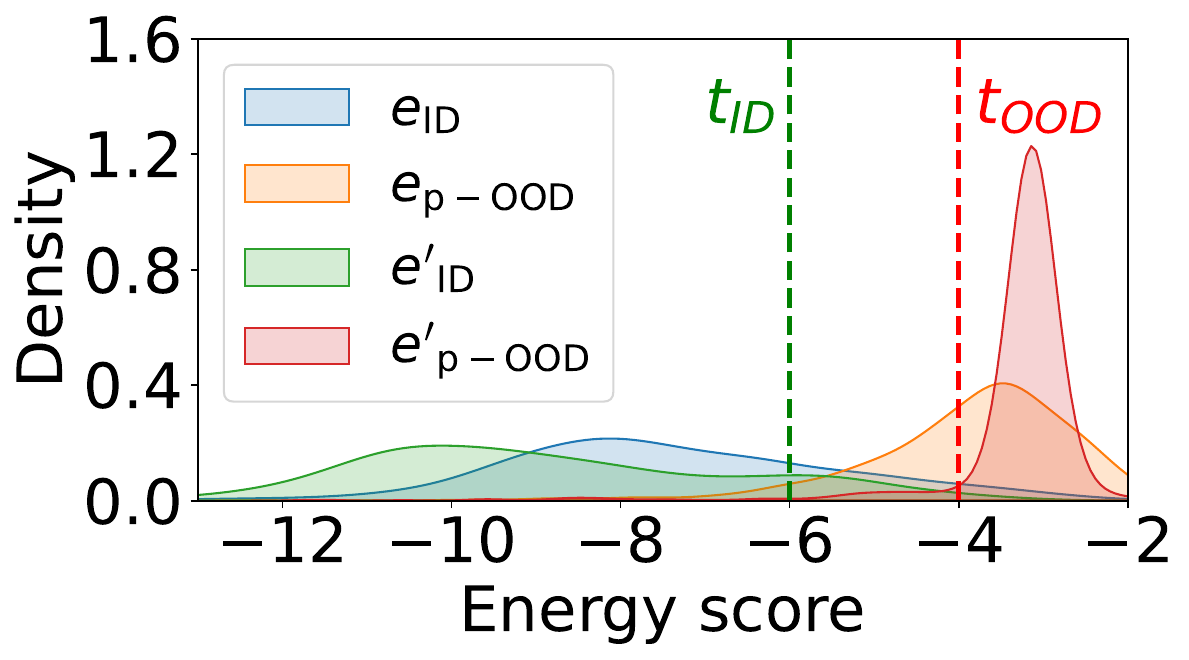}
        \caption{{\fontfamily{qcr}\selectfont Twitch} training (GOLD).}
        \label{fig:twitch_score_gap}
    \end{subfigure}
    \hfill
    \begin{subfigure}[b]{0.3\textwidth}
        \centering
        \includegraphics[width=\textwidth]{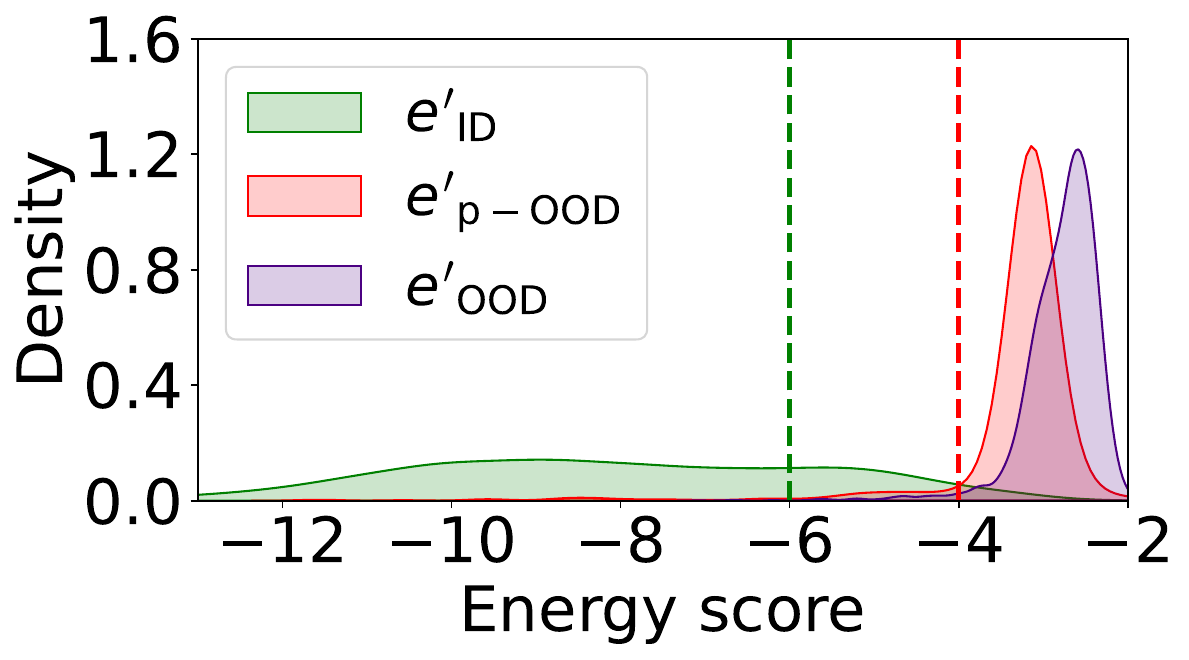}
        \caption{{\fontfamily{qcr}\selectfont Twitch} test (GOLD).}
        \label{fig:twitch_id_ood}
    \end{subfigure}
    \hfill
    \begin{subfigure}[b]{0.3\textwidth}
        \centering
        \includegraphics[width=\textwidth]{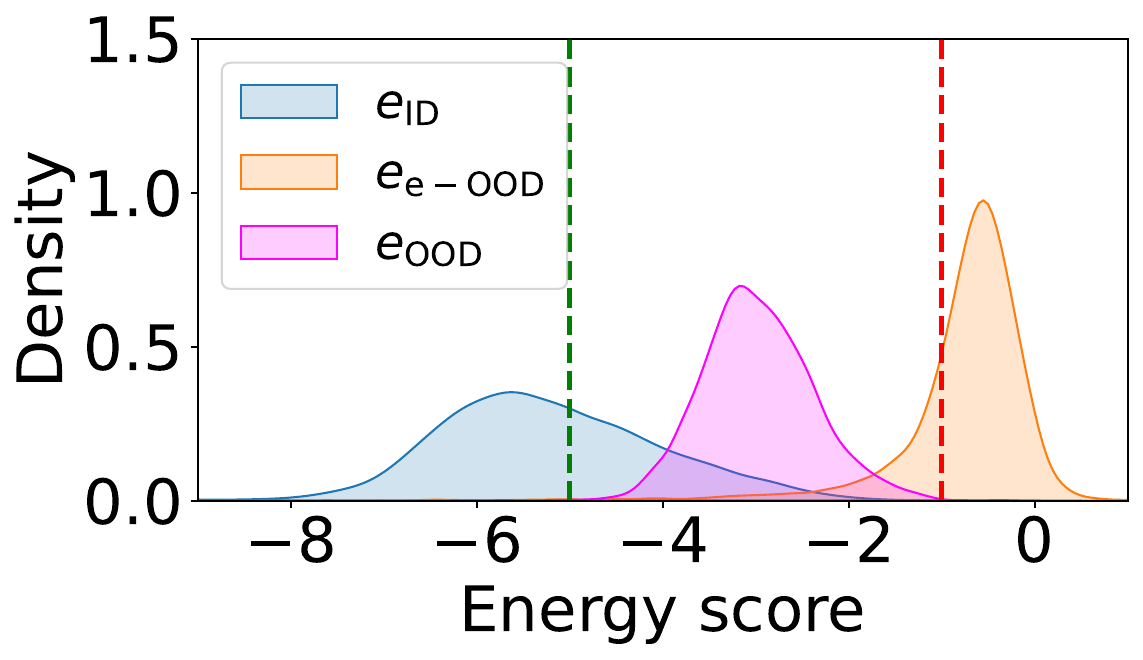}
        \caption{{\fontfamily{qcr}\selectfont Twitch} test (GNNSafe++).}
        \label{fig:twitch_gnnsafe_id_ood}
    \end{subfigure}
    \hfill
    \begin{subfigure}[b]{0.3\textwidth}
        \centering
        \includegraphics[width=\textwidth]{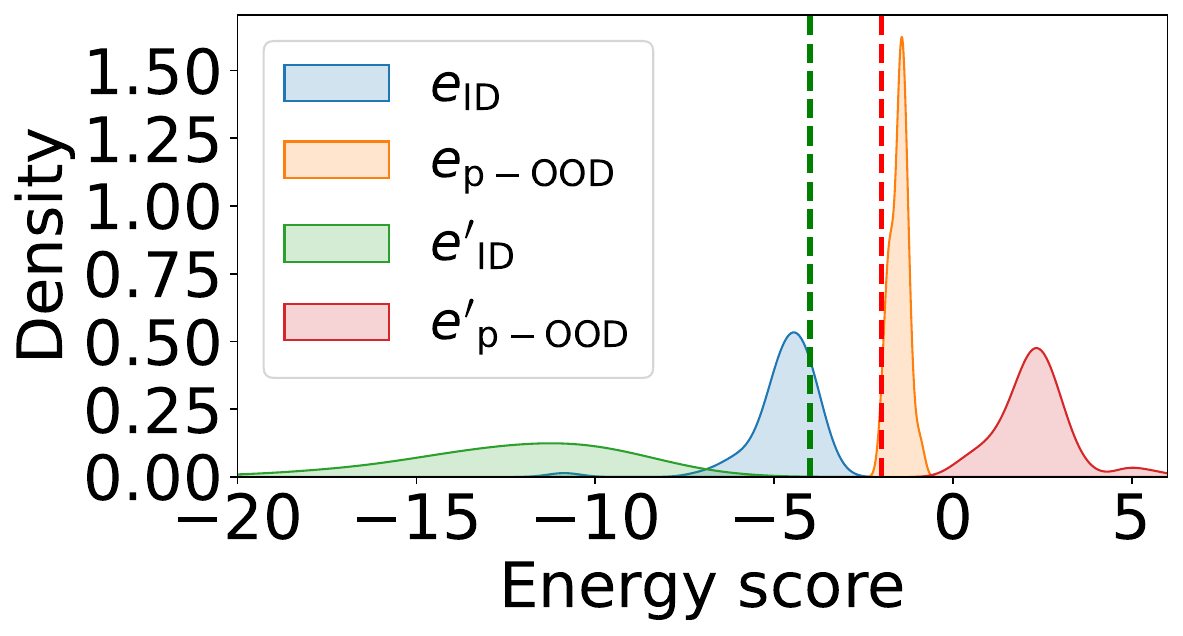}
        \caption{{\fontfamily{qcr}\selectfont Cora-L} training (GOLD).}
        \label{fig:cora_score_gap}
    \end{subfigure}
    \hfill
    \begin{subfigure}[b]{0.3\textwidth}
        \centering
        \includegraphics[width=\textwidth]{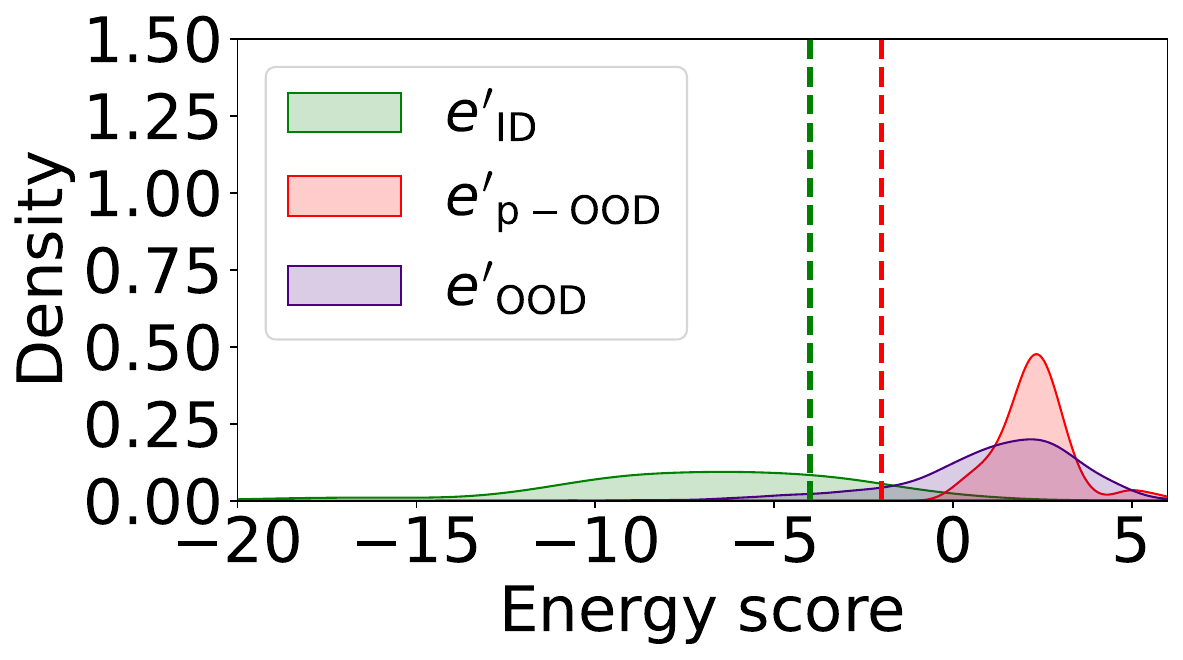}
        \caption{{\fontfamily{qcr}\selectfont Cora-L} test (GOLD).}
        \label{fig:cora_id_ood}
    \end{subfigure}
    \hfill
    \begin{subfigure}[b]{0.3\textwidth}
        \centering
        \includegraphics[width=\textwidth]{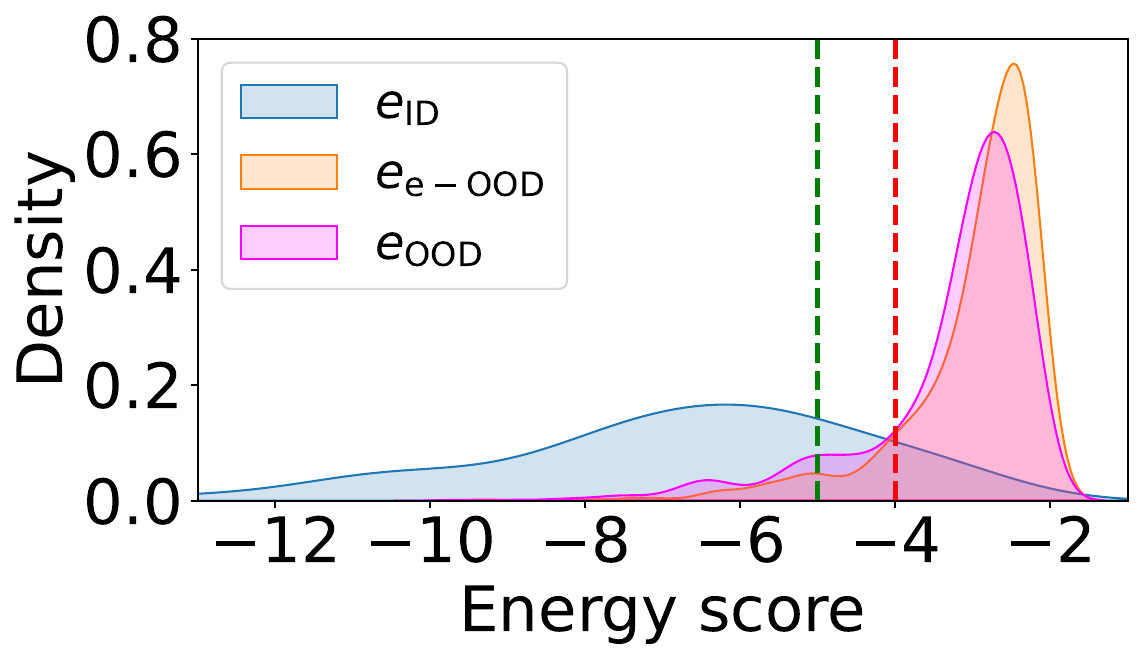}
        \caption{{\fontfamily{qcr}\selectfont Cora-L} test (GNNSafe++).}
        \label{fig:coral_gnnsafe_id_ood}
    \end{subfigure}
    \caption{Energy score distributions for {\fontfamily{qcr}\selectfont Twitch} and {\fontfamily{qcr}\selectfont Cora-L} with GOLD and \textsc{GNNSafe++}. The vertical green (red) dashed lines represent the thresholds $t_\text{ID}$ ($t_\text{OOD}$) from Eq.~\ref{eq: ereg}. $\mathbf{e}$ denotes original energy scores from the GNN, while $\mathbf{e}'$ are transformed scores from the detector, with subscripts for ID, OOD, p-OOD (pseudo), and e-OOD (exposed) data. \textbf{(a) \& (d)} show that transformed energy $\mathbf{e}'$ (\textcolor{LimeGreen}{green} and \textcolor{red}{red}) can be further diverged from the original energy $\mathbf{e}$ (\textcolor{Cyan}{blue} and \textcolor{orange}{orange}). \textbf{(b) \& (e)} indicate that GOLD can align the transformed energy $\mathbf{e}'$ for pseudo OOD (\textcolor{red}{red}) and real OOD (\textcolor{violet}{purple}) in testing. At the same time, the transformed energy $\mathbf{e}'$ of ID (\textcolor{LimeGreen}{green}) can be separated. \textbf{(c) \& (f)} demonstrate that energy separation of test ID (\textcolor{Cyan}{blue}) and OOD (\textcolor{Thistle}{pink}) in \textsc{GNNSafe++} is not effective, such that although the exposed OOD (\textcolor{orange}{orange}) can diverge far away from the ID (\textcolor{Cyan}{blue}), the real OOD (\textcolor{Thistle}{pink}) is still closer to the ID (\textcolor{Cyan}{blue}).}
    \label{fig:Energy_Score_Distribution}
\end{figure}

This experiment presents the energy distribution of GOLD and \textsc{GNNSafe}. Figures \ref{fig:twitch_score_gap} and \ref{fig:cora_score_gap} display \textbf{a distinct separation in the energy scores of ID and p-OOD, as well as ID and OOD, produced by the detector}, exemplifying the effectiveness of GOLD in distinguishing and amplifying the energy margin between ID and (p-)OOD data. Furthermore, Figures \ref{fig:twitch_id_ood} and \ref{fig:cora_id_ood} illustrate the energy score distributions of the test ID data, synthetic OOD data, and the test OOD data. These figures reveal an optimal and almost disjoint between ID and OOD data, where the thresholds $t_\text{ID}$ and $t_\text{OOD}$ indicate a clear energy boundary, thereby indicating the efficacy of GOLD in simulating pseudo-OOD data to facilitate effective OOD detection. Compared with the energy distribution of test ID data, test OOD data and exposed OOD data from \textsc{GNNSafe} in Figures~\ref{fig:twitch_gnnsafe_id_ood} and~\ref{fig:coral_gnnsafe_id_ood}, GOLD can further separate the energy scores between the test ID and OOD data with the pseudo-OOD data.

\begin{wraptable}{r}{0.5\linewidth}
    \centering
    \vspace{-1.3cm}
    \caption{Ablation study.}
    \vspace{-0.3cm}
    \resizebox{1\linewidth}{!}{
    \begin{tabular}{c|c|cc|cc|c}
    \toprule
    & \textbf{Metrics} & \textbf{\textsc{GNNSafe}} & \textbf{\textsc{GNNSafe++}} & \textbf{w/o Adv.} & \textbf{w/o Det.} & \textbf{GOLD}\\
    \midrule
    \multirow{4}{*}{\rotatebox{90}{\fontfamily{qcr}\selectfont Twitch}} 
        & AUROC & 66.82 & \underline{95.36} & 84.59 & \textcolor{purple}{77.70} & \textcolor{teal}{\textbf{99.46}} \\
        & AUPR  & 70.97 & \underline{97.12} & 88.69 & \textcolor{purple}{83.91} & \textcolor{teal}{\textbf{99.62}}\\
        & FPR95 & \textcolor{purple}{76.24} & \underline{33.57} & 59.71  & 79.84 & \textcolor{teal}{\textbf{1.78}}\\
        & ID ACC & 70.40 & 70.18 & 70.97 & 70.97  & 68.49\\
    \midrule
    \multirow{4}{*}{\rotatebox{90}{\fontfamily{qcr}\selectfont Cora}} 
        & AUROC & 91.25 & 92.98 & 89.64 & \textcolor{purple}{\underline{93.43}} & \textcolor{teal}{\textbf{95.84}} \\
        & AUPR  & 82.62 & 84.93 & 80.22 & \textcolor{purple}{\underline{86.78}} & \textcolor{teal}{\textbf{91.17}} \\
        & FPR95 & 47.38 & 38.44 & 46.33 & \textcolor{purple}{\underline{34.01}} & \textcolor{teal}{\textbf{17.83}} \\
        & ID ACC & 80.37 & 81.45 & 77.60 & 80.70 & 81.66\\
    \midrule                        
    \multirow{4}{*}{\rotatebox{90}{\fontfamily{qcr}\selectfont Arxiv}}
        & AUROC & \textcolor{purple}{71.06} & \textbf{74.77} & 69.76 & 69.91 & \textcolor{teal}{\underline{73.90}}\\
        & AUPR  & \textcolor{purple}{80.44} & \textbf{83.21} & 78.93 & 79.05 & \textcolor{teal}{\underline{82.52}}\\
        & FPR95 & \textcolor{purple}{87.01} & \textbf{77.43} & 88.16 & 89.67 & \textcolor{teal}{\underline{80.57}}\\
        & ID ACC & 53.39 & 53.50 & 49.89 & 49.66 & 50.59\\
    \specialrule{.1em}{.05em}{.05em} 
    \end{tabular}}
    \label{Table:Ablation_general}
    \vspace{-0.5cm}
\end{wraptable}
\subsection{Ablation Study}
\label{sec:ablation}
In the ablation study, two variants are studied: (1) pre-training the LDM without the adversarial pipeline (w/o Adv.), and (2) removing the MLP detector, using GNN energy scores instead (w/o Det.). In Table~\ref{Table:Ablation_general}, \textbf{both the adversarial training paradigm and the new detector significantly contribute to the GOLD}. The results reveal that without adversarial learning, the OOD detection performance has a significant drop for all situations. This underscores the efficacy of the adversarial framework in pseudo-OOD exposure. Moreover, we observe a more unstable performance when removing the detector. In this scenario, the model can still surpass other baselines on datasets like {\fontfamily{qcr}\selectfont Cora}, but it shows a significant drop on others. This juxtaposition exemplifies the necessity of the detector model in GOLD. Nonetheless, the results illustrate the importance of integrating all components to enhance the model's OOD detection capabilities. We provide additional explanation and visualisation of the ablation study in Appendix \ref{Appendix:ablation_vis}.

\begin{wraptable}{r}{0.5\linewidth}
    \vspace{-1.3cm}
    \centering
    \caption{Adversarial training analysis.}
    \vspace{-0.3cm}
    \resizebox{1\linewidth}{!}{
    \begin{tabular}{c|c|c|ccc|c}
    \toprule
    & \textbf{Metrics} & \textbf{\textsc{GNNSafe++}}& \textbf{Gen.\ Once}  & \textbf{Gen.\ Multi} & \textbf{Real OOD} & \textbf{GOLD}\\
    \midrule
    \multirow{4}{*}{\rotatebox{90}{\fontfamily{qcr}\selectfont Twitch}} 
        & AUROC & 95.36& 84.59  & 84.33 & \underline{97.58} & \textbf{99.46}\\
        & AUPR  & 97.12& 88.69  & 88.38 & \underline{98.50} & \textbf{99.62}\\
        & FPR95  & 33.57& 59.71  & 57.00 & \underline{14.39} & \textbf{1.78}\\
        & ID ACC & 70.18& 70.97  & 71.12 & 70.45 & 68.49\\
    \midrule
    \multirow{4}{*}{\rotatebox{90}{\fontfamily{qcr}\selectfont Cora}} 
        & AUROC & 92.98& 89.64  & 92.83 & \underline{95.59}   & \textbf{95.84} \\
        & AUPR  & 84.93 & 80.22 & 85.09 & \underline{90.05} & \textbf{91.17} \\
        & FPR95 & 38.44& 46.33  & 30.11 & \underline{21.54}  & \textbf{17.83} \\
        & ID ACC & 81.45& 77.60  & 80.56 & 78.42  & 81.66\\
    \midrule    
    \multirow{4}{*}{\rotatebox{90}{\fontfamily{qcr}\selectfont Arxiv}}   
        & AUROC & \underline{74.77} & 69.76   & 72.15  & \textbf{78.90} & 73.90\\
        & AUPR  & \underline{83.21}& 78.93 & 80.57  & \textbf{85.46} &82.52\\
        & FPR95  & \underline{77.43}& 88.16  & 82.02  & \textbf{68.94} & 80.57\\
        & ID ACC & 53.50& 49.89 & 50.77 & 49.99 & 50.59 \\
    \specialrule{.1em}{.05em}{.05em} 
    \end{tabular}} \label{Table:Ablation_adversarial}
    \vspace{-0.3cm}
\end{wraptable}
\subsection{Adversarial Training Analysis}
To further assess the proposed adversarial training framework, three variants of (pseudo-) OOD exposure are studied: (1) using an ID-pretrained LDM to generate once to train GOLD (Gen.\ Once), which is the same as w/o Adv. in Section~\ref{sec:ablation}; (2) using an ID-pretrained LDM to generate multiple rounds of pseudo-OOD along the GOLD training loops (Gen.\ Multi); (3) using real OOD data instead of pseudo-OOD to train GOLD (Real OOD).
The results are detailed in Table \ref{Table:Ablation_adversarial}, which highlights that \textbf{using an ID-pre-trained generative model would not improve OOD detection performance}. This is because without the adversarial training, the detector will be biased by a set of inaccurate and close-to-ID pseudo-OOD data generated by the pre-trained diffusion model. When incorporating real OOD data to substitute the pseudo-OOD in our framework, the Real OOD variant can achieve consistently better performance than Gen.\ Once and Gen.\ Multi. For {\fontfamily{qcr}\selectfont Arxiv}, Real OOD can surpass our default GOLD model with the advantage of OOD exposure in this dataset. Furthermore, a comparison of the results after removing the adversarial process highlights the superiority of the adversarial framework, as all adversarial-based methods outperform their non-adversarial baselines. This robust set of results validates the efficacy of our adversarial training paradigm in enhancing model performance for OOD detection. Despite these modifications, our synthetic-based OOD detection continues to maintain strong performance.

\subsection{Effectiveness of Energy Regulariser}
Extending beyond the previous analysis, we observed that the energy regularisers in GOLD are important factors for OOD detection, especially for the divergence regularisation. We provide a comprehensive assessment of the energy regularisers, $\mathcal{L}_\text{Unc}$ from Eq.~\ref{eq:uncertainty}, $\mathcal{L}_\text{EReg}$ from Eq.~\ref{eq: ereg}, and $\mathcal{L}_\text{DReg}$ from Eq.~\ref{eq:dreg}, across three datasets: {\fontfamily{qcr}\selectfont Twitch}, {\fontfamily{qcr}\selectfont Cora}, and {\fontfamily{qcr}\selectfont Amazon}, reporting the average performance across subsets in Table~\ref{Table:Ablation_regulariser}. \textbf{The default GOLD that incorporates all regularisers, consistently shows superior performance across all datasets, effectively indicating the contribution of the energy regularisers in OOD detection}.  Notably, each dataset exhibits different sensitivities to the absence or presence of specific regularisers. For instance, all datasets are significantly affected by the removal of $\mathcal{L}_\text{DReg}$, highlighting its critical role. There is a substantial performance drop for {\fontfamily{qcr}\selectfont Cora} without $\mathcal{L}_\text{EReg}$. Additionally, individual regulariser performance is context-dependent, with $\mathcal{L}_\text{DReg}$ emerging as particularly impactful, often driving better outcomes when combined with either of the other two regularisers. This is reflected in the best runner-up results, where $\mathcal{L}_\text{DReg}$ is combined with another regulariser, underscoring its influence as the most impactful of the three. Nonetheless, this analysis demonstrates the effectiveness of a holistic approach of combining all proposed regularisers, as shown by GOLD's consistently high performance across all metrics and datasets. The extended performance of each subset is provided in Appendix \ref{Appendix:Additional_exp}.

\begin{table}[!h]
    \centering
    \vspace{-0.1cm}
    \caption{Energy regulariser analysis.}
    \vspace{-0.3cm}
    \resizebox{1\linewidth}{!}{
    \begin{tabular}{ccc|cccc|cccc|cccc}
    \toprule
    \multirow{2}{*}{$\mathcal{L}_\text{Unc}$} & \multirow{2}{*}{$\mathcal{L}_\text{EReg}$} & \multirow{2}{*}{$\mathcal{L}_\text{DReg}$} & \multicolumn{4}{c|}{{\fontfamily{qcr}\selectfont Twitch}} & \multicolumn{4}{c|}{{\fontfamily{qcr}\selectfont Cora}}& \multicolumn{4}{c}{{\fontfamily{qcr}\selectfont Amazon}}\\
    & & & \textbf{AUROC} & \textbf{AUPR} & \textbf{FPR} & \textbf{ID Acc} & \textbf{AUROC} & \textbf{AUPR} & \textbf{FPR} & \textbf{ID Acc} & \textbf{AUROC} & \textbf{AUPR} & \textbf{FPR} & \textbf{ID Acc}\\
    \midrule
     &  &  & 
    86.44 & 80.64 & 79.84 & 68.97 &
    61.14 & 57.82 & 89.70 & 76.23 & 
    64.17 & 72.67 & 46.72 & 92.07 \\
    \midrule
    \checkmark & & & 
    10.18 & 40.62 & 97.84 & 70.15 &
    70.76 & 65.12 & 94.37 & 81.05& 
    67.20 & 71.73 & 68.13 & 93.70 \\
     & \checkmark& & 
    78.02& 83.37& 78.90& 70.98&
    66.00 & 63.32 & 54.06 & 80.44 & 
    48.65 & 61.93 & 77.91 & 93.45 \\
     &  & \checkmark& 
    69.04 & 76.88& 44.54& 70.79&
    84.39 & 74.57 & 68.49 & 76.08 & 
    97.72 & 96.83 & 8.28 & 92.79\\
    \midrule
    \checkmark & \checkmark&  &
    76.88 & 81.49 & 76.14 & 70.99 &
    34.63 & 39.27 & 96.84 &81.25 & 
    71.15  & 63.84 & 74.85 & 93.30\\
    \checkmark &  & \checkmark& 
    64.43 & 75.46 & 45.95 & 70.90&
    \textcolor{purple}{94.03} & 73.89 & 73.54 & 74.53 & 
    97.91 & 97.03 & 4.54 & 93.20 \\
    & \checkmark& \checkmark& 
    \textcolor{purple}{89.58} & \textcolor{purple}{93.12} & \textcolor{purple}{43.78} & 69.64 &
    93.28 & \textcolor{purple}{87.50} & \textcolor{purple}{31.04} & 79.88& 
    \textcolor{purple}{98.02} & \textcolor{purple}{98.42} & \textcolor{purple}{3.40} & 92.81 \\
    \midrule
    & GOLD & & 
    \textcolor{teal}{99.46} & \textcolor{teal}{99.62} & \textcolor{teal}{1.78} & 68.49 & 
    \textcolor{teal}{95.84} & \textcolor{teal}{91.17} & \textcolor{teal}{17.83} & 81.66& 
    \textcolor{teal}{98.81} & \textcolor{teal}{98.92} & \textcolor{teal}{2.07} & 92.99\\
    \specialrule{.1em}{.05em}{.05em} 
    \end{tabular}}
    \label{Table:Ablation_regulariser}
    \vspace{-0.3cm}
\end{table}


\begin{wraptable}{r}{0.6\linewidth}
    \vspace{-1.2cm}
    \centering
    \caption{Inference and training time (s) of GOLD.}
    \vspace{-0.2cm}
    \resizebox{\linewidth}{!}{
    \begin{tabular}{l|cc|cc|cc|cc}
    \toprule
       & \multicolumn{2}{c|}{\textsc{GNNSafe}} & \multicolumn{2}{c|}{\textsc{GNNSafe++}} & \multicolumn{2}{c|}{GOLD w/ VAE} & \multicolumn{2}{c}{GOLD w/ LDM}\\
        & Inf. & Train.  & Inf. & Train.  & Inf. & Train. & Inf. & Train.\\
    \midrule
    \texttt{Twitch}  & \textbf{\textcolor{ForestGreen}{0.08}}  & 2.41  & \textbf{\textcolor{ForestGreen}{0.09}}  & 4.74    & \textbf{\textcolor{ForestGreen}{0.09}}  & 2.78  & \textbf{\textcolor{ForestGreen}{0.10}}  & 8.96   \\
    \texttt{Cora-F}  & \textbf{\textcolor{ForestGreen}{0.03}}  & 4.40  & \textbf{\textcolor{ForestGreen}{0.03}}  & 5.32  & \textbf{\textcolor{ForestGreen}{0.04}}  & 3.91  & \textbf{\textcolor{ForestGreen}{0.04}}  & 5.93   \\
    \texttt{Amazon-F}     & \textbf{\textcolor{ForestGreen}{0.04}}  & 13.51     & \textbf{\textcolor{ForestGreen}{0.05}}  & 18.40   & \textbf{\textcolor{ForestGreen}{0.05}}  & 12.52 & \textbf{\textcolor{ForestGreen}{0.07}}  & 39.04    \\
    \texttt{Coauthor-F}   & \textbf{\textcolor{ForestGreen}{0.35}}  & 57.80    & \textbf{\textcolor{ForestGreen}{0.36}}  & 67.83   & \textbf{\textcolor{ForestGreen}{0.35}}  & 55.65 & \textbf{\textcolor{ForestGreen}{0.37}}  & 89.74 \\
    \texttt{Arxiv}        & \textbf{\textcolor{ForestGreen}{0.40}}  & 85.23  & \textbf{\textcolor{ForestGreen}{0.40 }} & 132.36  & \textbf{\textcolor{ForestGreen}{0.45}}  & 80.77 & \textbf{\textcolor{ForestGreen}{0.47}}  & 244.95  \\
    \bottomrule
    \end{tabular}
    }
    \label{Table:Inference_speed}
    \vspace{-0.5cm}
\end{wraptable}

\subsection{Computational Cost} \label{sec:computational_cost}
Table~\ref{Table:Inference_speed} shows that \textbf{GOLD generally achieves a very close inference time, and a faster (w/ VAE) or comparable (w/LDM) training time relative to the GNNSafe(++) baseline}. This is under the situation that the non-OOD exposed GOLD outperforms the existing non-OOD exposure methods, while matching or surpassing the real-OOD exposed SOTA baselines, all under the same backbone. In addition to the high-performing LDM variant, a lightweight VAE is also experimented, providing an efficient alternative with comparable performance. Thus, we consider this training cost as an acceptable trade-off for improved OOD detection performance, and is discussed in Appendix~\ref{Appendix:Limitation}. However, we highlight that GOLD can achieve a similar inference time as the baselines, regardless of the LGM, as shown in Table~\ref{Table:Inference_speed}. This reveals a competitive application of GOLD while having a strong performance. We provide detailed results in Appendix~\ref{Appendix:computational_cost}.


\section{Related Work}
Our work intersects with three major research areas: \textbf{1) Non-OOD-Exposure OOD Detection} that purely relies on ID data for detecting OOD instances, this involves score-based methods, feature learning, and techniques specific for graph-structured data~\citep{ConfOOD, Hendrycks17softmax, MSP, GenOE, energy, SGOOD, grasp, GKDE, GOOD-D, GraphDE, GNNSafe, NODESAFE}; \textbf{2) OOD Exposure-Based OOD Detection}, a prominent line of work that adopts auxiliary OOD data to assist training, often achieving higher performance than non-OOD-exposure based methods~\citep{OE, energy, Textual-OODExposure, DivOE, ATOL, SAL,GNNSafe, GDE_OOD}; and \textbf{3) OOD Generation,} a more recent field that aims to synthesise OOD-like data to assist OOD detection~\citep{manifold, Likelihood, LRegret, GenAnalysis, GenUnknown, Hierarc,ConfOOD,VOS,NPOS}. Notably for graph data, \textsc{GNNSafe} considers the inter-dependence nature of node instances and proposes an energy propagation schema, and explores an OOD-exposed variant \textsc{GNNSafe++}~\citep{GNNSafe}. \textsc{NODESafe/++} builds upon \textsc{GNNSafe/++} and proposes additional regularisation terms to reduce and bound the generation of extreme energy scores~\citep{NODESAFE}. \cite{GDE_OOD} proposes a generalised Dirichlet energy score for graph OOD detection. A detailed review of related work is provided in Appendix~\ref{Appendix:related_work}.

\section{Conclusion} \label{sec:conclusion}
In this paper, we propose GOLD, a novel graph OOD detection framework with a latent generative model trained in a novel implicit adversarial paradigm. Unlike methods that rely on pre-trained generative models or real OOD data requiring auxiliary data inputs, GOLD synthesises pseudo-OOD data to inherit OOD characteristics through the implicit adversarial framework, solely based on ID data. An effective OOD detector head is further designed to address the difficulties with multiple classes in the logit space, optimising the energy score for improved detection. 
Extensive experiments show the efficacy of GOLD, outperforming SOTA non- and OOD-exposed methods.
We hope this work inspires future synthetic-based graph OOD detection research for real-world applications.

\newpage
\section{Acknowledgement}
This research has been partially supported by Australian Research Council Discovery Projects (DP230101196, DP24010306, DE250100919 and CE200100025).

\section{Reproducibility Statement}
To support reproducible research, we summarise our efforts as below:
\begin{enumerate}
    \item \textbf{Baselines \& Datasets.} We follow the baseline from~\citep{GNNSafe} and utilise publicly available datasets. The details are described in Section \ref{sec:Experiments} and Appendix \ref{Appendix:dataset}.
    \item \textbf{Model training.} Our implementation of the energy-based OOD detector builds upon the open-sourced work \textsc{GNNSafe} by \cite{GNNSafe}, \url{https://github.com/qitianwu/GraphOOD-GNNSafe}. Detailed implementation setting is provided in Section \ref{sec:Experiments} and Appendix \ref{Appendix:implementation_details}.
    \item \textbf{Methodology.} Our GOLD framework is fully documented in Section \ref{sec:GOLD_Method}. In addition, we provide a detailed pseudo code in Algorithm \ref{alg:GOLD}.
    \item \textbf{Evaluation Metrics.} We discuss the evaluation metrics used in Section \ref{sec:Experiments} and Appendix \ref{Appendix:Metrics}.
\end{enumerate}

\newpage
\bibliography{iclr2025_conference}

\begin{thebibliography}{94}
\providecommand{\natexlab}[1]{#1}
\providecommand{\url}[1]{\texttt{#1}}
\expandafter\ifx\csname urlstyle\endcsname\relax
  \providecommand{\doi}[1]{doi: #1}\else
  \providecommand{\doi}{doi: \begingroup \urlstyle{rm}\Url}\fi

\bibitem[Abu{-}El{-}Haija et~al.(2019)Abu{-}El{-}Haija, Perozzi, Kapoor, Alipourfard, Lerman, Harutyunyan, Steeg, and Galstyan]{MixHop}
Sami Abu{-}El{-}Haija, Bryan Perozzi, Amol Kapoor, Nazanin Alipourfard, Kristina Lerman, Hrayr Harutyunyan, Greg~Ver Steeg, and Aram Galstyan.
\newblock Mixhop: Higher-order graph convolutional architectures via sparsified neighborhood mixing.
\newblock In \emph{ICML}, 2019.

\bibitem[Ahmedt{-}Aristizabal et~al.(2021)Ahmedt{-}Aristizabal, Armin, Denman, Fookes, and Petersson]{GraphMedicalDiagnosis}
David Ahmedt{-}Aristizabal, Mohammad~Ali Armin, Simon Denman, Clinton Fookes, and Lars Petersson.
\newblock Graph-based deep learning for medical diagnosis and analysis: Past, present and future.
\newblock \emph{Sensors}, 21\penalty0 (14):\penalty0 4758, 2021.
\newblock \doi{10.3390/S21144758}.
\newblock URL \url{https://doi.org/10.3390/s21144758}.

\bibitem[Bao et~al.(2024)Bao, Wu, Jiang, Chen, Sun, and Yan]{GDE_OOD}
Tianyi Bao, Qitian Wu, Zetian Jiang, Yiting Chen, Jiawei Sun, and Junchi Yan.
\newblock Graph out-of-distribution detection goes neighborhood shaping.
\newblock In \emph{ICML}, 2024.

\bibitem[Bazhenov et~al.(2022)Bazhenov, Ivanov, Panov, Zaytsev, and Burnaev]{GOODD-uncertainty}
Gleb Bazhenov, Sergei Ivanov, Maxim Panov, Alexey Zaytsev, and Evgeny Burnaev.
\newblock Towards {OOD} detection in graph classification from uncertainty estimation perspective.
\newblock \emph{CoRR}, 2022.

\bibitem[Bitterwolf et~al.(2020)Bitterwolf, Meinke, and Hein]{GOOD-cert}
Julian Bitterwolf, Alexander Meinke, and Matthias Hein.
\newblock Certifiably adversarially robust detection of out-of distribution data.
\newblock In \emph{NeurIPS}, 2020.

\bibitem[Cao et~al.(2020)Cao, Huang, Hui, and Cohen]{medicalDiagnoisis}
Tianshi Cao, Chinwei Huang, David~Yu{-}Tung Hui, and Joseph~Paul Cohen.
\newblock A benchmark of medical out of distribution detection.
\newblock \emph{CoRR}, abs/2007.04250, 2020.
\newblock URL \url{https://arxiv.org/abs/2007.04250}.

\bibitem[Chen et~al.(2021)Chen, Li, Wu, Liang, and Jha]{ATOM}
Jiefeng Chen, Yixuan Li, Xi~Wu, Yingyu Liang, and Somesh Jha.
\newblock {ATOM:} robustifying out-of-distribution detection using outlier mining.
\newblock In \emph{ECML PKDD}, 2021.

\bibitem[Chen et~al.(2023)Chen, Chen, Maul, Li, and Yin]{FocalOE}
Qichao Chen, Zhiyuan Chen, Tom{\'{a}}s Maul, Kuan Li, and Jianping Yin.
\newblock Outlier exposure with focal loss for out-of-distribution detection.
\newblock In \emph{ACAI}, 2023.

\bibitem[Chen et~al.(2022)Chen, Zhang, Bian, Yang, Ma, Xie, Liu, Han, and Cheng]{CIGA}
Yongqiang Chen, Yonggang Zhang, Yatao Bian, Han Yang, Kaili Ma, Binghui Xie, Tongliang Liu, Bo~Han, and James Cheng.
\newblock Learning causally invariant representations for out-of-distribution generalization on graphs.
\newblock In \emph{NeurIPS}, 2022.

\bibitem[Choi \& Chung(2020)Choi and Chung]{RND}
Sung{-}Ik Choi and Sae{-}Young Chung.
\newblock Novelty detection via blurring.
\newblock In \emph{ICLR}, 2020.

\bibitem[Dai et~al.(2017)Dai, Yang, Yang, Cohen, and Salakhutdinov]{BadGAN}
Zihang Dai, Zhilin Yang, Fan Yang, William~W. Cohen, and Ruslan Salakhutdinov.
\newblock Good semi-supervised learning that requires a bad {GAN}.
\newblock In \emph{NeurIPS}, 2017.

\bibitem[Ding et~al.(2021)Ding, Kong, Chen, Kirchenbauer, Goldblum, Wipf, Huang, and Goldstein]{GDSBenchmark}
Mucong Ding, Kezhi Kong, Jiuhai Chen, John Kirchenbauer, Micah Goldblum, David Wipf, Furong Huang, and Tom Goldstein.
\newblock A closer look at distribution shifts and out-of-distribution generalization on graphs.
\newblock In \emph{NeurIPS Workshop}, 2021.

\bibitem[Ding \& Shi(2023)Ding and Shi]{SGOOD}
Zhihao Ding and Jieming Shi.
\newblock {SGOOD:} substructure-enhanced graph-level out-of-distribution detection.
\newblock \emph{CoRR}, 2023.

\bibitem[Djurisic et~al.(2023)Djurisic, Bozanic, Ashok, and Liu]{ASH}
Andrija Djurisic, Nebojsa Bozanic, Arjun Ashok, and Rosanne Liu.
\newblock Extremely simple activation shaping for out-of-distribution detection.
\newblock In \emph{ICLR}, 2023.

\bibitem[Dong et~al.(2022)Dong, Guo, Li, Ting, Liu, and Kung]{NMD}
Xin Dong, Junfeng Guo, Ang Li, Wei{-}Te Ting, Cong Liu, and H.~T. Kung.
\newblock Neural mean discrepancy for efficient out-of-distribution detection.
\newblock In \emph{CVPR}, 2022.

\bibitem[Du et~al.(2022)Du, Wang, Cai, and Li]{VOS}
Xuefeng Du, Zhaoning Wang, Mu~Cai, and Yixuan Li.
\newblock {VOS:} learning what you don't know by virtual outlier synthesis.
\newblock In \emph{ICLR}, 2022.

\bibitem[Du et~al.(2023)Du, Sun, Zhu, and Li]{Dream-OOD}
Xuefeng Du, Yiyou Sun, Jerry Zhu, and Yixuan Li.
\newblock Dream the impossible: Outlier imagination with diffusion models.
\newblock In \emph{NeurIPS}, 2023.

\bibitem[Du et~al.(2024)Du, Fang, Diakonikolas, and Li]{SAL}
Xuefeng Du, Zhen Fang, Ilias Diakonikolas, and Yixuan Li.
\newblock How does unlabeled data provably help out-of-distribution detection?
\newblock In \emph{ICLR}, 2024.

\bibitem[Evdaimon et~al.(2024)Evdaimon, Nikolentzos, Chatzianastasis, Abdine, and Vazirgiannis]{NGG}
Iakovos Evdaimon, Giannis Nikolentzos, Michail Chatzianastasis, Hadi Abdine, and Michalis Vazirgiannis.
\newblock Neural graph generator: Feature-conditioned graph generation using latent diffusion models.
\newblock \emph{CoRR}, 2024.

\bibitem[Giuffr{\`{e}} \& Shung(2023)Giuffr{\`{e}} and Shung]{HealthOOD}
Mauro Giuffr{\`{e}} and Dennis~L. Shung.
\newblock Harnessing the power of synthetic data in healthcare: innovation, application, and privacy.
\newblock \emph{npj Digit. Medicine}, 2023.

\bibitem[Grathwohl et~al.(2020)Grathwohl, Wang, Jacobsen, Duvenaud, Norouzi, and Swersky]{EBM-NN}
Will Grathwohl, Kuan{-}Chieh Wang, J{\"{o}}rn{-}Henrik Jacobsen, David Duvenaud, Mohammad Norouzi, and Kevin Swersky.
\newblock Your classifier is secretly an energy based model and you should treat it like one.
\newblock In \emph{ICLR}, 2020.

\bibitem[Gui et~al.(2022)Gui, Li, Wang, and Ji]{GOOD}
Shurui Gui, Xiner Li, Limei Wang, and Shuiwang Ji.
\newblock {GOOD:} {A} graph out-of-distribution benchmark.
\newblock In \emph{NeurIPS}, 2022.

\bibitem[Guo et~al.(2023)Guo, Yang, Chen, Liu, Shi, and Du]{AAGOD}
Yuxin Guo, Cheng Yang, Yuluo Chen, Jixi Liu, Chuan Shi, and Junping Du.
\newblock A data-centric framework to endow graph neural networks with out-of-distribution detection ability.
\newblock In \emph{KDD}, 2023.

\bibitem[Hamilton et~al.(2017)Hamilton, Ying, and Leskovec]{GraphSage}
William~L. Hamilton, Zhitao Ying, and Jure Leskovec.
\newblock Inductive representation learning on large graphs.
\newblock In \emph{NeurIPS}, 2017.

\bibitem[Hein et~al.(2019)Hein, Andriushchenko, and Bitterwolf]{WhyReLU}
Matthias Hein, Maksym Andriushchenko, and Julian Bitterwolf.
\newblock Why relu networks yield high-confidence predictions far away from the training data and how to mitigate the problem.
\newblock In \emph{CVPR}, 2019.

\bibitem[Hendrycks \& Gimpel(2017{\natexlab{a}})Hendrycks and Gimpel]{Hendrycks17softmax}
Dan Hendrycks and Kevin Gimpel.
\newblock A baseline for detecting misclassified and out-of-distribution examples in neural networks.
\newblock In \emph{ICLR}, 2017{\natexlab{a}}.

\bibitem[Hendrycks \& Gimpel(2017{\natexlab{b}})Hendrycks and Gimpel]{MSP}
Dan Hendrycks and Kevin Gimpel.
\newblock A baseline for detecting misclassified and out-of-distribution examples in neural networks.
\newblock In \emph{ICLR}, 2017{\natexlab{b}}.

\bibitem[Hendrycks et~al.(2019)Hendrycks, Mazeika, and Dietterich]{OE}
Dan Hendrycks, Mantas Mazeika, and Thomas~G. Dietterich.
\newblock Deep anomaly detection with outlier exposure.
\newblock In \emph{ICLR}, 2019.

\bibitem[Ho et~al.(2020)Ho, Jain, and Abbeel]{DDPM}
Jonathan Ho, Ajay Jain, and Pieter Abbeel.
\newblock Denoising diffusion probabilistic models.
\newblock In \emph{NeurIPS}, 2020.

\bibitem[Hsu et~al.(2020)Hsu, Shen, Jin, and Kira]{Gen-ODIN}
Yen{-}Chang Hsu, Yilin Shen, Hongxia Jin, and Zsolt Kira.
\newblock Generalized {ODIN:} detecting out-of-distribution image without learning from out-of-distribution data.
\newblock In \emph{CVPR}, 2020.

\bibitem[Hu et~al.(2020)Hu, Fey, Zitnik, Dong, Ren, Liu, Catasta, and Leskovec]{Arxiv}
Weihua Hu, Matthias Fey, Marinka Zitnik, Yuxiao Dong, Hongyu Ren, Bowen Liu, Michele Catasta, and Jure Leskovec.
\newblock Open graph benchmark: Datasets for machine learning on graphs.
\newblock In \emph{NeurIPS}, 2020.

\bibitem[Huang et~al.(2022{\natexlab{a}})Huang, Wang, Fang, and Chen]{LMN}
Tiancheng Huang, Donglin Wang, Yuan Fang, and Zhengyu Chen.
\newblock End-to-end open-set semi-supervised node classification with out-of-distribution detection.
\newblock In \emph{IJCAI}, 2022{\natexlab{a}}.

\bibitem[Huang et~al.(2022{\natexlab{b}})Huang, Chen, Fang, Menkovski, Zhao, Yin, Pei, Mocanu, Wang, Pechenizkiy, and Liu]{UGTs}
Tianjin Huang, Tianlong Chen, Meng Fang, Vlado Menkovski, Jiaxu Zhao, Lu~Yin, Yulong Pei, Decebal~Constantin Mocanu, Zhangyang Wang, Mykola Pechenizkiy, and Shiwei Liu.
\newblock You can have better graph neural networks by not training weights at all: Finding untrained gnns tickets.
\newblock In \emph{LoG}, 2022{\natexlab{b}}.

\bibitem[Huang et~al.(2022{\natexlab{c}})Huang, Wang, Xia, Wang, and Zhang]{regOOD}
Wenjian Huang, Hao Wang, Jiahao Xia, Chengyan Wang, and Jianguo Zhang.
\newblock Density-driven regularization for out-of-distribution detection.
\newblock In \emph{NeurIPS}, 2022{\natexlab{c}}.

\bibitem[Ji et~al.(2022)Ji, Zhang, Wu, Wu, Huang, Xu, Rong, Li, Ren, Xue, Lai, Xu, Feng, Liu, Luo, Zhou, Huang, Zhao, and Bian]{DrugOOD}
Yuanfeng Ji, Lu~Zhang, Jiaxiang Wu, Bingzhe Wu, Long{-}Kai Huang, Tingyang Xu, Yu~Rong, Lanqing Li, Jie Ren, Ding Xue, Houtim Lai, Shaoyong Xu, Jing Feng, Wei Liu, Ping Luo, Shuigeng Zhou, Junzhou Huang, Peilin Zhao, and Yatao Bian.
\newblock Drugood: Out-of-distribution {(OOD)} dataset curator and benchmark for ai-aided drug discovery - {A} focus on affinity prediction problems with noise annotations.
\newblock \emph{CoRR}, 2022.

\bibitem[Jiang et~al.(2023)Jiang, Liu, Fang, Chen, Liu, Zheng, and Han]{RPRW}
Xue Jiang, Feng Liu, Zhen Fang, Hong Chen, Tongliang Liu, Feng Zheng, and Bo~Han.
\newblock Detecting out-of-distribution data through in-distribution class prior.
\newblock In \emph{ICML}, 2023.

\bibitem[Kingma \& Ba(2015)Kingma and Ba]{Adam}
Diederik~P. Kingma and Jimmy Ba.
\newblock Adam: {A} method for stochastic optimization.
\newblock In \emph{ICLR}, 2015.

\bibitem[Kingma \& Welling(2014)Kingma and Welling]{VAE}
Diederik~P. Kingma and Max Welling.
\newblock Auto-encoding variational bayes.
\newblock In \emph{ICLR}, 2014.

\bibitem[Kipf \& Welling(2017)Kipf and Welling]{GCN}
Thomas~N. Kipf and Max Welling.
\newblock Semi-supervised classification with graph convolutional networks.
\newblock In \emph{ICLR}, 2017.

\bibitem[Koo et~al.(2024)Koo, Choi, and Hwang]{GenOE}
Jiin Koo, Sungjoon Choi, and Sangheum Hwang.
\newblock Generalized outlier exposure: Towards a trustworthy out-of-distribution detector without sacrificing accuracy.
\newblock \emph{Neurocomputing}, 2024.

\bibitem[Lafon et~al.(2023)Lafon, Ramzi, Rambour, and Thome]{EBFeat}
Marc Lafon, Elias Ramzi, Cl{\'{e}}ment Rambour, and Nicolas Thome.
\newblock Hybrid energy based model in the feature space for out-of-distribution detection.
\newblock In \emph{ICML}, 2023.

\bibitem[Lang et~al.(2023)Lang, Zheng, Li, Sun, Huang, and Li]{NLPOODD}
Hao Lang, Yinhe Zheng, Yixuan Li, Jian Sun, Fei Huang, and Yongbin Li.
\newblock A survey on out-of-distribution detection in {NLP}.
\newblock \emph{CoRR}, 2023.

\bibitem[Lee et~al.(2018{\natexlab{a}})Lee, Lee, Lee, and Shin]{ConfOOD}
Kimin Lee, Honglak Lee, Kibok Lee, and Jinwoo Shin.
\newblock Training confidence-calibrated classifiers for detecting out-of-distribution samples.
\newblock In \emph{ICLR}, 2018{\natexlab{a}}.

\bibitem[Lee et~al.(2018{\natexlab{b}})Lee, Lee, Lee, and Shin]{Mahalanobis}
Kimin Lee, Kibok Lee, Honglak Lee, and Jinwoo Shin.
\newblock A simple unified framework for detecting out-of-distribution samples and adversarial attacks.
\newblock In \emph{NeurIPS}, 2018{\natexlab{b}}.

\bibitem[Lee et~al.(2023{\natexlab{a}})Lee, Jo, and Hwang]{MOOD-Molecule}
Seul Lee, Jaehyeong Jo, and Sung~Ju Hwang.
\newblock Exploring chemical space with score-based out-of-distribution generation.
\newblock In \emph{ICML}, 2023{\natexlab{a}}.

\bibitem[Lee et~al.(2023{\natexlab{b}})Lee, Yin, and Zhang]{HealthForecast}
Seungyeon Lee, Changchang Yin, and Ping Zhang.
\newblock Stable clinical risk prediction against distribution shift in electronic health records.
\newblock \emph{Patterns}, 2023{\natexlab{b}}.

\bibitem[Li et~al.(2022{\natexlab{a}})Li, Wang, Zhang, and Zhu]{GOODSurvey}
Haoyang Li, Xin Wang, Ziwei Zhang, and Wenwu Zhu.
\newblock Out-of-distribution generalization on graphs: {A} survey.
\newblock \emph{CoRR}, 2022{\natexlab{a}}.

\bibitem[Li et~al.(2023)Li, Wang, Zhang, and Zhu]{OOD-GNN}
Haoyang Li, Xin Wang, Ziwei Zhang, and Wenwu Zhu.
\newblock {OOD-GNN:} out-of-distribution generalized graph neural network.
\newblock \emph{{IEEE} Trans. Knowl. Data Eng.}, 2023.

\bibitem[Li et~al.(2024)Li, Chen, Liu, Wang, He, Cheng, and Ao]{advOOD}
Kuan Li, YiWen Chen, Yang Liu, Jin Wang, Qing He, Minhao Cheng, and Xiang Ao.
\newblock Boosting the adversarial robustness of graph neural networks: An ood perspective.
\newblock In \emph{ICLR}, 2024.

\bibitem[Li et~al.(2022{\natexlab{b}})Li, Wu, Nie, and Yan]{GraphDE}
Zenan Li, Qitian Wu, Fan Nie, and Junchi Yan.
\newblock Graphde: {A} generative framework for debiased learning and out-of-distribution detection on graphs.
\newblock In \emph{NeurIPS}, 2022{\natexlab{b}}.

\bibitem[Liang et~al.(2018)Liang, Li, and Srikant]{ODIN}
Shiyu Liang, Yixuan Li, and R.~Srikant.
\newblock Enhancing the reliability of out-of-distribution image detection in neural networks.
\newblock In \emph{ICLR}, 2018.

\bibitem[Lin et~al.(2021)Lin, Roy, and Li]{MOOD}
Ziqian Lin, Sreya~Dutta Roy, and Yixuan Li.
\newblock {MOOD:} multi-level out-of-distribution detection.
\newblock In \emph{CVPR}, 2021.

\bibitem[Liu et~al.(2023{\natexlab{a}})Liu, Yang, Lu, Chen, Li, Zhang, Bai, Fang, Sun, Yu, and Shi]{GFM}
Jiawei Liu, Cheng Yang, Zhiyuan Lu, Junze Chen, Yibo Li, Mengmei Zhang, Ting Bai, Yuan Fang, Lichao Sun, Philip~S. Yu, and Chuan Shi.
\newblock Towards graph foundation models: {A} survey and beyond.
\newblock \emph{CoRR}, 2023{\natexlab{a}}.

\bibitem[Liu et~al.(2020)Liu, Wang, Owens, and Li]{energy}
Weitang Liu, Xiaoyun Wang, John~D. Owens, and Yixuan Li.
\newblock Energy-based out-of-distribution detection.
\newblock In \emph{NeurIPS}, 2020.

\bibitem[Liu et~al.(2023{\natexlab{b}})Liu, Ding, Liu, and Pan]{GOOD-D}
Yixin Liu, Kaize Ding, Huan Liu, and Shirui Pan.
\newblock {GOOD-D:} on unsupervised graph out-of-distribution detection.
\newblock In \emph{WSDM}, 2023{\natexlab{b}}.

\bibitem[Ma et~al.(2023)Ma, Sun, Ding, and Wu]{grasp}
Longfei Ma, Yiyou Sun, Kaize Ding, and Fei Wu.
\newblock Score propagation as a catalyst for graph out-of-distribution detection: A theoretical and empirical study.
\newblock In \emph{ICLR}, 2023.

\bibitem[McAuley et~al.(2015)McAuley, Targett, Shi, and van~den Hengel]{AmazonPhoto}
Julian~J. McAuley, Christopher Targett, Qinfeng Shi, and Anton van~den Hengel.
\newblock Image-based recommendations on styles and substitutes.
\newblock In \emph{SIGIR}, 2015.

\bibitem[Nalisnick et~al.(2019)Nalisnick, Matsukawa, Teh, G{\"{o}}r{\"{u}}r, and Lakshminarayanan]{GenUnknown}
Eric~T. Nalisnick, Akihiro Matsukawa, Yee~Whye Teh, Dilan G{\"{o}}r{\"{u}}r, and Balaji Lakshminarayanan.
\newblock Do deep generative models know what they don't know?
\newblock In \emph{ICLR}, 2019.

\bibitem[Papadopoulos et~al.(2021)Papadopoulos, Rajati, Shaikh, and Wang]{OECC}
Aristotelis{-}Angelos Papadopoulos, Mohammad~Reza Rajati, Nazim Shaikh, and Jiamian Wang.
\newblock Outlier exposure with confidence control for out-of-distribution detection.
\newblock \emph{Neurocomputing}, 2021.

\bibitem[Park et~al.(2023)Park, Mok, Jung, Lee, and Yoon]{Textual-OODExposure}
Sangha Park, Jisoo Mok, Dahuin Jung, Saehyung Lee, and Sungroh Yoon.
\newblock On the powerfulness of textual outlier exposure for visual ood detection.
\newblock In \emph{NeurIPS}, 2023.

\bibitem[Ren et~al.(2019)Ren, Liu, Fertig, Snoek, Poplin, DePristo, Dillon, and Lakshminarayanan]{LRatio_OODD}
Jie Ren, Peter~J. Liu, Emily Fertig, Jasper Snoek, Ryan Poplin, Mark~A. DePristo, Joshua~V. Dillon, and Balaji Lakshminarayanan.
\newblock Likelihood ratios for out-of-distribution detection.
\newblock In \emph{NeurIPS}, 2019.

\bibitem[Rombach et~al.(2022)Rombach, Blattmann, Lorenz, Esser, and Ommer]{sd}
Robin Rombach, Andreas Blattmann, Dominik Lorenz, Patrick Esser, and Bj{\"{o}}rn Ommer.
\newblock High-resolution image synthesis with latent diffusion models.
\newblock In \emph{CVPR}, 2022.

\bibitem[Rozemberczki \& Sarkar(2021)Rozemberczki and Sarkar]{Twitch_Gamers}
Benedek Rozemberczki and Rik Sarkar.
\newblock Twitch gamers: a dataset for evaluating proximity preserving and structural role-based node embeddings.
\newblock \emph{CoRR}, 2021.

\bibitem[Schirrmeister et~al.(2020)Schirrmeister, Zhou, Ball, and Zhang]{Hierarc}
Robin Schirrmeister, Yuxuan Zhou, Tonio Ball, and Dan Zhang.
\newblock Understanding anomaly detection with deep invertible networks through hierarchies of distributions and features.
\newblock In \emph{NeurIPS}, 2020.

\bibitem[Sen et~al.(2008)Sen, Namata, Bilgic, Getoor, Gallagher, and Eliassi{-}Rad]{Cora}
Prithviraj Sen, Galileo Namata, Mustafa Bilgic, Lise Getoor, Brian Gallagher, and Tina Eliassi{-}Rad.
\newblock Collective classification in network data.
\newblock \emph{{AI} Mag.}, 2008.

\bibitem[Serr{\`{a}} et~al.(2020)Serr{\`{a}}, {\'{A}}lvarez, G{\'{o}}mez, Slizovskaia, N{\'{u}}{\~{n}}ez, and Luque]{Likelihood}
Joan Serr{\`{a}}, David {\'{A}}lvarez, Vicen{\c{c}} G{\'{o}}mez, Olga Slizovskaia, Jos{\'{e}}~F. N{\'{u}}{\~{n}}ez, and Jordi Luque.
\newblock Input complexity and out-of-distribution detection with likelihood-based generative models.
\newblock In \emph{ICLR}, 2020.

\bibitem[Shen et~al.(2024)Shen, Wang, Zhou, Pan, and Wang]{MOL_DIF}
Xu~Shen, Yili Wang, Kaixiong Zhou, Shirui Pan, and Xin Wang.
\newblock Optimizing ood detection in molecular graphs: A novel approach with diffusion models.
\newblock \emph{CoRR}, 2024.

\bibitem[Song \& Wang(2022)Song and Wang]{OODGAT}
Yu~Song and Donglin Wang.
\newblock Learning on graphs with out-of-distribution nodes.
\newblock In \emph{KDD}, 2022.

\bibitem[Stadler et~al.(2021)Stadler, Charpentier, Geisler, Z{\"{u}}gner, and G{\"{u}}nnemann]{GPN}
Maximilian Stadler, Bertrand Charpentier, Simon Geisler, Daniel Z{\"{u}}gner, and Stephan G{\"{u}}nnemann.
\newblock Graph posterior network: Bayesian predictive uncertainty for node classification.
\newblock In \emph{NeurIPS}, 2021.

\bibitem[Sun \& Li(2022)Sun and Li]{DICE}
Yiyou Sun and Yixuan Li.
\newblock {DICE:} leveraging sparsification for out-of-distribution detection.
\newblock In \emph{ECCV}, 2022.

\bibitem[Sun et~al.(2021)Sun, Guo, and Li]{React}
Yiyou Sun, Chuan Guo, and Yixuan Li.
\newblock React: Out-of-distribution detection with rectified activations.
\newblock In \emph{NeurIPS}, 2021.

\bibitem[Tao et~al.(2023)Tao, Du, Zhu, and Li]{NPOS}
Leitian Tao, Xuefeng Du, Jerry Zhu, and Yixuan Li.
\newblock Non-parametric outlier synthesis.
\newblock In \emph{ICLR}, 2023.

\bibitem[Trivedi et~al.(2024)Trivedi, Heimann, Anirudh, Koutra, and Thiagarajan]{G_UQ}
Puja Trivedi, Mark Heimann, Rushil Anirudh, Danai Koutra, and Jayaraman~J. Thiagarajan.
\newblock Accurate and scalable estimation of epistemic uncertainty for graph neural networks.
\newblock \emph{ICLR}, 2024.

\bibitem[Velickovic et~al.(2018)Velickovic, Cucurull, Casanova, Romero, Li{\`{o}}, and Bengio]{GAT}
Petar Velickovic, Guillem Cucurull, Arantxa Casanova, Adriana Romero, Pietro Li{\`{o}}, and Yoshua Bengio.
\newblock Graph attention networks.
\newblock In \emph{ICLR}, 2018.

\bibitem[Vernekar et~al.(2019)Vernekar, Gaurav, Abdelzad, Denouden, Salay, and Czarnecki]{manifold}
Sachin Vernekar, Ashish Gaurav, Vahdat Abdelzad, Taylor Denouden, Rick Salay, and Krzysztof Czarnecki.
\newblock Out-of-distribution detection in classifiers via generation.
\newblock \emph{CoRR}, 2019.

\bibitem[Vyas et~al.(2018)Vyas, Jammalamadaka, Zhu, Das, Kaul, and Willke]{Ensemble}
Apoorv Vyas, Nataraj Jammalamadaka, Xia Zhu, Dipankar Das, Bharat Kaul, and Theodore~L. Willke.
\newblock Out-of-distribution detection using an ensemble of self supervised leave-out classifiers.
\newblock In \emph{ECCV}, 2018.

\bibitem[Wang \& Li(2023)Wang and Li]{SLW}
Han Wang and Yixuan Li.
\newblock A graph-theoretic framework for joint ood generalization and detection.
\newblock In \emph{CoRR}, 2023.

\bibitem[Wang et~al.(2021)Wang, Liu, Bocchieri, and Li]{Wang21energy}
Haoran Wang, Weitang Liu, Alex Bocchieri, and Yixuan Li.
\newblock Can multi-label classification networks know what they don't know?
\newblock In \emph{NeurIPS}, 2021.

\bibitem[Wang et~al.(2024)Wang, He, Zhang, Liu, Wang, Pan, Jin, and Chua]{GOODAT}
Luzhi Wang, Dongxiao He, He~Zhang, Yixin Liu, Wenjie Wang, Shirui Pan, Di~Jin, and Tat{-}Seng Chua.
\newblock {GOODAT:} towards test-time graph out-of-distribution detection.
\newblock In \emph{AAAI}, 2024.

\bibitem[Wang et~al.(2020)Wang, Dai, Wipf, and Zhu]{GenAnalysis}
Ziyu Wang, Bin Dai, David~P. Wipf, and Jun Zhu.
\newblock Further analysis of outlier detection with deep generative models.
\newblock In \emph{NeurIPS}, 2020.

\bibitem[Wu et~al.(2023{\natexlab{a}})Wu, Chen, and Deng]{DFDD}
Aming Wu, Da~Chen, and Cheng Deng.
\newblock Deep feature deblurring diffusion for detecting out-of-distribution objects.
\newblock In \emph{ICCV}, 2023{\natexlab{a}}.

\bibitem[Wu et~al.(2023{\natexlab{b}})Wu, Chen, Yang, and Yan]{GNNSafe}
Qitian Wu, Yiting Chen, Chenxiao Yang, and Junchi Yan.
\newblock Energy-based out-of-distribution detection for graph neural networks.
\newblock In \emph{ICLR}, 2023{\natexlab{b}}.

\bibitem[Xiao et~al.(2020)Xiao, Yan, and Amit]{LRegret}
Zhisheng Xiao, Qing Yan, and Yali Amit.
\newblock Likelihood regret: An out-of-distribution detection score for variational auto-encoder.
\newblock In \emph{NeurIPS}, 2020.

\bibitem[Yang et~al.(2021)Yang, Zhou, Li, and Liu]{OODD}
Jingkang Yang, Kaiyang Zhou, Yixuan Li, and Ziwei Liu.
\newblock Generalized out-of-distribution detection: {A} survey.
\newblock \emph{CoRR}, 2021.

\bibitem[Yang et~al.(2023{\natexlab{a}})Yang, Zhou, and Liu]{FS-OOD}
Jingkang Yang, Kaiyang Zhou, and Ziwei Liu.
\newblock Full-spectrum out-of-distribution detection.
\newblock \emph{Int. J. Comput. Vis.}, 2023{\natexlab{a}}.

\bibitem[Yang et~al.(2023{\natexlab{b}})Yang, Lu, and Gan]{EMP}
Lina Yang, Bin Lu, and Xiaoying Gan.
\newblock Graph open-set recognition via entropy message passing.
\newblock In \emph{ICDM}, 2023{\natexlab{b}}.

\bibitem[Yang et~al.(2022)Yang, Zeng, Wu, Jia, and Yan]{MoleOOD}
Nianzu Yang, Kaipeng Zeng, Qitian Wu, Xiaosong Jia, and Junchi Yan.
\newblock Learning substructure invariance for out-of-distribution molecular representations.
\newblock In \emph{NeurIPS}, 2022.

\bibitem[Yang et~al.(2024)Yang, Liang, Liu, Gui, Yao, and Zhang]{NODESAFE}
Shenzhi Yang, Bin Liang, An~Liu, Lin Gui, Xingkai Yao, and Xiaofang Zhang.
\newblock Bounded and uniform energy-based out-of-distribution detection for graphs.
\newblock In \emph{ICML}, 2024.

\bibitem[Yu et~al.(2023)Yu, Liang, and He]{LiSA}
Junchi Yu, Jian Liang, and Ran He.
\newblock Mind the label shift of augmentation-based graph {OOD} generalization.
\newblock In \emph{CVPR}, 2023.

\bibitem[Zhao et~al.(2020)Zhao, Chen, Hu, and Cho]{GKDE}
Xujiang Zhao, Feng Chen, Shu Hu, and Jin{-}Hee Cho.
\newblock Uncertainty aware semi-supervised learning on graph data.
\newblock In \emph{NeurIPS}, 2020.

\bibitem[Zheng et~al.(2023)Zheng, Wang, Fang, Xia, Liu, Liu, and Han]{ATOL}
Haotian Zheng, Qizhou Wang, Zhen Fang, Xiaobo Xia, Feng Liu, Tongliang Liu, and Bo~Han.
\newblock Out-of-distribution detection learning with unreliable out-of-distribution sources.
\newblock In \emph{NeurIPS}, 2023.

\bibitem[Zhou et~al.(2024)Zhou, Wang, and Zhang]{LatentDIF}
Cai Zhou, Xiyuan Wang, and Muhan Zhang.
\newblock Latent graph diffusion: {A} unified framework for generation and prediction on graphs.
\newblock \emph{CoRR}, 2024.

\bibitem[Zhou et~al.(2022)Zhou, Kutyniok, and Ribeiro]{OODLink}
Yangze Zhou, Gitta Kutyniok, and Bruno Ribeiro.
\newblock {OOD} link prediction generalization capabilities of message-passing gnns in larger test graphs.
\newblock In \emph{NeurIPS}, 2022.

\bibitem[Zhu et~al.(2023)Zhu, Geng, Yao, Liu, Niu, Sugiyama, and Han]{DivOE}
Jianing Zhu, Yu~Geng, Jiangchao Yao, Tongliang Liu, Gang Niu, Masashi Sugiyama, and Bo~Han.
\newblock Diversified outlier exposure for out-of-distribution detection via informative extrapolation.
\newblock In \emph{NeurIPS}, 2023.

\end{thebibliography}
\bibliographystyle{iclr2025_conference}

\newpage
\appendix
\section{Appendix}

In the Appendix, we provide additional supplementary material to the main paper. The structure is as follows:
\begin{itemize}
    \item We provide the proof for Proposition 1. in \ref{Appendix:Proof}.
    \item An extended related work is detailed in \ref{Appendix:related_work}.
    \item Potential Limitations is discussed in \ref{Appendix:Limitation}.
    \item Preliminary GNN description is described in \ref{Appendix:GNN}.
    \item We provide the description of datasets in \ref{Appendix:dataset}.
    \item The evaluation metrics and implementation details are provided in \ref{Appendix:Metrics} and \ref{Appendix:implementation_details}.
    \item Additional experiment results, including extended subset performance, ablation study visualisations, empirical evaluations of logits vs. softmax scores, and computational cost are detailed in \ref{Appendix:Additional_exp}, 
    \ref{Appendix:ablation_vis}
    \ref{Appendix:Logits vs. SFM}, \ref{Appendix:computational_cost}.
    \item Descriptions of the latent generative models: 1) Latent diffusion model, and 2) Variational autoencoder are provided in \ref{Appendix:latent_generative_model}.
    
\end{itemize}

\subsection{Proof for Proposition 1.} \label{Appendix:Proof}
\begin{proof}
Let $l_{\theta_{[y]}}$ denote the logits of the MLP detector with parameter $\theta$ for class y, $\phi$ denote the softmax function. Assume the hyper-parameters $\lambda = \gamma = 1$.

Note that:
\begin{equation}
\begin{split}
\frac{\partial \log (e^a_\theta + e^b_\theta)}{\partial \theta} = \frac{e^a_\theta \frac{\partial a_\theta}{\partial \theta} + e^b_\theta \frac{\partial b_\theta}{\partial \theta}}{e^a_\theta + e^b_\theta}\\
\end{split}
\end{equation}

The gradient of $\min_{\text{MLP}} -(\mathcal{L}_{\text{Unc}} + \mathcal{L}_{\text{DReg}})$ w.r.t $\theta$ is given by:  
\begin{equation}
\begin{split}
- \frac{\partial \mathcal{L}_{\text{Unc}}}{\partial \theta} & = - \mathbb{E}_{i\sim P_\text{ID}} \frac{\partial \log[\phi(l_\theta(e_i))_{[0]}]}{\partial \theta} - \mathbb{E}_{j\sim P_\text{p-OOD}} \frac{\partial \log[\phi(l_\theta(e_j))_{[1]}]}{\partial \theta} \\
& = - \mathbb{E}_{i\sim P_\text{ID}} \frac{\partial \log[\frac{e^{l_\theta(e_i)_{[0]}}}{e^{l_\theta(e_i)_{[0]}} + e^{l_\theta(e_i)_{[1]}}}]}{\partial \theta} - \mathbb{E}_{j\sim P_\text{p-OOD}} \frac{\partial \log[\frac{e^{l_\theta(e_j)_{[1]}}}{e^{l_\theta(e_j)_{[0]}} + e^{l_\theta(e_j)_{[1]}}}]}{\partial \theta} \\
& = - \mathbb{E}_{i\sim P_\text{ID}} \frac{\partial [l_\theta(e_i)_{[0]} -\log (e^{l_\theta(e_i)_{[0]}} + e^{l_\theta(e_i)_{[1]}})]}{\partial \theta} \\
& - \mathbb{E}_{j\sim P_\text{p-OOD}} \frac{[\partial l_\theta(e_j)_{[1]} - \log (e^{l_\theta(e_j)_{[0]}} + e^{l_\theta(e_j)_{[1]}})]}{\partial \theta} \\
& = \mathbb{E}_{i\sim P_\text{ID}} \left[ -\frac{\partial l_\theta(e_i)_{[0]}}{\partial \theta} + \frac{e^{\l_\theta(e_i)_{[0]}} \frac{\partial \l_\theta(e_i)_{[0]}}{\partial \theta} + e^{\l_\theta(e_i)_{[1]}} \frac{\partial \l_\theta(e_i)_{[1]}}{\partial \theta}}{e^{\l_\theta(e_i)_{[0]}} + e^{\l_\theta(e_i)_{[1]}}} \right]\\
& +\mathbb{E}_{j\sim P_\text{p-OOD}} \left[- \frac{\partial l_\theta(e_j)_{[1]}}{\partial \theta} +\frac{e^{\l_\theta(e_j)_{[0]}} \frac{\partial \l_\theta(e_j)_{[0]}}{\partial \theta} + e^{\l_\theta(e_j)_{[1]}} \frac{\partial \l_\theta(e_j)_{[1]}}{\partial \theta}}{e^{\l_\theta(e_j)_{[0]}} + e^{\l_\theta(e_j)_{[1]}}} \right]
\end{split}
\end{equation}

Notice that $\max(0, e_i - e'_i)$ and $\max(0, e'_j - e_j)$ are positive and monotonic, the optimised $\theta$ that minimises the functions (arg min) would also minimise $\max(0, e_i - e'_i)^2$ and $\max(0, e'_j - e_j)^2$, thus, we consider the gradient of a surrogate function of $\mathcal{L}_{\text{DReg}}$ as $\mathcal{L}_{\text{DReg}_S}$:

\begin{equation}
\begin{split}
- \frac{\partial \mathcal{L}_{\text{DReg}_S}}{\partial \theta} & = -\frac{\partial}{\partial \theta} \mathbb{E}_{i\sim P_\text{ID}}\operatorname{max}\left( 0, e_i - e'_i\right) - \frac{\partial}{\partial \theta} \mathbb{E}_{j\sim P_\text{p-OOD}}\operatorname{max}\left(0, e'_j -e_j\right) \\
& \text{If $e_i - e'_i \le 0$ or $e'_j - e_j \le 0$ the gradient is 0, else:}\\
& = -\frac{\partial}{\partial \theta} \mathbb{E}_{i\sim P_\text{ID}} \left[e_i + \log(e^{\l_\theta(e_i)_{[0]}} + e^{\l_\theta(e_i)_{[1]}}) \right] \\
& - \frac{\partial}{\partial \theta} \mathbb{E}_{j\sim P_\text{p-OOD}} \left[- \log(e^{\l_\theta(e_j)_{[0]}}  + e^{\l_\theta(e_j)_{[1]}}) -e_j \right]\\
& = -\mathbb{E}_{i\sim P_\text{ID}} \frac{\partial \log(e^{\l_\theta(e_i)_{[0]}} + e^{\l_\theta(e_i)_{[1]}})}{\partial \theta} \\
& - \mathbb{E}_{j\sim P_\text{p-OOD}}  \frac{\partial - \log(e^{\l_\theta(e_j)_{[0]}}  + e^{\l_\theta(e_j)_{[1]}})}{\partial \theta} \\
& = -\mathbb{E}_{i\sim P_\text{ID}} \frac{e^{\l_\theta(e_i)_{[0]}} \frac{\partial \l_\theta(e_i)_{[0]}}{\partial \theta} + e^{\l_\theta(e_i)_{[1]}} \frac{\partial \l_\theta(e_i)_{[1]}}{\partial \theta}}{e^{\l_\theta(e_i)_{[0]}} + e^{\l_\theta(e_i)_{[1]}}}\\
& + \mathbb{E}_{j\sim P_\text{p-OOD}}  \frac{e^{\l_\theta(e_j)_{[0]}} \frac{\partial \l_\theta(e_j)_{[0]}}{\partial \theta} + e^{\l_\theta(e_j)_{[1]}} \frac{\partial \l_\theta(e_j)_{[1]}}{\partial \theta}}{e^{\l_\theta(e_j)_{[0]}} + e^{\l_\theta(e_j)_{[1]}}}
\end{split}
\end{equation}

\begin{equation}
\begin{split}
- (\frac{\partial \mathcal{L}_{\text{Unc}}}{\partial \theta} + \frac{\partial \mathcal{L}_{\text{DReg}_S}}{\partial \theta}) & = \mathbb{E}_{i\sim P_\text{ID}} \left[-\frac{\partial l_\theta(e_i)_{[0]}}{\partial \theta} + \frac{e^{\l_\theta(e_i)_{[0]}} \frac{\partial \l_\theta(e_i)_{[0]}}{\partial \theta} + e^{\l_\theta(e_i)_{[1]}} \frac{\partial \l_\theta(e_i)_{[1]}}{\partial \theta}}{e^{\l_\theta(e_i)_{[0]}} + e^{\l_\theta(e_i)_{[1]}}} \right]\\
& +\mathbb{E}_{j\sim P_\text{p-OOD}} \left[ - \frac{\partial l_\theta(e_j)_{[1]}}{\partial \theta} +\frac{e^{\l_\theta(e_j)_{[0]}} \frac{\partial \l_\theta(e_j)_{[0]}}{\partial \theta} + e^{\l_\theta(e_j)_{[1]}} \frac{\partial \l_\theta(e_j)_{[1]}}{\partial \theta}}{e^{\l_\theta(e_j)_{[0]}} + e^{\l_\theta(e_j)_{[1]}}} \right] \\
&-\mathbb{E}_{i\sim P_\text{ID}} \frac{e^{\l_\theta(e_i)_{[0]}} \frac{\partial \l_\theta(e_i)_{[0]}}{\partial \theta} + e^{\l_\theta(e_i)_{[1]}} \frac{\partial \l_\theta(e_i)_{[1]}}{\partial \theta}}{e^{\l_\theta(e_i)_{[0]}} + e^{\l_\theta(e_i)_{[1]}}}\\
& + \mathbb{E}_{j\sim P_\text{p-OOD}}  \frac{e^{\l_\theta(e_j)_{[0]}} \frac{\partial \l_\theta(e_j)_{[0]}}{\partial \theta} + e^{\l_\theta(e_j)_{[1]}} \frac{\partial \l_\theta(e_j)_{[1]}}{\partial \theta}}{e^{\l_\theta(e_j)_{[0]}} + e^{\l_\theta(e_j)_{[1]}}}\\
& \text{Define the energy w.r.t. label y as $E(e, y) = - l_\theta(e)_{[y]}$}\\
& = \mathbb{E}_{i\sim P_\text{ID}} \frac{\partial E(e_i, 0)}{\partial \theta} +\mathbb{E}_{j\sim P_\text{p-OOD}} \frac{\partial E(e_j, 1)}{\partial \theta} \\
& - 2 \left(\phi(\l_\theta(e_j))_{[0]} \frac{\partial E(e_j, 0)}{\partial \theta} + \phi(\l_\theta(e_j))_{[1]} \frac{\partial E(e_j, 1)}{\partial \theta}\right)\\
& = \mathbb{E}_{i\sim P_\text{ID}} \frac{\partial E(e_i, 0)}{\partial \theta} +\mathbb{E}_{j\sim P_\text{p-OOD}} (1-2\phi(\l_\theta(e_j))_{[1]}) \frac{\partial E(e_j, 1)}{\partial \theta} \\
& - 2 \phi(\l_\theta(e_j))_{[0]} \frac{\partial E(e_j, 0)}{\partial \theta}
\end{split}
\end{equation}

From the above equation, the training procedure that overall minimises the first order gradient of the negative sum of $\mathcal{L}_{\text{Unc}}$ and the surrogate function of $\mathcal{L}_{\text{DReg}}$ will decrease the energy score $E(e_i, 0; l_\theta)$ for in-distribution data, and increase the energy score $E(e_j, 1; l_\theta)$ and $E(e_j, 0; l_\theta)$ for pseudo-OOD data, given $\phi(\l_\theta(e_j))_{[1]} > 0.5$ as the detector continues to improve detection performance.
\end{proof}

\subsection{Logits vs. Softmax discrepancy} \label{Appendix:Logits vs. SFM}
\begin{figure}[h!]
    \centering
    \begin{subfigure}{0.45\linewidth}
        \includegraphics[width=\linewidth]{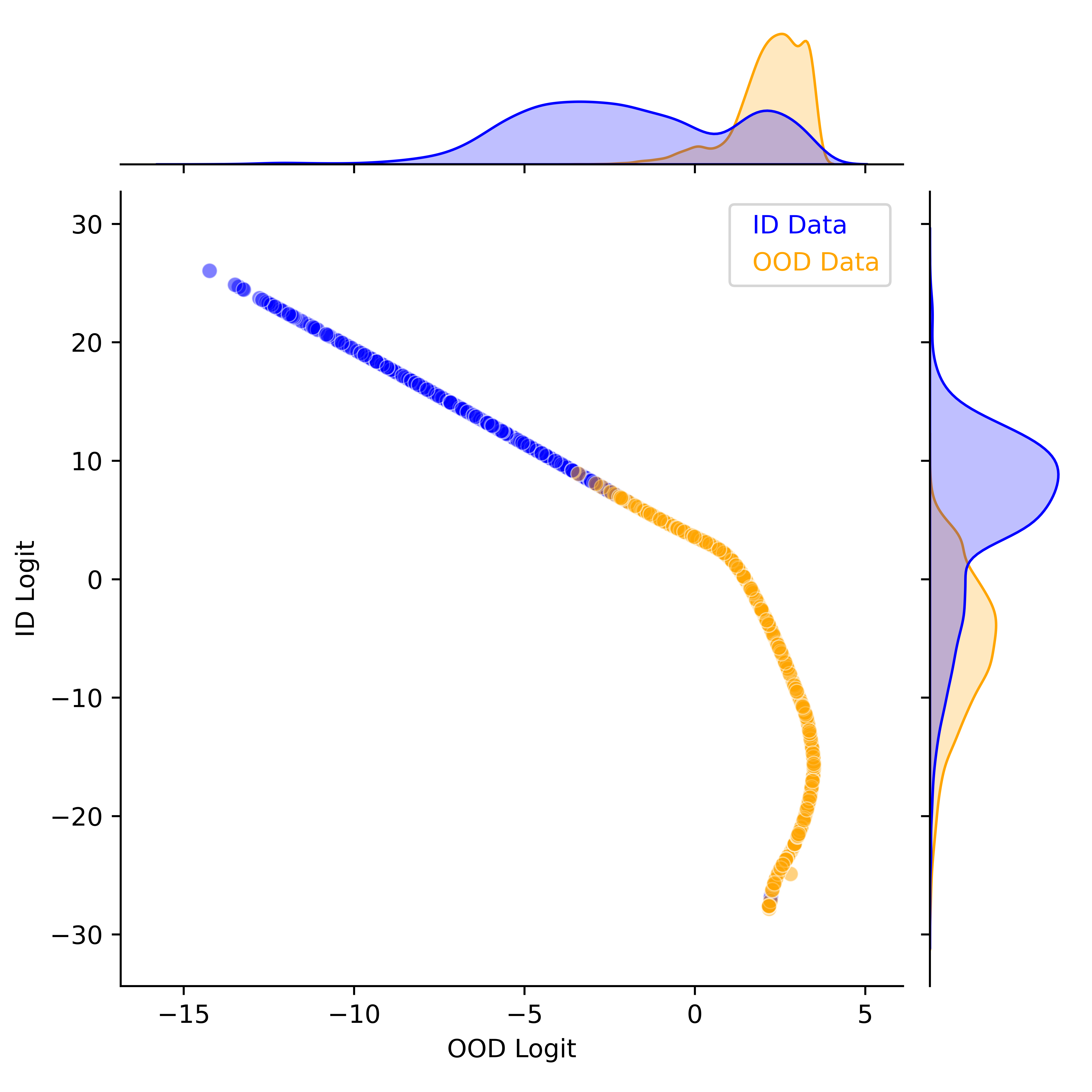}
        \caption{Logits Joint distribution}
        \label{fig:figure1}
    \end{subfigure}
    \hspace{0.05\linewidth} 
    \begin{subfigure}{0.45\linewidth}
        \includegraphics[width=\linewidth]{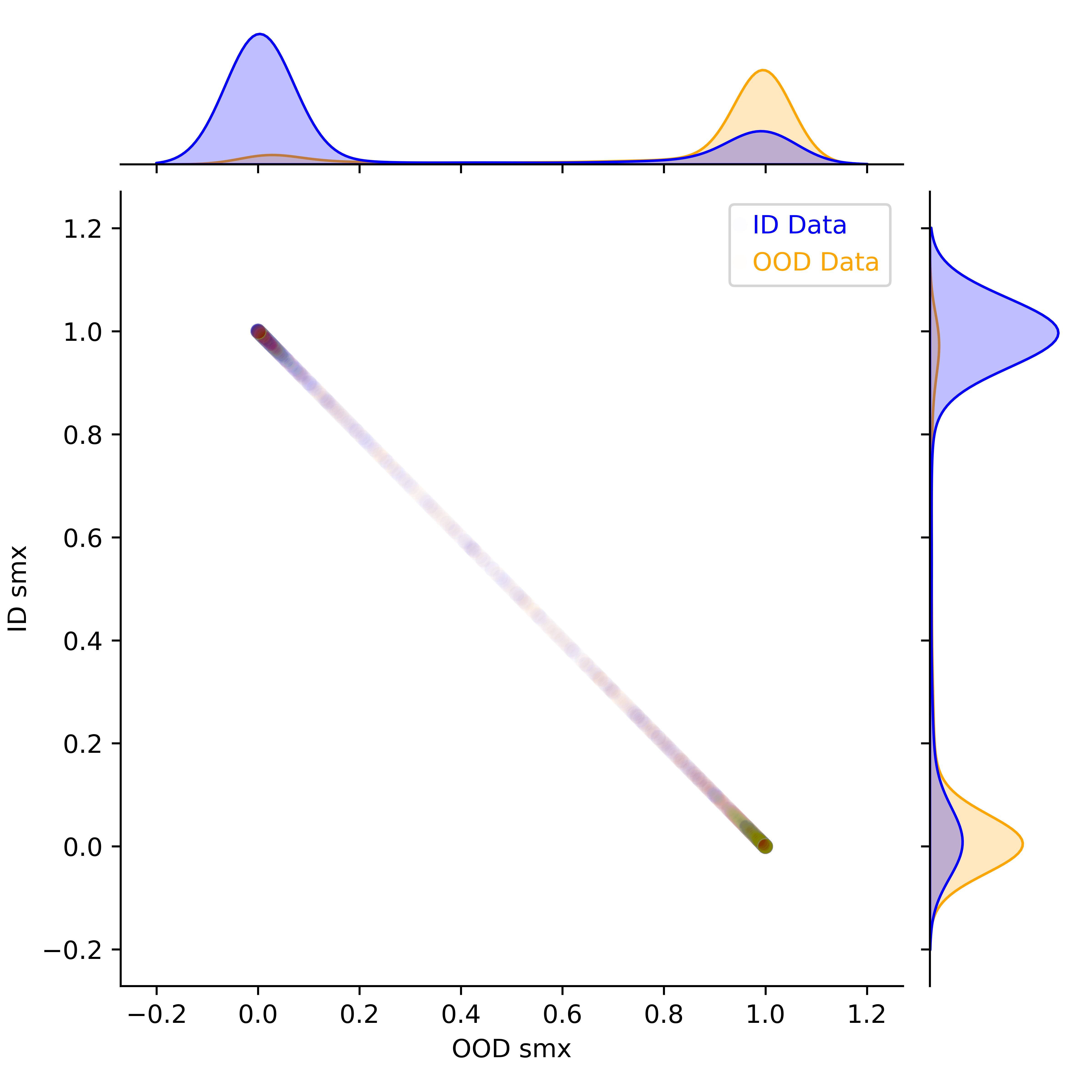} 
        \caption{Softmax Joint distribution}
        \label{fig:figure2}
    \end{subfigure}
    \caption{Logits vs. Softmax joint distribution plot for Twitch dataset.}
    \label{fig:Logits_smf}
\end{figure}

In this section, we present an empirical evaluation of the Logits vs. Softmax discrepancy between ID and OOD data from the detector. It is evident from Figure~\ref{fig:Logits_smf}, while the softmax confidence scores present high confidence for ID and OOD instances, where the majority of the scores corresponding to the respective class were close to 1 (based on the marginal distribution), the energy scores provide more meaningful information for distinguishing between them. Notably, ID data typically possess higher and positive ID Logits and lower OOD logits than OOD data. Thereby, leading to more distinguishable energy scores than softmax for OOD detection. 

\subsection{Extended Related Work} \label{Appendix:related_work}
\paragraph{Non-OOD-Exposure OOD Detection.} OOD detection is a fundamental task extensively studied in diverse machine learning domains~\citep{ConfOOD, EBFeat, RPRW, regOOD, MSP, Mahalanobis, GenOE, OECC, FocalOE}. A representative line of work that relies on purely ID data is based on the model's output including using softmax score~\citep{Hendrycks17softmax, ODIN}, using energy score~\citep{energy, Wang21energy, NODESAFE}, and activation pruning-based methods~\citep{ASH,DICE, React}. Other approaches involve confidence enhancement~\citep{Gen-ODIN, WhyReLU, Ensemble}, feature learning~\citep{MOOD, NMD}, and adversarial strategies~\citep{GOOD-cert, ATOM, RND}. More recent studies have applied OOD detection to graph-structured data~\citep{SGOOD, grasp, GOODAT, LMN, UGTs, G_UQ, EMP, advOOD, SLW,GOOD,GOODD-uncertainty}. For node-level detection, \textsc{GNNSafe} considers the inter-dependence nature of node instances and proposes an energy propagation schema~\citep{GNNSafe}. \textsc{NODESafe} builds upon \textsc{GNNSafe} and proposes additional regularisation terms to reduce and bound the generation of extreme energy scores~\citep{NODESAFE}. GKDE proposes a multi-source uncertainty framework to estimate the node-level Dirichlet distributions to assist OOD detection~\citep{GKDE}. GPN applies Bayesian posterior and density estimation to estimate the uncertainty for each node~\citep{GPN}. For graph-level detection, recent methods include modelling distribution shifts through a graph generative process, overseeing from a data-centric perspective, and unsupervised methods~\citep{GraphDE, AAGOD, GOOD-D}. 

\paragraph{OOD Exposure-Based OOD Detection.}
OOD exposure is another prominent line of work that adopts auxiliary OOD data to assist training~\citep{OE, energy, Textual-OODExposure, DivOE, ATOL, SAL,GNNSafe}. The aforementioned \textsc{GNNSafe} model also considers an additional version \textsc{GNNSafe++} to adopt OOD exposure and has shown greater performance than the standard model~\citep{GNNSafe}. \cite{NODESAFE} also presents \textsc{NODESafe++} as an extended OOD exposed version. \cite{GDE_OOD} proposes a generalised Dirichlet energy score for OOD detection. Our proposed GOLD method attempts to take advantage of the effectiveness of OOD exposure by synthesising samples that exhibit OOD characteristics. Thus, avoiding the necessity of real OOD data during training.

\paragraph{OOD Generation.}
Recent studies begin to work on synthesising OOD data~\citep{manifold, Likelihood, LRegret, GenAnalysis, GenUnknown, Hierarc,ConfOOD,VOS,NPOS}. A GAN-based approach is proposed to generate OOD data by jointly training a confidence classifier~\citep{ConfOOD}. VOS generates synthetic outliers from low-probability regions of multivariate Gaussian distributions~\citep{VOS}. Recently,pre-trained diffusion models have been widely employed for OOD generation including DFDD~\citep{DFDD}, Dream-OOD~\citep{Dream-OOD}. Several initial graph-level OOD studies have been initiated, predominantly for molecule~\citep{MOOD-Molecule,MOL_DIF}. A score-based OOD molecule generation model is proposed by MOOD~\citep{MOOD-Molecule}, which employs an OOD-controlled reverse-time diffusion. A recent work PGR-MOOD~\citep{MOL_DIF} proposes to rely on a pre-trained molecule diffusion for generation. These methods typically rely on pre-trained models that are trained with additional data. In contrast, GOLD does not rely on pre-trained generative models to synthesise pseudo-OOD data.

\subsection{Potential Limitations} \label{Appendix:Limitation}
In our concluding remarks, we highlight that our methodology leverages a generative model (specifically, diffusion model) to generate effective pseudo-OOD instances for OOD detection. To curb computational expenses, we employ a latent diffusion model, which reduces the computational demands of direct input space manipulation. Despite this, training-time efficiency may still be impacted. Nonetheless, during the inference phase, our model does not necessitate the generation of extra data, thus mitigating the impact of high latency. Moreover, we have experimented with a lightweight VAE as the latent generative model, which can achieve a competitive computational time as the standard SOTA baselines. Additionally, our approach currently targets node-level prediction tasks; however, we envisage its applicability to graph-level OOD detection, which we leave for future research. Following the propositions of \cite{energy} and \cite{GNNSafe}, our framework incorporates an energy-bounded regulariser that ideally ensures ID scores are lower than those of OOD samples, as illustrated in our visualisations in Section \ref{sec:vis}. Extended experiments detailed in Table~\ref{Appendix:Cora_reg_effectiveness} reveal that using only the energy regulariser results in AUROC scores near the single-digit range. This outcome highlights the regulariser's limitations and challenges the assumption that OOD energy scores consistently exceed ID scores, thereby undermining the effectiveness of the OOD metric in true detection performance. Nevertheless, our framework introduces an additional regulariser, which effectively addresses these discrepancies, as showcased by our consistently positive results.

\subsection{GNN} \label{Appendix:GNN}
GNNs, by their very nature, excel in modelling the complex relationship of node-dependence in graphs. Central to their success is the message-passing mechanism, which iteratively aggregates neighbouring information towards the centre node to capture both local and global knowledge. Denote the learnt representation of node $i$ at the $l$-th layer as $\mathbf{h}_i^{(l)}$, a typical Graph Convolutional Network (GCN) executes recursive layer propagation via:
\begin{equation}
\mathbf{H}^{(l)} = \sigma ( \mathbf{D}^{-1/2} \mathbf{\Tilde{A}}\mathbf{D}^{-1/2} \mathbf{H}^{(l - 1)} \mathbf{W}^{(l)}), \mathbf{H}^{l-1} = [{\mathbf{h}}_i^{l-1}], \mathbf{H}^{(0)} = {\mathbf{X}} \label{eq:GCN} 
\end{equation}
with $\mathbf{\Tilde{A}} = \mathbf{A} + \mathbf{I}$, where $\mathbf{I}$ is the identity matrix, $\mathbf{D}$ is the diagonal degree matrix of $\mathbf{\Tilde{A}} $, $\sigma$ is a non-linear activation function (i.e., ReLU), and $\mathbf{W}^{(l)}$ is the corresponding weight matrix at layer $l$~\citep{GCN}.


\subsection{Description of Datasets} \label{Appendix:dataset}
The datasets utilised in this study are publicly available benchmark datasets for graph learning. We follow the same data collection and processing protocol in \cite{GNNSafe} and utilised the data loader for the {\fontfamily{qcr}\selectfont ogbn-Arxiv} dataset provided by the OGB package\footnote{https://github.com/snap-stanford/ogb?tab=readme-ov-file}, and others from the Pytorch Geometric Package\footnote{https://pytorch-geometric.readthedocs.io/en/latest/modules/datasets.html}. For all datasets, we follow the provided splits and generation process in \cite{GNNSafe}. We provide a brief description of the datasets below:

The {\fontfamily{qcr}\selectfont TwitchGamers - Explicit} dataset consists of multiple subgraphs, each representing a social network from a different region~\citep{Twitch_Gamers}. The nodes within these subgraphs indicate Twitch gamers, while the edges depict the follower relationships between two users. Node features include embeddings based on the games played by Twitch users, and for this study, we focus on the label that indicates whether a user broadcasts mature content (i.e., Explicit). We utilise subgraph DE as ID data, and subgraphs ES, FR, RU as testing data. Dataset details are provided in Table \ref{appendix:twitch_table}.

\begin{table} [H]
    \centering
    \resizebox{0.8\linewidth}{!}{\begin{tabular}{c|c|c|c|c|c}
        \toprule
        {\fontfamily{qcr}\selectfont Twitch} & Splits & $\#$ Nodes & $\#$ Edges & Feature Dimension & $\#$ Classes\\
        \midrule
        {\fontfamily{qcr}\selectfont Twitch-DE} & ID& 9498 & 315774 & 128 & 2\\
        {\fontfamily{qcr}\selectfont Twitch-ES} & OOD&4648 & 123412 & 128 & 2\\
       {\fontfamily{qcr}\selectfont Twitch-FR} & OOD&6551 & 231883 & 128 & 2 \\
        {\fontfamily{qcr}\selectfont Twitch-RU}& OOD&4385 & 78993 & 128 & 2 \\
        \bottomrule
    \end{tabular}}
    \caption{{\fontfamily{qcr}\selectfont Twitch} dataset overview}
    \label{appendix:twitch_table}
\end{table}

The {\fontfamily{qcr}\selectfont Cora} dataset is a citation network where each node represents a published paper, and each edge reflects a citation relationship between papers~\citep{Cora}. The dataset consists of seven labels. Since {\fontfamily{qcr}\selectfont Cora} does not contain an explicit domain attribute to partition into OOD subgraphs, we follow the provided protocol in \cite{GNNSafe}, and synthetically create the OOD data as mentioned in Section \ref{sec:datasets}. Dataset details are provided in Table \ref{appendix:cora_table}.

\begin{table} [H]
    \centering
    \resizebox{0.8\linewidth}{!}{\begin{tabular}{c|c|c|c|c|c}
        \toprule
        {\fontfamily{qcr}\selectfont Cora} & Splits&$\#$ Nodes & $\#$ Edges & Feature Dimension & $\#$ Classes\\
        \midrule
        {\fontfamily{qcr}\selectfont Cora-S} &ID & 2708  & 10556  & 1433 & 7\\
        {\fontfamily{qcr}\selectfont Cora-S} &OOD & 2708  & 6696 & 1433 & 7\\
        \midrule
        {\fontfamily{qcr}\selectfont Cora-F} &ID & 2708  & 10556  & 1433 & 7\\
        {\fontfamily{qcr}\selectfont Cora-F} &OOD & 2708  & 10556 & 1433 & 7\\
        \midrule
        {\fontfamily{qcr}\selectfont Cora-L} &ID & 904   & 10556  & 1433 & 3\\
        {\fontfamily{qcr}\selectfont Cora-L} &OOD & 986   & 10556 & 1433 & 3\\
        
        \bottomrule
    \end{tabular}}
    \caption{{\fontfamily{qcr}\selectfont Cora} dataset overview}
    \label{appendix:cora_table}
\end{table}

The {\fontfamily{qcr}\selectfont Amazon-Photo} dataset forms an item co-purchasing network on Amazon, where each node represents a product and each edge signifies that the linked products are frequently bought together~\citep{AmazonPhoto}. Node labels categorise the products. Similar to the {\fontfamily{qcr}\selectfont Cora} dataset, we employ three synthetic methods to create the OOD data due to the lack of a clear domain for partition. Dataset details are provided in Table \ref{appendix:amazon_table}.

\begin{table} [H]
    \centering
    \resizebox{0.8\linewidth}{!}{\begin{tabular}{c|c|c|c|c|c}
        \toprule
        {\fontfamily{qcr}\selectfont Amazon-Photo} & Splits&$\#$ Nodes & $\#$ Edges & Feature Dimension & $\#$ Classes\\
        \midrule
        {\fontfamily{qcr}\selectfont Amazon-S} &ID & 7650    & 238162  & 745 & 8\\
        {\fontfamily{qcr}\selectfont Amazon-S} &OOD & 7650   & 149168 & 745 & 8\\
        \midrule
        {\fontfamily{qcr}\selectfont Amazon-F} &ID & 7650   & 238162  & 745 & 8\\
        {\fontfamily{qcr}\selectfont Amazon-F} &OOD & 7650   & 238162 & 745 & 8\\
        \midrule
        {\fontfamily{qcr}\selectfont Amazon-L} &ID & 3095    & 238162  & 745 & 3\\
        {\fontfamily{qcr}\selectfont Amazon-L} &OOD & 3673     & 238162 & 745 & 4\\
        
        \bottomrule
    \end{tabular}}
    \caption{{\fontfamily{qcr}\selectfont Amazon-Photo} dataset overview}
    \label{appendix:amazon_table}
\end{table}

The {\fontfamily{qcr}\selectfont Coauthor-CS} dataset describes a network of computer science coauthors. Such that each node represents an author, and edges connect any two authors who have collaborated on a paper. The dataset aims to classify authors into their respective fields of study based on the keywords from their publications, which are also used as node features. Due to the lack of a clear domain to split the data, OOD graphs were constructed following the same protocol aforementioned. Dataset details are provided in Table \ref{appendix:coauthor_table}.

\begin{table} [H]
    \centering
    \resizebox{0.8\linewidth}{!}{\begin{tabular}{c|c|c|c|c|c}
        \toprule
        {\fontfamily{qcr}\selectfont Coauthor-CS} & Splits&$\#$ Nodes & $\#$ Edges & Feature Dimension & $\#$ Classes\\
        \midrule
        {\fontfamily{qcr}\selectfont Coauthor-S} &ID & 18333 & 163788 & 6805 & 15\\
        {\fontfamily{qcr}\selectfont Coauthor-S} &OOD & 18333 & 92802 & 6805 & 15\\
        \midrule
        {\fontfamily{qcr}\selectfont Coauthor-F} &ID & 18333  & 163788  & 6805 & 15\\
        {\fontfamily{qcr}\selectfont Coauthor-F} &OOD & 18333 & 163788  & 6805 & 15 \\
        \midrule
        {\fontfamily{qcr}\selectfont Coauthor-L} &ID & 13290 & 163788  & 6805 & 10\\
        {\fontfamily{qcr}\selectfont Coauthor-L} &OOD & 3649  & 163788 & 6805 & 4\\
        
        \bottomrule
    \end{tabular}}
    \caption{{\fontfamily{qcr}\selectfont Coauthor-CS} dataset overview}
    \label{appendix:coauthor_table}
\end{table}

The {\fontfamily{qcr}\selectfont ogbn-Arxiv} dataset curated an extensive dataset from 1960 to 2020, where each node represents a paper, labelled by its subject area for classification~\citep{Arxiv}. Edges reflect the citation relationships among papers, and each node is associated with a 128-dimensional vector derived from word embeddings of its title and abstract. Following \cite{GNNSafe}, we utilise time information to partition the graph, where data before 2015 are used as ID data and papers published after 2017 are used as OOD data. Dataset details are provided in Table \ref{appendix:arxiv_table}.

\begin{table} [H]
    \centering
    \resizebox{0.8\linewidth}{!}{\begin{tabular}{c|c|c|c|c|c}
        \toprule
        {\fontfamily{qcr}\selectfont ogbn-Arxiv} & Splits&$\#$ Nodes & $\#$ Edges & Feature Dimension & $\#$ Classes\\
        \midrule
        {\fontfamily{qcr}\selectfont Arxiv-2015} &ID & 53160   & 152226  & 128 & 40 \\
        {\fontfamily{qcr}\selectfont Arxiv-2018} &OOD & 29799  & 622466 & 128 & 40\\
        {\fontfamily{qcr}\selectfont Arxiv-2019} & OOD & 39711   & 1061197  & 128 & 40\\
        {\fontfamily{qcr}\selectfont Arxiv-2020} &OOD & 8892   & 1166243 & 128 & 40\\
        
        \bottomrule
    \end{tabular}}
    \caption{{\fontfamily{qcr}\selectfont ogbn-Arxiv} dataset overview}
    \label{appendix:arxiv_table}
\end{table}

\subsection{Evaluation Metrics} \label{Appendix:Metrics}
In this section, we provide a detailed description of the metrics used for evaluation. Following common practice in OOD detection~\citep{GNNSafe, energy, GOOD-D}, we employed three key metrics to measure the performance of detecting OOD instances: (1) the Area Under the Receiver Operating Characteristic curve (AUROC); (2) the Area Under the Precision-Recall curve (AUPR); and (3) the false positive rate (FPR95) of OOD examples when the true positive rate of ID examples is 95\%. AUROC measures the trade-off between the true positive rate (TPR) and the false positive rate (FPR) at different threshold levels, providing insights into the model's ability to accurately distinguish between ID and OOD instances. However, in highly imbalanced datasets with only a few OOD instances, AUROC might be overly optimistic. AUPR, on the other hand, offers a more realistic performance measure by accounting for both precision and recall. FPR95 provides further insight into the model's performance under high-sensitivity conditions, indicating the probability of misclassifying in-distribution samples as OOD when the TPR is 95\%.

\subsection{Implementation Details} \label{Appendix:implementation_details}
\begin{figure}[!h]
    \centering
    \begin{subfigure}[b]{0.35\textwidth}
        \centering
        \includegraphics[width=\textwidth]{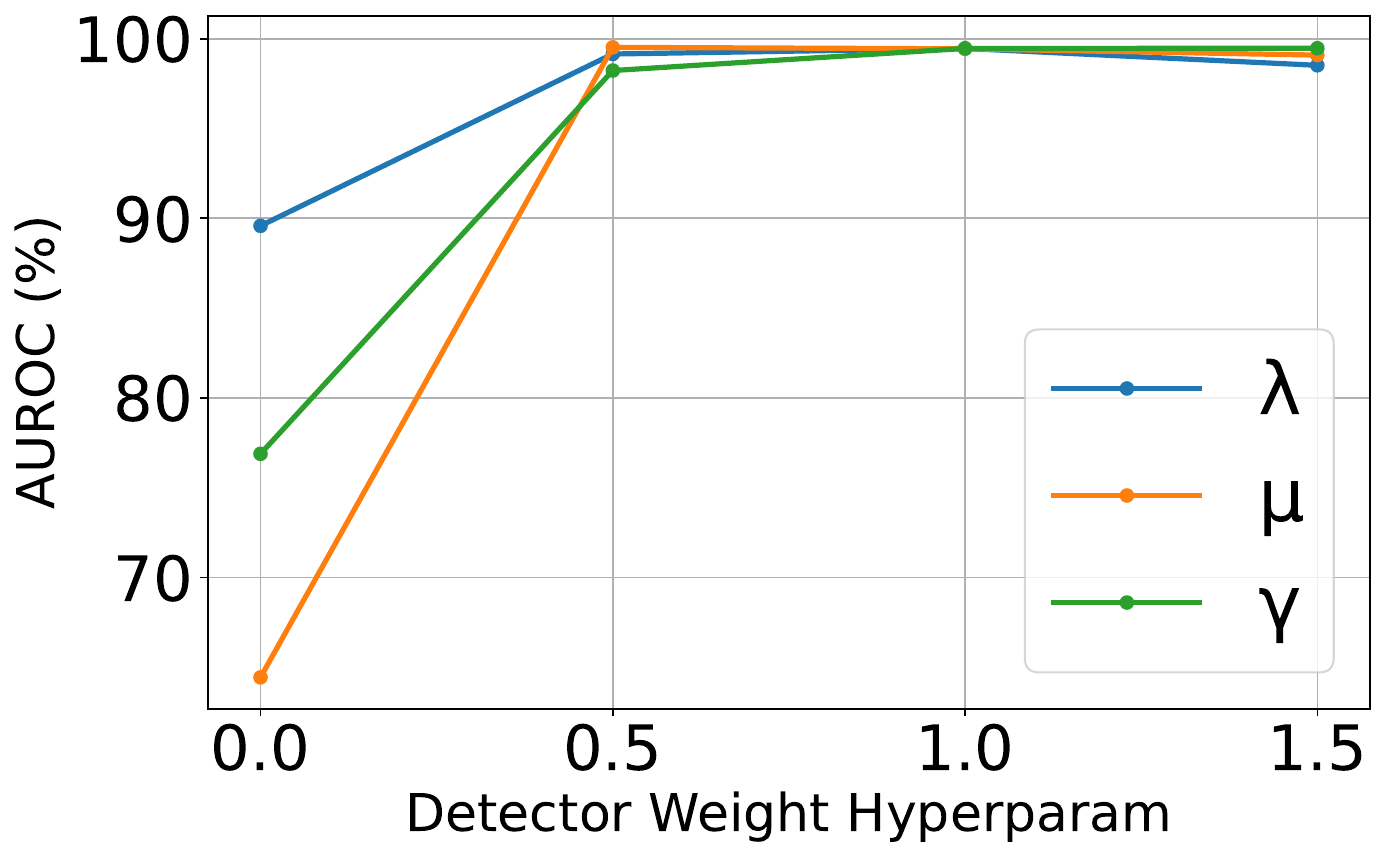}
        \caption{}
        \label{fig:AUROC_HYPER}
    \end{subfigure}
    \begin{subfigure}[b]{0.35\textwidth}
        \centering
        \includegraphics[width=\textwidth]{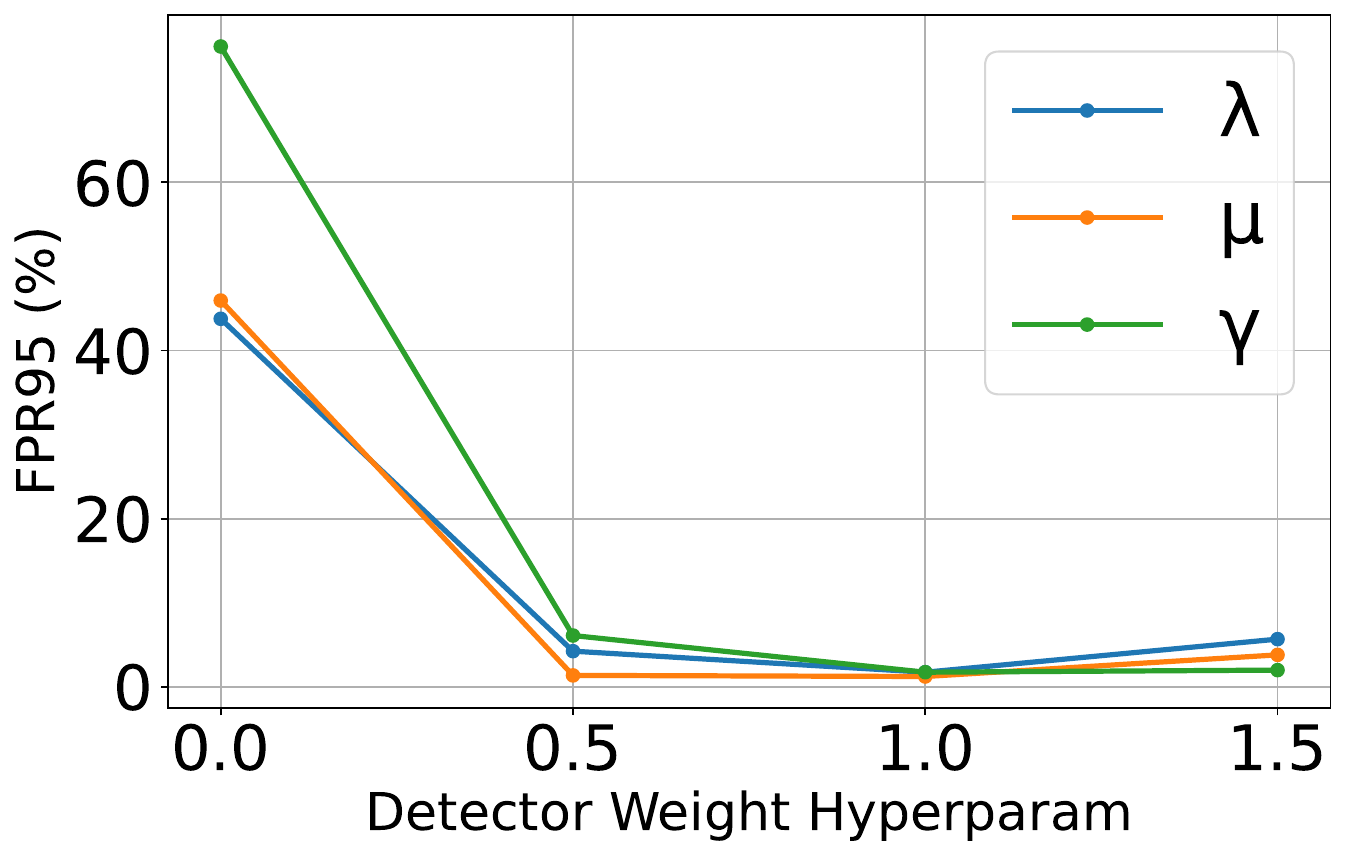}
        \caption{}
        \label{fig:FPR_HYPER}
    \end{subfigure}
    \caption{The {\fontfamily{qcr}\selectfont Twitch} dataset was utilised for conducting hyper-parameter sensitivity analysis. (\ref{fig:AUROC_HYPER}) and (\ref{fig:FPR_HYPER}) are Hyper-parameter sensitivity of different weights in Eq.~\ref{eq:div} for Detector measured by AUROC and FPR95.} 
    \label{Appendix:Hyperparam}
\end{figure}

We utilised the publicly available benchmark (i.e., datasets and baselines) provided by \cite{GNNSafe}, and fully respect their CC-BY 4.0 license. The experiments were conducted using Python 3.8.0 and PyTorch 2.2.2 with Cuda 12.1, using Tesla V100 GPUs with 32GB memory for experiments. The datasets were obtained from Pytorch Geometric 2.0.3 and OGB 1.3.3 under the MIT license. Extending beyond the thresholds provided in \cite{GNNSafe}, we tuned the margins $t_\text{ID}$ and $t_\text{OOD}$ with various ranges for different dataset (i.e., for {\fontfamily{qcr}\selectfont Twitch} $t_\text{ID}\in\{-5, -4, -3\}, t_\text{OOD} \in \{1, 2, 3\} $). The detector loss weights $\lambda,\mu,\gamma$ are tuned in the range of $\{0, 0.3, 0.5, 0.7, 1, 1.5\}$, depending on the dataset. Hyperparameter sensitivity analysis for the detector and classifier loss objective can be found in Figure \ref{Appendix:Hyperparam}. The LGM training step $M_1$ is configured in the range of $\{100, 200, 600, 800\}$, and the classifier and detector update $M_2$ is tuned from $\{5-20\}$ subject to the dataset, with early stopping applied to ensure the ID accuracy does not reduce significantly. Regarding baseline models, we utilised the provided benchmark in \cite{GNNSafe}, which includes modified versions of the different baseline models.
This involves adapting to the same encoder GCN backbone (i.e., a hidden size of 64 and layer number of 2) for MSP, ODIN, Mahalanobis, Energy and Energy FT. We also considered latest SOTA OOD Detection method by \cite{NODESAFE} using their reported results.

\subsection{Additional Experiment Results} \label{Appendix:Additional_exp}
In this section, we provide additional experimental results to supplement the results provided in the maintext. Specifically, we present detailed OOD detection performance of the subsets for each OOD dataset (i.e., subgraphs for {\fontfamily{qcr}\selectfont Twitch}, three types of OOD data for {\fontfamily{qcr}\selectfont Cora}, {\fontfamily{qcr}\selectfont Amazon}, and {\fontfamily{qcr}\selectfont Coauthor}, and different years for {\fontfamily{qcr}\selectfont Arxiv}) in Tables \ref{Appendix:twitch_full} to \ref{Appendix:arxiv_full}, complementing Table \ref{Table:overall_performance} in the main text. Furthermore, in Tables \ref{Appendix:Full_ablation_general} and \ref{Appendix:Full_ablation_adversarial}, we report an extended version of the ablation study and adversarial training effectiveness, covering the subsets of {\fontfamily{qcr}\selectfont Twitch}, {\fontfamily{qcr}\selectfont Cora}, and {\fontfamily{qcr}\selectfont Arxiv}, supplementing Table \ref{Table:Ablation_general} and \ref{Table:Ablation_adversarial} in the maintext. Lastly, we provide the full tables for the energy regulariser analysis in Table \ref{Table:Ablation_regulariser} for {\fontfamily{qcr}\selectfont Twitch}, {\fontfamily{qcr}\selectfont Cora}, and {\fontfamily{qcr}\selectfont Amazon} in Tables \ref{Appendix:Twtich_reg_effectiveness}, \ref{Appendix:Cora_reg_effectiveness}, and \ref{Appendix:Amazon_reg_effectiveness}, respectively.

\begin{table}[H]
    \centering
    \caption{Model performance on OOD sub-graphs ES, FR and RU of {\fontfamily{qcr}\selectfont Twitch} dataset.}
    \resizebox{1\linewidth}{!}{
    \begin{tabular}{c|c|ccccccc|ccc|c}
    \specialrule{.1em}{.05em}{.05em} 
    \multirow{2}{*}{\textbf{Dataset}} & \multirow{2}{*}{\textbf{Metrics}} & \multicolumn{7}{c}{\textbf{Non-OOD Exposure}} & \multicolumn{3}{c}{\textbf{Real OOD Exposure}} & \multicolumn{1}{c}{\textbf{Ours}}\\
    & & \textbf{MSP} & \textbf{ODIN} & \textbf{Mahalanobis} & \textbf{Energy} & \textbf{GKDE} & \textbf{GPN} & \textbf{\textsc{GNNSafe}} & \textbf{OE} & \textbf{Energy FT} & \textbf{\textsc{GNNSafe++}} & \textbf{GOLD}\\
    \midrule
    \multirow{4}{*}{{\fontfamily{qcr}\selectfont Twitch-ES}} & AUROC & 37.72 & 83.83 & 45.66 & 38.80 & 48.70 & 53.00 & 49.07 & 55.97 & 80.73 & 94.54 & 99.72 $\pm$ 0.03 \\
                            & AUPR  & 53.08 & 80.43 & 58.82 & 54.26 & 61.05 & 64.24 & 57.62 & 69.49 & 87.56 & 97.17 & 99.82 $\pm$ 0.02 \\
                            & FPR95   & 98.09 & 33.28 & 95.48 & 95.70 & 95.37 & 95.05 & 93.98 & 94.94 & 76.76 & 44.06 & 0.44 $\pm$0.13 \\
                            & ID ACC & 68.72 & 70.79 & 70.51 & 70.40 & 67.44 & 68.09 & 70.40 & 70.73 & 70.52 & 70.18 & 68.49 $\pm$ 0.13 \\
    \midrule
    \multirow{4}{*}{{\fontfamily{qcr}\selectfont Twitch-FR}} & AUROC & 21.82 & 59.82 & 40.40 & 57.21 & 49.19 & 51.25 & 63.49 & 45.66 & 79.66 & 93.45 & 99.08 $\pm$ 0.19 \\
                            & AUPR  & 38.27 & 64.63 & 46.69 & 61.48 & 52.94 & 55.37 & 66.25 & 54.03 & 81.20 & 95.44 & 99.25 $\pm$ 0.15 \\
                            & FPR95   & 99.25 & 92.57 & 95.54 & 91.57 & 95.04 & 93.92 & 90.80 & 95.48 & 76.39 & 51.06 & 3.77$\pm$ 0.92 \\
                            & ID ACC & 68.72 & 70.79 & 70.51 & 70.40 & 67.44 & 68.09 & 70.40 & 70.73 & 70.52 & 70.18 & 68.49 $\pm$ 0.13 \\
    \midrule
    \multirow{4}{*}{{\fontfamily{qcr}\selectfont Twitch-RU}} & AUROC & 41.23 & 58.67 & 55.68 & 57.72 & 46.48 & 50.89 & 87.90 & 55.72 & 93.12 & 98.10 & 99.58 $\pm$ 0.06 \\
                            & AUPR  & 56.06 & 72.58 & 66.42 & 66.68 & 62.11 & 65.14 & 89.05 & 70.18 & 95.36 & 98.74 & 99.78$\pm$ 0.04 \\
                            & FPR95   & 95.01 & 93.98 & 90.13 & 87.57 & 95.62 & 99.93 & 43.95 & 95.07 & 30.72 & 5.59 & 
                            1.14 $\pm$ 0.35 \\
                            & ID ACC & 68.72 & 70.79 & 70.51 & 70.40 & 67.44 & 68.09 & 70.40 & 70.73 & 70.52 & 70.18 & 68.49 $\pm$ 0.13 \\
    \specialrule{.1em}{.05em}{.05em} 
\end{tabular}} \label{Appendix:twitch_full}
\end{table}

\begin{table}[H]
    \centering
    \caption{Model performance on {\fontfamily{qcr}\selectfont Cora} with three types of OOD (\textbf{S}tructure manipulation, \textbf{F}eature interpolation, and \textbf{L}abel leave-out).}
    \resizebox{1\linewidth}{!}{
    \begin{tabular}{c|c|ccccccc|ccc|c}
    \specialrule{.1em}{.05em}{.05em} 
    \multirow{2}{*}{\textbf{Dataset}} & \multirow{2}{*}{\textbf{Metrics}} & \multicolumn{7}{c}{\textbf{Non-OOD Exposure}} & \multicolumn{3}{c}{\textbf{Real OOD Exposure}} & \multicolumn{1}{c}{\textbf{Ours}}\\
    & & \textbf{MSP} & \textbf{ODIN} & \textbf{Mahalanobis} & \textbf{Energy} & \textbf{GKDE} & \textbf{GPN} & \textbf{\textsc{GNNSafe}} & \textbf{OE} & \textbf{Energy FT} & \textbf{\textsc{GNNSafe++}} & \textbf{GOLD}\\
    \midrule
\multirow{4}{*}{{\fontfamily{qcr}\selectfont Cora-S}} & AUROC & 70.90 & 49.92 & 46.68 & 71.73 & 68.61 & 77.47 & 87.52 & 67.98 & 75.88 & 90.62 & 95.48 $\pm$ 0.28 \\
                            & AUPR  & 45.73 & 27.01 & 29.03 & 46.08 & 44.26 & 53.26 & 77.46 & 46.93 & 49.18 & 81.88 & 91.06 $\pm$ 0.32 \\
                            & FPR95   & 87.30 & 100.00 & 98.19 & 88.74 & 84.34 & 76.22 & 73.15 & 95.31 & 67.73 & 53.51 & 21.86 $\pm$ 0.97 \\
                            & ID ACC & 75.50 & 74.90 & 74.90 & 76.00 & 73.70 & 76.50 & 75.80 & 71.80 & 75.50 & 76.10 & 77.4 $\pm$ 0.56 \\
\midrule
\multirow{4}{*}{{\fontfamily{qcr}\selectfont Cora-F}} & AUROC & 85.39 & 49.88 & 49.93 & 86.15 & 82.79 & 85.88 & 93.44 & 81.83 & 88.15 & 95.56 & 96.64 $\pm$ 0.15 \\
                            & AUPR  & 73.70 & 26.96 & 31.95 & 74.42 & 66.52 & 73.79 & 88.19 & 70.84 & 75.99 & 90.27 & 93.82 $\pm$ 0.24 \\
                            & FPR95   & 64.88 & 100.00 & 99.93 & 65.81 & 68.24 & 56.17 & 38.92 & 83.79 & 47.53 & 27.73 & 14.35 $\pm$ 2.05 \\
                            & ID ACC & 75.30 & 75.00 & 74.90 & 76.10 & 74.80 & 77.00 & 76.40 & 73.30 & 75.30 & 76.80 & 76.77 $\pm$ 0.21\\
\midrule
\multirow{4}{*}{{\fontfamily{qcr}\selectfont Cora-L}} & AUROC & 91.36 & 49.80 & 67.62 & 91.40 & 57.23 & 90.34 & 92.80 & 89.47 & 91.36 & 92.75 & 95.40 $\pm$ 0.17 \\
                            & AUPR  & 78.03 & 24.27 & 42.31 & 78.14 & 27.50 & 77.40 & 82.21 & 77.01 & 78.49 & 82.64 & 88.65 $\pm$ 0.25 \\
                            & FPR95   & 34.99 & 100.00 & 90.77 & 41.08 & 88.95 & 37.42 & 30.83 & 46.55 & 37.83 & 34.08 & 17.28 $\pm$ 0.50 \\
                            & ID ACC & 88.92 & 88.92 & 88.92 & 88.92 & 89.87 & 91.46 & 88.92 & 87.97 & 90.51 & 91.46 & 90.82 $\pm$ 0.55 \\
\specialrule{.1em}{.05em}{.05em} 
    \end{tabular}} \label{Appendix:cora_full}
\end{table}

\begin{table}[H]
    \centering
    \caption{Model performance on {\fontfamily{qcr}\selectfont Amazon} with three types of OOD  (\textbf{S}tructure manipulation, \textbf{F}eature interpolation, and \textbf{L}abel leave-out).}
    \resizebox{1\linewidth}{!}{
    \begin{tabular}{c|c|ccccccc|ccc|c}
    \specialrule{.1em}{.05em}{.05em} 
    \multirow{2}{*}{\textbf{Dataset}} & \multirow{2}{*}{\textbf{Metrics}} & \multicolumn{7}{c}{\textbf{Non-OOD Exposure}} & \multicolumn{3}{c}{\textbf{Real OOD Exposure}} & \multicolumn{1}{c}{\textbf{Ours}}\\
    & & \textbf{MSP} & \textbf{ODIN} & \textbf{Mahalanobis} & \textbf{Energy} & \textbf{GKDE} & \textbf{GPN} & \textbf{\textsc{GNNSafe}} & \textbf{OE} & \textbf{Energy FT} & \textbf{\textsc{GNNSafe++}} & \textbf{GOLD}\\
    \midrule
\multirow{4}{*}{{\fontfamily{qcr}\selectfont Amazon-S}} & AUROC & 98.27 & 93.24 & 71.69 & 98.51 & 76.39 & 97.17 & 99.58 & 99.60 & 98.83 & 99.82 & 99.99 $\pm$ 0.03 \\
 & AUPR & 98.54 & 95.26 & 79.01 & 98.72 & 81.58 & 96.39 & 99.76 & 99.61 & 99.14 & 99.89 & 99.99 $\pm$ 0.02 \\
 & FPR95 & 6.13 & 65.44 & 99.91 & 4.97 & 99.25 & 11.65 & 0.00 & 0.51 & 1.31 & 0.00 & 0 $\pm$ 0 \\
 & ID ACC & 92.84 & 92.84 & 92.79 & 92.86 & 87.57 & 88.51 & 92.53 & 92.61 & 92.79 & 92.22 & 92.03 $\pm$ 0.24 \\

\midrule
\multirow{4}{*}{{\fontfamily{qcr}\selectfont Amazon-F}} & AUROC & 97.31 & 81.15 & 76.50 & 97.87 & 58.96 & 87.91 & 98.55 & 98.39 & 98.68 & 99.64 & 99.17 $\pm$ 0.02 \\
 & AUPR & 95.16 & 78.47 & 71.14 & 95.64 & 66.76 & 84.77 & 98.99 & 96.24 & 96.82 & 99.68 & 99.31 $\pm$ 0.06 \\
 & FPR95 & 8.72 & 100.0 & 76.12 & 6.00 & 99.28 & 49.11 & 0.31 & 4.34 & 2.84 & 0.13 & 0.14 $\pm$ 0.03 \\
 & ID ACC & 92.89 & 92.71 & 92.86 & 92.96 & 86.18 & 90.05 & 92.81 & 92.30 & 92.52 & 92.39 & 91.76 $\pm$ 0.57 \\

\midrule
\multirow{4}{*}{{\fontfamily{qcr}\selectfont Amazon-L}} & AUROC & 93.97 & 65.97 & 73.25 & 93.81 & 65.58 & 92.72 & 97.35 & 95.39 & 96.61 & 97.51 & 97.26 $\pm$ 0.27  \\
 & AUPR & 91.32 & 57.80 & 66.89 & 91.13 & 65.20 & 90.34 & 97.12 & 92.53 & 94.92 & 97.07 & 97.46 $\pm$ 0.29 \\
 & FPR95 & 26.65 & 90.23 & 74.30 & 28.48 & 96.87 & 37.16 & 6.59 & 17.72 & 13.78 & 6.18 & 6.06 $\pm$ 1.81 \\
 & ID ACC & 95.76 & 96.08 & 95.76 & 95.72 & 89.37 & 90.07 & 95.76 & 95.72 & 94.83 & 95.84 & 95.18 $\pm$ 0.81\\
\specialrule{.1em}{.05em}{.05em} 
    \end{tabular}} \label{Appendix:amazon_full}
\end{table}

\begin{table}[H]
    \centering
    \caption{Model performance on {\fontfamily{qcr}\selectfont Coauthor} with three types of OOD  (\textbf{S}tructure manipulation, \textbf{F}eature interpolation, and \textbf{L}abel leave-out).}
    \resizebox{1\linewidth}{!}{
    \begin{tabular}{c|c|ccccccc|ccc|c}
    \specialrule{.1em}{.05em}{.05em} 
    \multirow{2}{*}{\textbf{Dataset}} & \multirow{2}{*}{\textbf{Metrics}} & \multicolumn{7}{c}{\textbf{Non-OOD Exposure}} & \multicolumn{3}{c}{\textbf{Real OOD Exposure}} & \multicolumn{1}{c}{\textbf{Ours}}\\
    & & \textbf{MSP} & \textbf{ODIN} & \textbf{Mahalanobis} & \textbf{Energy} & \textbf{GKDE} & \textbf{GPN} & \textbf{\textsc{GNNSafe}} & \textbf{OE} & \textbf{Energy FT} & \textbf{\textsc{GNNSafe++}} & \textbf{GOLD}\\
    \midrule
    \multirow{4}{*}{{\fontfamily{qcr}\selectfont Coauthor-S}} & AUROC & 95.30 & 52.14 & 80.46 & 96.18 & 65.87 & 34.67 & 99.60 & 97.86 & 98.84 & 99.99 & 99.62 $\pm$ 0.02 \\
 & AUPR & 94.37 & 48.83 & 76.65 & 95.25 & 72.65 & 40.21 & 99.69 & 96.81 & 97.78 & 99.99 & 99.78 $\pm$ 0.01 \\
 & FPR95 & 24.75 & 99.92 & 70.75 & 18.02 & 99.48 & 99.57 & 0.26 & 9.23 & 3.97 & 0.02 & 0.01 $\pm$ 0.01 \\
 & ID ACC & 92.47 & 92.34 & 92.33 & 92.75 & 88.62 & 89.45 & 92.73 & 92.60 & 92.61 & 92.92 & 91.41 $\pm$ 0.16 \\
\midrule
\multirow{4}{*}{{\fontfamily{qcr}\selectfont Coauthor-F}} & AUROC & 97.05 & 51.54 & 93.23 & 97.88 & 80.69 & 81.77 & 99.64 & 99.04 & 99.43 & 99.97 & 99.78 $\pm$ 0.15 \\
 & AUPR & 96.93 & 45.50 & 90.88 & 97.69 & 86.47 & 80.56 & 99.66 & 98.80 & 99.25 & 99.95 & 99.86 $\pm$ 0.09 \\
 & FPR95 & 15.55 & 100.0 & 28.10 & 9.75 & 96.57 & 74.46 & 0.51 & 4.44 & 2.25 & 0.09 & 0.03 $\pm$ 0.01 \\
 & ID ACC & 92.45 & 92.39 & 92.34 & 92.75 & 84.72 & 87.05 & 92.73 & 92.64 & 92.50 & 92.87 & 91.81 $\pm$ 0.26 \\
\midrule
\multirow{4}{*}{{\fontfamily{qcr}\selectfont Coauthor-L}} & AUROC & 94.88 & 51.44 & 85.36 & 95.87 & 61.15 & 93.24 & 97.23 & 96.04 & 96.23 & 97.89 & 97.63 $\pm$ 0.16 \\
 & AUPR & 97.99 & 74.79 & 93.61 & 98.34 & 81.39 & 97.55 & 98.98 & 98.50 & 98.51 & 99.24 & 99.06 $\pm$ 0.07 \\
 & FPR95 & 23.81 & 100.0 & 45.41 & 18.69 & 94.60 & 34.78 & 12.06 & 18.17 & 17.07 & 9.43 & 9.46 $\pm$ 0.3 \\
 & ID ACC & 95.18 & 95.15 & 95.19 & 95.20 & 89.05 & 91.68 & 95.21 & 95.10 & 95.20 & 95.24 & 94.84 $\pm$ 0.03 \\
\specialrule{.1em}{.05em}{.05em} 
    \end{tabular}} \label{Appendix:coauthor_full}
\end{table}

\begin{table}[H]
    \centering
    \caption{Model performance on OOD datasets of paper published in 2018, 2019, and 2020 on {\fontfamily{qcr}\selectfont Arxiv}.}
    \resizebox{1\linewidth}{!}{
    \begin{tabular}{c|c|ccccccc|ccc|c}
    \specialrule{.1em}{.05em}{.05em} 
    \multirow{2}{*}{\textbf{Dataset}} & \multirow{2}{*}{\textbf{Metrics}} & \multicolumn{7}{c}{\textbf{Non-OOD Exposure}} & \multicolumn{3}{c}{\textbf{Real OOD Exposure}} & \multicolumn{1}{c}{\textbf{Ours}}\\
    & & \textbf{MSP} & \textbf{ODIN} & \textbf{Mahalanobis} & \textbf{Energy} & \textbf{GKDE} & \textbf{GPN} & \textbf{\textsc{GNNSafe}} & \textbf{OE} & \textbf{Energy FT} & \textbf{\textsc{GNNSafe++}} & \textbf{GOLD}\\
    \midrule
        \multirow{4}{*}{{\fontfamily{qcr}\selectfont Arxiv-2018}} & AUROC & 61.66 & 53.49 & 57.08 & 61.75 & 56.29 & OOM & 66.47 & 67.72 & 69.58 & 70.40 & 69.74 $\pm$ 0.28 \\
        & AUPR & 70.63 & 63.06 & 65.09 & 70.41 & 66.78 & OOM & 74.99 & 75.74 & 76.31 & 78.62 & 77.12 $\pm$ 0.23 \\
        & FPR95 & 91.67 & 100.0 & 93.69 & 91.74 & 94.31 & OOM & 89.44 & 86.67 & 82.10 & 81.47 & 83.20 $\pm$ 0.57 \\
        & ID ACC & 53.78 & 51.39 & 51.59 & 53.36 & 50.76 & OOM & 53.39 & 52.39 & 53.26 & 53.50 & 50.59 $\pm$ 0.53 \\
        \midrule
        \multirow{4}{*}{{\fontfamily{qcr}\selectfont Arxiv-2019}} & AUROC & 63.07 & 53.95 & 56.76 & 63.16 & 57.87 & OOM & 68.36 & 69.33 & 70.58 & 72.16 & 72.46 $\pm$ 0.35 \\
        & AUPR & 66.00 & 56.07 & 57.85 & 65.78 & 62.34 & OOM & 71.57 & 72.15 & 72.03 & 75.43 & 75.41 $\pm$ 0.38 \\
        & FPR95 & 90.82 & 100.0 & 94.01 & 90.96 & 93.97 & OOM & 88.02 & 85.52 & 81.30 & 79.33 & 81.16 $\pm$ 0.58 \\
        & ID ACC & 53.78 & 51.39 & 51.59 & 53.36 & 50.76 & OOM & 53.39 & 52.39 & 53.26 & 53.50 & 50.59 $\pm$ 0.53 \\
        \midrule
        \multirow{4}{*}{{\fontfamily{qcr}\selectfont Arxiv-2020}} & AUROC & 67.00 & 55.78 & 56.92 & 67.70 & 60.79 & OOM & 78.35 & 72.35 & 74.53 & 81.75 & 79.50 $\pm$ 0.11 \\
        & AUPR & 90.92 & 87.41 & 85.95 & 91.15 & 88.74 & OOM & 94.76 & 92.57 & 93.08 & 95.57 & 95.02 $\pm$ 0.04 \\
        & FPR95 & 89.28 & 100.0 & 95.01 & 89.69 & 93.31 & OOM & 83.57 & 83.28 & 78.36 & 71.50 & 77.36 $\pm$ 0.75 \\
        & ID ACC & 53.78 & 51.39 & 51.59 & 53.36 & 50.76 & OOM & 53.39 & 52.39 & 53.26 & 53.50 & 50.59 $\pm$ 0.53 \\
    \specialrule{.1em}{.05em}{.05em} 
    \end{tabular}}\label{Appendix:arxiv_full}
\end{table}

\begin{table}[H]
    \centering
    \caption{Extended ablation performance of individual subsets.}
    \resizebox{0.8\linewidth}{!}{
    \begin{tabular}{c|c|cc|cc|c}
\toprule
    \textbf{Dataset} & \textbf{Metrics} & \textbf{\textsc{GNNSafe}} & \textbf{\textsc{GNNSafe++}} & \textbf{w/o Adv.} & \textbf{w/o Det.} & \textbf{GOLD}\\
    \midrule
    \multirow{4}{*}{{\fontfamily{qcr}\selectfont Twitch-ES}} 
        & AUROC & 49.07 & 94.54 & 69.10 & 57.65 & 99.72 \\
        & AUPR  & 57.62 & 97.17 & 75.86 & 65.82 & 99.82\\
        & FPR95 & 93.98 & 44.06 & 85.82 & 91.65 & 0.44 \\
        & ID ACC & 70.40 & 70.18 & 70.97 & 70.97  & 68.49\\
    \midrule
    \multirow{4}{*}{{\fontfamily{qcr}\selectfont Twitch-FR}} 
        & AUROC & 63.49 & 93.45 & 93.86 & 88.98 & 99.08 \\
        & AUPR  & 66.25 & 95.44 & 95.45 & 92.61 & 99.25\\
        & FPR95 & 90.80 & 51.06 & 39.44  & 70.84 & 3.77\\
        & ID ACC & 70.40 & 70.18 & 70.97 & 70.97  & 68.49\\
    \midrule
    \multirow{4}{*}{{\fontfamily{qcr}\selectfont Twitch-RU}} 
        & AUROC & 87.90 & 98.10 & 90.81 & 86.48 & 99.58 \\
        & AUPR  & 89.05 & 98.74 & 94.75 & 93.30 & 99.78\\
        & FPR95 & 43.95 & 5.59 & 53.87 & 77.01 & 1.14\\
        & ID ACC & 70.40 & 70.18 & 70.97 & 70.97  & 68.49\\
    \midrule    
    \multirow{4}{*}{{\fontfamily{qcr}\selectfont Cora-S}} & AUROC & 87.52 & 90.62 & 90.05  & 93.33   & 95.48 \\
                            & AUPR   & 77.46 & 81.88 & 83.04 & 87.13 & 91.06 \\
                            & FPR95   & 73.15 & 53.51 & 59.45 & 31.98  & 21.86 \\
                            & ID ACC & 75.80 & 76.10 & 67.70  & 75.60  & 77.40\\
\midrule
    \multirow{4}{*}{{\fontfamily{qcr}\selectfont Cora-F}} & AUROC & 93.44 & 95.56 & 94.43 & {95.24}  & 96.64\\
                            & AUPR  & 88.19 & 90.27 & 91.74& 91.23 & 93.82\\
                            & FPR95   & 38.92 & 27.73 & 29.54  & 26.74 & 14.35\\
                            & ID ACC & 76.40 & 76.80 & 76.50 & 75.70  & 76.77 \\
    \midrule
    \multirow{4}{*}{{\fontfamily{qcr}\selectfont Cora-L}} & AUROC & 89.47 & 92.75 & 84.45  & 91.71& 95.40\\
                            & AUPR  & 82.21 & 82.64 & 65.90  & 81.98& 88.65 \\
                            & FPR95  & 30.83 & 34.08 &50.00 & 43.31& 17.28 \\
                            & ID ACC & 88.92 & 91.46 & 88.60  &90.80  & 90.82\\
\midrule
\multirow{4}{*}{{\fontfamily{qcr}\selectfont Arxiv-2018}}
        & AUROC & 66.47 & 70.40 & 64.97  & 65.55  & 69.74\\
        & AUPR  & 74.99 & 78.62 & 72.71 & 73.63  & 77.12\\
        & FPR95  & 89.44 & 81.47 & 90.12 & 91.19   & 83.20 \\
        & ID ACC & 53.39 & 53.50 & 49.89 & 49.66  & 50.59 \\
    \midrule
\multirow{4}{*}{{\fontfamily{qcr}\selectfont Arxiv-2019}}    
        & AUROC & 68.36 & 72.16 & 67.21  & 67.13  & 72.46\\
        & AUPR  & 71.57 & 75.43 & 69.70 & 69.06  & 75.41\\
        & FPR95  & 88.02 & 79.33 & 88.97 & 90.27  & 81.16\\
        & ID ACC & 53.39 & 53.50 & 49.89 & 49.66  & 50.59 \\
        \midrule
\multirow{4}{*}{{\fontfamily{qcr}\selectfont Arxiv-2020}}    
        & AUROC & 78.35 & 81.75 & 77.11  & 77.04  & 79.50\\
        & AUPR  & 94.76 & 95.57 & 94.39 & 94.45  & 95.02\\
        & FPR95  & 83.57 & 71.50 & 85.40 & 87.54  & 77.36\\
        & ID ACC & 53.39 & 53.50 & 49.89 & 49.66  & 50.59 \\
    \specialrule{.1em}{.05em}{.05em} 
    \end{tabular}
    } 
    \label{Appendix:Full_ablation_general}
\end{table}

\begin{table}[H]
    \centering
    \caption{Extended adversarial training effectiveness analysis of individual subsets.}
    \resizebox{0.8\linewidth}{!}{
    \begin{tabular}{c|c|c|ccc|c}
    \toprule
    \textbf{Dataset} & \textbf{Metrics} & \textbf{\textsc{GNNSafe++}}& \textbf{Dif.\ Once}  & \textbf{Dif.\ Multi} & \textbf{Real OOD} & \textbf{GOLD}\\
    \midrule
    \multirow{4}{*}{{\fontfamily{qcr}\selectfont Twitch-ES}} 
        & AUROC & 94.54& 69.10  & 66.52 & 98.99 & 99.72 \\
        & AUPR   & 97.17& 75.86 & 73.49 & 99.52 & 99.82\\
        & FPR95 & 44.06 & 85.82 & 87.07 & 1.38 & 0.44 \\
        & ID ACC & 70.18 & 70.40  & 71.12 & 70.45  & 68.49\\
    \midrule
    \multirow{4}{*}{{\fontfamily{qcr}\selectfont Twitch-FR}} 
        & AUROC & 93.45& 93.86  & 95.20 & 94.51 & 99.08 \\
        & AUPR   & 95.44& 95.45 & 96.54 & 96.33 & 99.25\\
        & FPR95 & 51.06 & 39.44 & 31.08  & 40.62 & 3.77\\
        & ID ACC & 70.18 & 70.40  & 71.12 & 70.45  & 68.49\\
    \midrule
    \multirow{4}{*}{{\fontfamily{qcr}\selectfont Twitch-RU}} 
        & AUROC & 98.10& 90.81  & 91.28 & 99.24 & 99.58 \\
        & AUPR  & 98.74& 94.75  & 95.10 & 99.65 & 99.78\\
        & FPR95 & 5.59 & 53.87 & 82.84 & 1.16 & 1.14\\
        & ID ACC & 70.18 & 70.40  & 71.12 & 70.45  & 68.49\\
    \midrule    
    \multirow{4}{*}{{\fontfamily{qcr}\selectfont Cora-S}} 
        & AUROC & 90.62 & 90.05  & 95.69  & 94.12 & 95.48 \\
        & AUPR  & 81.88 & 83.04  & 91.59 & 89.91 & 91.06 \\
        & FPR95   & 53.51 & 59.45 & 22.30 &  37.11 & 21.86 \\
        & ID ACC & 76.10& 67.70  & 76.10 & 75.90 & 77.40\\
    \midrule
    \multirow{4}{*}{{\fontfamily{qcr}\selectfont Cora-F}} 
        & AUROC & 95.56& 94.43  & 96.02 & 97.60 & 96.64\\
        & AUPR  & 90.27& 91.74  & 92.99  & 94.23 & 93.82\\
        & FPR95  & 27.73 & 29.54 & 18.94 &10.27 & 14.35\\
        & ID ACC & 76.80 & 76.50 & 77.30  & 71.70 & 76.77 \\
    \midrule
    \multirow{4}{*}{{\fontfamily{qcr}\selectfont Cora-L}} 
        & AUROC & 92.75& 84.45  & 86.76 & 95.04 & 95.40 \\
        & AUPR   & 82.64 & 65.90 & 70.70 & 86.01& 88.65 \\
        & FPR95  & 34.08 & 50.00 & 49.09 & 17.24 & 17.28 \\
        & ID ACC & 91.46  & 88.60 & 88.29 & 87.65 & 90.82\\
    \midrule
\multirow{4}{*}{{\fontfamily{qcr}\selectfont Arxiv-2018}}
        & AUROC & 70.40& 64.97  & 67.26  & 75.32  & 69.74\\
        & AUPR  & 78.62& 72.71  & 74.72 & 80.89  & 77.12\\
        & FPR95  & 81.47 & 90.12 & 85.23 & 72.40   & 83.20 \\
        & ID ACC & 53.50 & 53.39  & 50.77 & 49.99  & 50.59 \\
    \midrule
\multirow{4}{*}{{\fontfamily{qcr}\selectfont Arxiv-2019}}    
        & AUROC & 72.16 & 67.21 & 69.66  & 77.98 & 72.46\\
        & AUPR  & 75.43 & 69.70 & 72.05 & 79.56  & 75.41\\
        & FPR95  & 79.33& 88.97  & 83.33 & 95.92  & 81.16\\
        & ID ACC & 53.50& 53.39  & 50.77 & 49.99   & 50.59 \\
        \midrule
\multirow{4}{*}{{\fontfamily{qcr}\selectfont Arxiv-2020}}    
        & AUROC & 81.75& 77.11   & 79.52  &83.41 & 79.50\\
        & AUPR  & 95.57& 94.39  & 94.95 & 95.92  & 95.02\\
        & FPR95  & 71.50& 85.40  & 77.52 & 64.93  & 77.36\\
        & ID ACC & 53.50& 53.39  & 50.77 &49.99   & 50.59 \\
    \specialrule{.1em}{.05em}{.05em} 
    \end{tabular}} \label{Appendix:Full_ablation_adversarial}
\end{table}

\begin{table}[H]
    \centering
    \caption{Extended energy regulariser effectiveness analysis on {\fontfamily{qcr}\selectfont Twitch}.}
    \resizebox{1\linewidth}{!}{
    \begin{tabular}{ccc|cccc|cccc|cccc}
    \toprule
    \multirow{2}{*}{$\mathcal{L}_\text{Unc}$} & \multirow{2}{*}{$\mathcal{L}_\text{EReg}$} & \multirow{2}{*}{$\mathcal{L}_\text{DReg}$} & \multicolumn{4}{c|}{{\fontfamily{qcr}\selectfont Twitch-ES}}& \multicolumn{4}{c|}{{\fontfamily{qcr}\selectfont Twitch-FR}}& \multicolumn{4}{c}{{\fontfamily{qcr}\selectfont Twitch-RU}}\\
    & & & AUROC & AUPR & FPR & ID Acc & AUROC & AUPR & FPR & ID Acc& AUROC & AUPR & FPR & ID Acc\\
    \midrule
     & & &
    74.33 & 76.23 & 52.97& 68.97& 
   98.34 & 98.72 & 2.58& 68.97 &
   26.92& 48.67& 91.70 & 68.97\\
    \midrule
     \checkmark& & & 
    17.39 & 44.31 & 98.04 & 70.15& 
   5.69 & 34.24& 99.07& 70.15 &
   7.48 & 43.30 & 96.42 &70.15 \\
    & \checkmark& & 
    60.03 & 68.63& 92.36& 70.98 & 
    90.21& 92.64 & 66.57 & 70.98 &
    83.81 & 88.85 & 77.77& 70.98 \\
    & & \checkmark& 
    18.69 & 44.81 & 97.89 & 70.79& 
    96.40 & 94.65 & 8.85 & 70.79 &
    92.04 & 91.19 & 26.89 & 70.79\\
    \midrule
     \checkmark& \checkmark& &
    59.40  & 68.20 & 92.77 & 70.99 & 
    91.91& 93.23 & 56.07 & 70.99 0&
   79.32 & 83.03 & 79.59 & 70.99 \\
     \checkmark& & \checkmark& 
    7.50 & 42.10 & 99.07 & 70.90 & 
    99.04 & 98.15 & 1.30 & 70.90 &
    86.76 & 86.13 & 37.49 & 70.90 \\
    & \checkmark& \checkmark& 
    90.55 & 94.20 & 41.98 & 69.64& 
    88.95 & 91.30 & 45.86 & 69.64 &
    90.35 & 80.94 & 42.09 & 69.64 \\
    \midrule
    & GOLD & & 
    99.72 &99.82 & 0.44& 68.49& 
    99.08 & 99.25 & 3.77 & 68.49& 
    99.58 & 99.78 & 1.14& 68.49\\
    \specialrule{.1em}{.05em}{.05em} 
    \end{tabular}} \label{Appendix:Twtich_reg_effectiveness}
\end{table}

\begin{table}[H]
    \centering
    \caption{Extended energy regulariser effectiveness analysis on {\fontfamily{qcr}\selectfont Cora}.}
    \resizebox{1\linewidth}{!}{
    \begin{tabular}{ccc|cccc|cccc|cccc}
    \toprule
    \multirow{2}{*}{$\mathcal{L}_\text{Unc}$} & \multirow{2}{*}{$\mathcal{L}_\text{EReg}$} & \multirow{2}{*}{$\mathcal{L}_\text{DReg}$} & \multicolumn{4}{c|}{{\fontfamily{qcr}\selectfont Cora-S}}& \multicolumn{4}{c|}{{\fontfamily{qcr}\selectfont Cora-F}}& \multicolumn{4}{c}{{\fontfamily{qcr}\selectfont Cora-L}}\\
    & & & AUROC & AUPR & FPR & ID Acc & AUROC & AUPR & FPR & ID Acc& AUROC & AUPR & FPR & ID Acc\\
    \midrule
     & & &
    86.44 & 91.26 & 35.20 & 67.70 & 
    13.00 & 16.79 & 99.19 & 70.80 &
    83.98 & 76.01 & 90.06 & 90.19\\
    \midrule
     \checkmark& & & 
    68.06 & 67.14 & 89.18 & 80.00& 
    68.01 & 77.40 & 45.70 & 75.80 &
    76.20 & 60.36 & 96.65 & 87.34 \\
    & \checkmark& & 
    94.70 & 90.09 & 30.54 & 74.30 & 
    10.41 & 15.72 & 100 & 76.20 &
    92.90 & 84.14 & 31.64 & 90.82 \\
    & & \checkmark& 
    89.18 & 80.96 & 65.18 & 72.80& 
    79.83 & 74.36 & 90.69 & 70.00&
    84.16 & 68.39 & 49.59 & 85.44\\
    \midrule
     \checkmark& \checkmark& &
    47.95  & 49.68 & 91.32 & 77.60 & 
    47.63 & 54.31 & 99.41 & 76.60&
    8.30 & 13.82 & 99.80 & 89.55\\
     \checkmark& & \checkmark& 
    86.26 & 76.48 & 64.59 & 69.70 & 
    79.40 & 73.66 & 93.65 & 67.50 &
    86.43 & 71.52 & 62.37 & 86.39 \\
    & \checkmark& \checkmark& 
    93.94 & 88.86 & 28.99 & 75.60 & 
    95.54 & 92.69 & 22.05 & 76.70 &
    90.35 & 80.94 & 42.09 & 87.34 \\
    \midrule
    & GOLD & & 
    95.48 & 91.06 & 21.86 & 77.40& 
    96.64 & 93.82 & 14.35 & 76.77& 
    95.40 & 88.65 & 17.28 & 90.82\\
    \specialrule{.1em}{.05em}{.05em} 
    \end{tabular}} \label{Appendix:Cora_reg_effectiveness}
\end{table}

\begin{table}[H]
    \centering
    \caption{Extended energy regulariser effectiveness analysis on {\fontfamily{qcr}\selectfont Amazon}.}
    \resizebox{1\linewidth}{!}{
    \begin{tabular}{ccc|cccc|cccc|cccc}
    \toprule
    \multirow{2}{*}{$\mathcal{L}_\text{Unc}$} & \multirow{2}{*}{$\mathcal{L}_\text{EReg}$} & \multirow{2}{*}{$\mathcal{L}_\text{DReg}$} & \multicolumn{4}{c|}{{\fontfamily{qcr}\selectfont Amazon-S}}& \multicolumn{4}{c|}{{\fontfamily{qcr}\selectfont Amazon-F}}& \multicolumn{4}{c}{{\fontfamily{qcr}\selectfont Amazon-L}}\\
    & & & AUROC & AUPR & FPR & ID Acc & AUROC & AUPR & FPR & ID Acc& AUROC & AUPR & FPR & ID Acc\\
    \midrule
     & & &
    96.67 & 97.48 & 20.16 & 88.35 & 
    1.21 & 26.70 & 99.35 & 92.11 &
    94.64 & 93.84 & 20.66 & 95.76 \\
    \midrule
     \checkmark& & & 
    84.90 & 90.80 & 100.00 & 92.17 & 
    21.53 & 30.92 & 86.18 & 92.96 &
    95.17 & 93.45 & 18.21 & 95.96 \\
    & \checkmark& & 
    52.82 & 69.21 & 100.00 & 92.74 & 
    1.58 & 26.81 & 98.47 & 91.89 &
    91.53 & 89.77 & 35.26 & 95.72 \\
    & & \checkmark& 
    100.00 & 100.00 & 0.00 & 91.35 & 
    99.76 & 99.55 & 0.63 & 91.62 &
    93.39 & 90.94 & 24.20 & 95.39 \\
    \midrule
     \checkmark& \checkmark& &
    85.72 & 91.06 & 99.97 & 92.48 & 
    57.99 & 44.03 & 55.73 & 92.63 &
    69.74 & 56.45 & 68.85 & 94.79 \\
     \checkmark& & \checkmark& 
    100.00 & 100.00 & 0.00 & 92.33 & 
    98.61 & 99.09 & 0.29 & 91.76 &
    95.11 & 91.99 & 13.34 & 95.52 \\
    & \checkmark& \checkmark& 
    98.58 & 99.21 & 0.00 & 92.25 & 
    98.36 & 98.75 & 0.60 & 92.04 &
    97.13 & 97.29 & 9.61 & 94.14 \\
    \midrule
    & GOLD & & 
    99.98 & 99.99 & 0.00 & 92.03& 
    99.17 & 99.31 & 0.14 & 91.76& 
    97.26 & 97.46 & 6.06 & 95.18\\
    \specialrule{.1em}{.05em}{.05em} 
    \end{tabular}} \label{Appendix:Amazon_reg_effectiveness}
\end{table}

\subsection{Ablation study visualisation} \label{Appendix:ablation_vis}
\begin{figure}[h]
    \centering
    \begin{subfigure}[b]{0.25\textwidth}
        \centering
        \includegraphics[width=\textwidth]{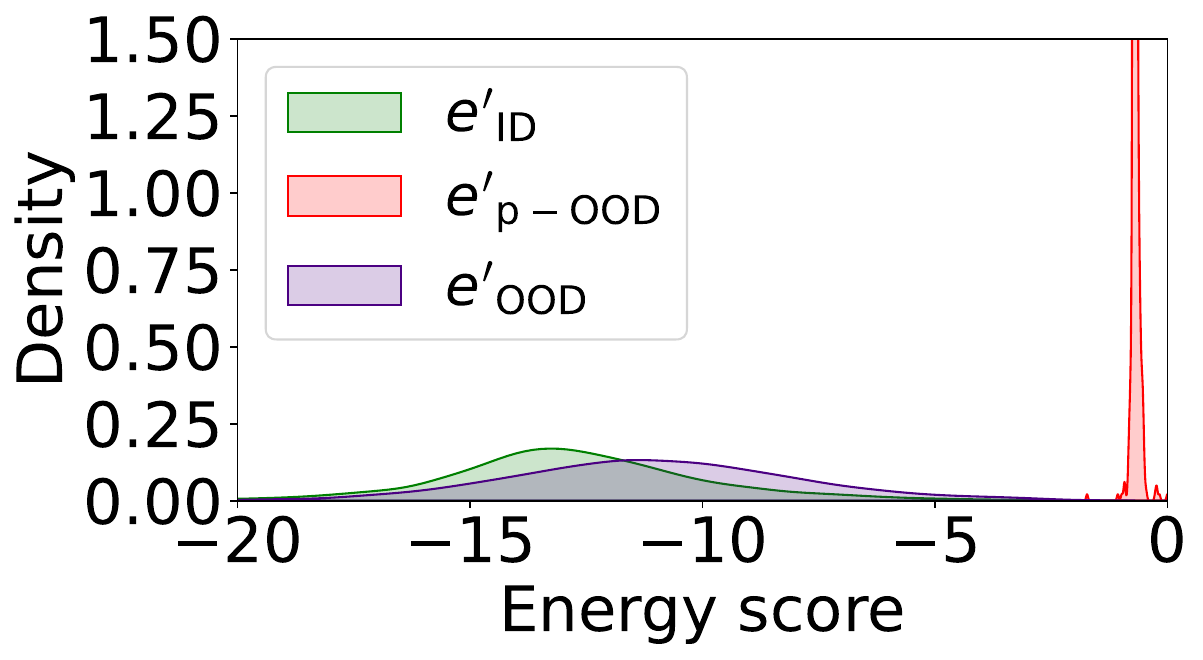}
        \caption{Twitch w/o Adv.}
    \end{subfigure}
    \hfill
    \begin{subfigure}[b]{0.25\textwidth}
        \centering
        \includegraphics[width=\textwidth]{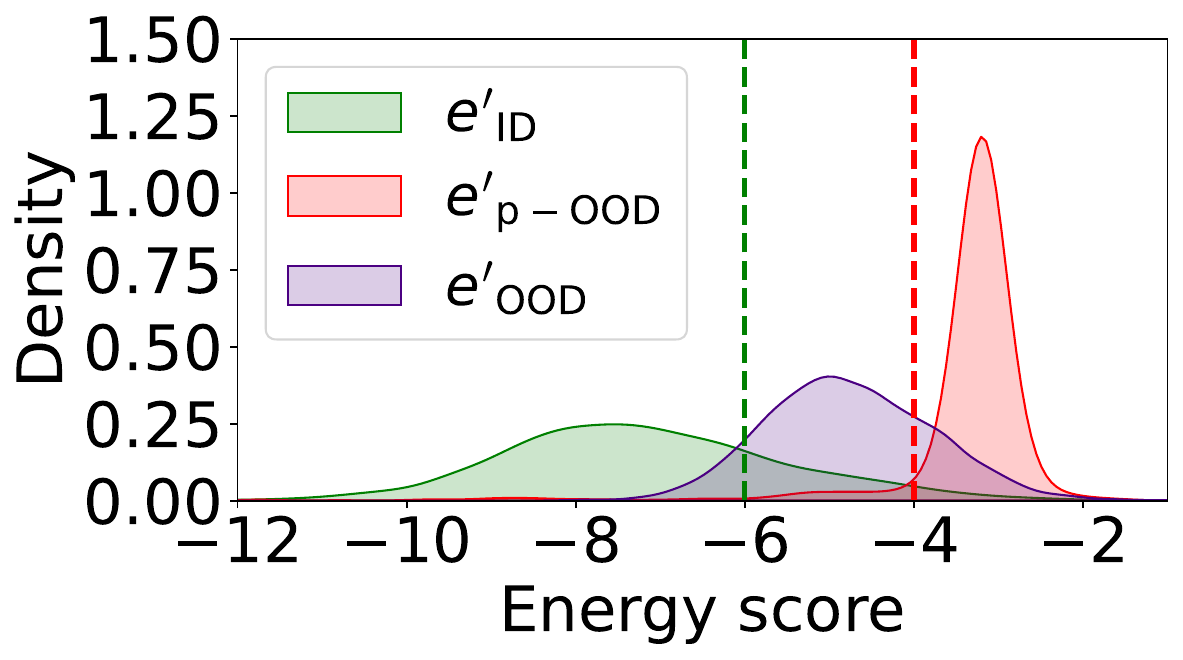}
        \caption{Twitch w/o Det.}
    \end{subfigure}
    \hfill
    \begin{subfigure}[b]{0.25\textwidth}
        \centering
        \includegraphics[width=\textwidth]{fig/Twitch-IDvsOOD_Enlarged_v2.pdf}
        \caption{Twtich GOLD}
    \end{subfigure}
    \hfill
    \begin{subfigure}[b]{0.25\textwidth}
        \centering
        \includegraphics[width=\textwidth]{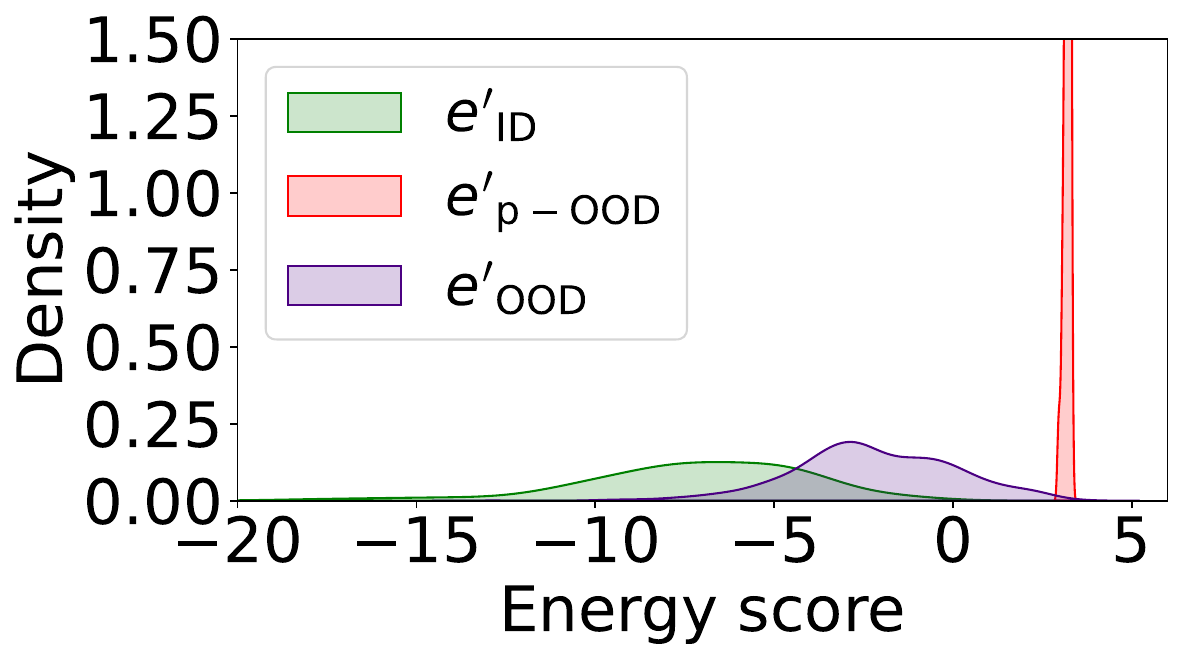}
        \caption{Cora-L w/o Adv.}
    \end{subfigure}
    \hfill
    \begin{subfigure}[b]{0.25\textwidth}
        \centering
        \includegraphics[width=\textwidth]{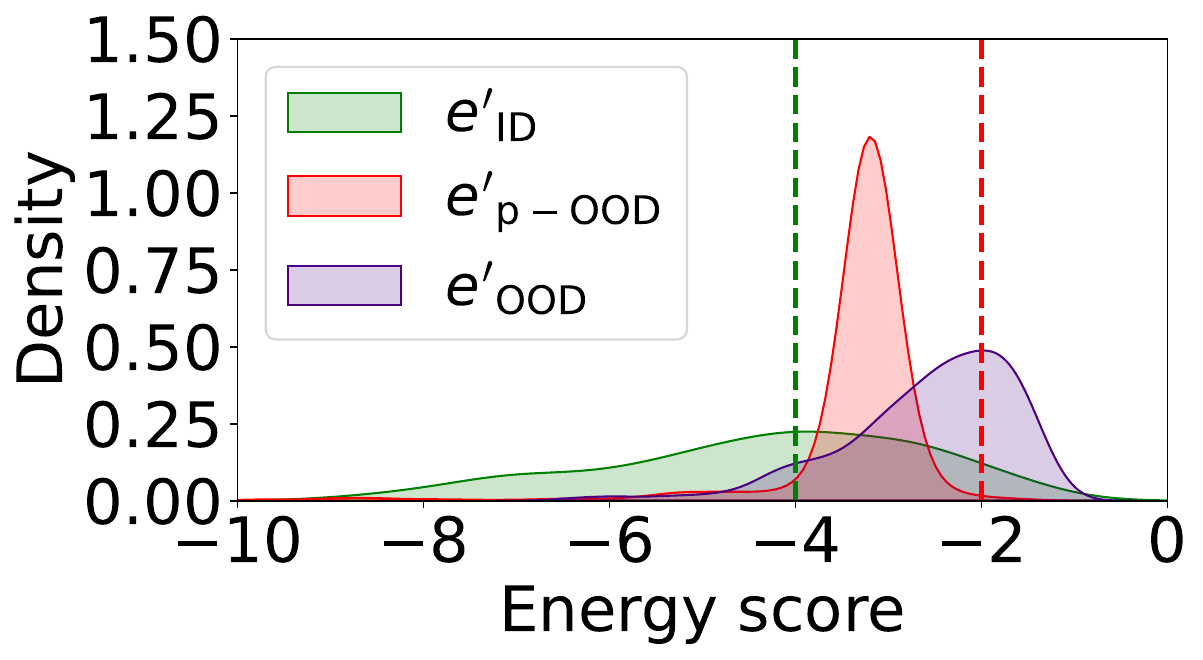}
        \caption{Cora-L w/o Det.}
    \end{subfigure}
    \hfill
    \begin{subfigure}[b]{0.25\textwidth}
        \centering
        \includegraphics[width=\textwidth]{fig/CoraL-IDvsOOD_Enlarged_v2.pdf}
        \caption{Cora-L GOLD}
    \end{subfigure}
    \caption{
    \textbf{Visualisation of the energy score distributions of GOLD without adversarial training or the use of detector for {\fontfamily{qcr}\selectfont Twitch} and {\fontfamily{qcr}\selectfont Cora-L} datasets.} (a) and (d) illustrate the energy score gaps w/o adversarial training, where the energy of p-OOD data will be diverged too far and fail to diverge the energy of real OOD. (b) and (e) shows the energy scores derived from the GNN classifier without our proposed detector, where the energy scores cannot be effectively separated. (c) and (f) demonstrates the ability of GOLD to effectively distinguish the ID and OOD energy distributions, illustrating the effectiveness of the adversarial and detector components.}
    \vspace{-0.5cm}
    \label{Appendix:Additional_Ablation_fig}
\end{figure}

To explore how the different modules contribute to OOD detection, we present further energy distribution visualisations in Figure \ref{Appendix:Additional_Ablation_fig}.
The adversarial training can help to maintain the closeness between synthetic and ID data, preventing the synthetic samples from diverging too far from real ID/OOD data to bias the detector. Our method alternates between two tasks: (1) the latent diffusion model pulls latent embeddings of ID data and generated pseudo-OOD embeddings closer, while (2) the GNN \& detector push their energies apart. Without this component, the pseudo-OOD distribution diverges significantly compared to GOLD in Figure \ref{Appendix:Additional_Ablation_fig} c/f, where synthetic embeddings are regularly updated. This divergence makes OOD data indistinct from ID data, leading to poor performance in the ablation study.
The detector can decrease the overlap of energy distribution between the ID and OOD samples, leading to better energy-based OOD detection. Figure \ref{Appendix:Additional_Ablation_fig} b/e shows that w/o detector will lead to a large overlap of energy distribution between ID and OOD samples. This overlap occurs because the energy scores, derived from prediction logits of the GNN classifier, become indistinct as the number of predicted classes increases and when the classifier struggles to distinguish certain classes. Thus, introducing a dedicated detector to further discern energy scores enhances detection by reducing the number of output classes.

\subsection{Latent Generative Model} \label{Appendix:latent_generative_model}
In this section, we provide description of the two latent generative models utilised: The variational autoencoder and the latent diffusion model.

\subsubsection{Variational Autoencoder}
The variational autoencoder (VAE) is a generative model consisting of an encoder that learns latent variables from training data and a decoder that then uses those latent variables to reconstruct the input data. VAEs are trained to optimise a lower bound on the marginal log-likelihood $\log p_\theta(x)$ over the data by using a learned approximate posterior $q_\phi(h|x)$, as follows:

$$\mathcal{L}(\theta, \phi; x) = \mathbb{E}_{q_\phi(h|x)}[\log p_\theta(x|h)] - D_{KL}(q_\phi(h|x) || p(h))$$
s.t the first term is the reconstruction loss, and the second term is the KL divergence of the approximate from the true posterior. The trained approximate posterior $q_\phi(h|x)$ would thus act as an encoder that maps the data $x$ to a lower dimensional latent representation, and latent samples $h$ can be drawn via the reparametrisation trick: 
$$ h = \mu_\phi(x) + \sigma_\phi(x) \odot \epsilon, \text{ where } \epsilon \sim \mathcal{N}(0, I) \text{ if the models are Gaussian}. $$

We set the encoder hidden dimension size to be 512, the decoder dimension to be 256, and layer sizes to be 2.

\subsubsection{Latent Diffusion Model}
The latent diffusion model consists of a forward diffusion and a backward denoising process on a set of latent representations~\citep{DDPM, LatentDIF, NGG}. In our GOLD, a latent node representation $\mathbf{h}_0\in\mathbb{R}^{d'}$ is initialised at timestep 0 from the GNN encodings $\mathbf{H}$ in Eq.~\ref{eq:GCN_emb_logits}. At the forward process, the model progressively adds Gaussian noise to the latent node representation  $\mathbf{h}_0$, according to a known variance schedule $\beta_1, \cdots, \beta_T$, for $0 < \beta_1 < . . . \beta_T < 1$. This process will produce a sequence of increasingly noisy vectors $(\mathbf{h}_1, \cdots \mathbf{h}_T)$ with timestep $t = \{1,2,3, \dots, T\}$. Denoting $a_t=1-\beta_t$ and $\bar{a}_t=\prod_{i=1}^t a_i$, we can derive a closed form for obtaining the representation at any timestep $t$ given the initial representation $\mathbf{h}_0$:
\begin{equation}
\mathbf{h}_t \sim \mathcal{N}\left(\sqrt{\bar{a}_t} \mathbf{h_0},\left(1-\bar{a}_t\right) \mathbf{I}\right).
\end{equation}
The backward denoising process involves predicting the noise added to the representation at a given timestep via a denoising model $D$ (e.g., MLP). To train the latent diffusion model, we minimise the mean squared error loss between the added noise $\bm{\epsilon} ~ \sim \mathcal{N}(\mathbf{0},\mathbf{I})$ and the predicted noise from the noisy representation $\mathbf{h}_t$ at a given timestep $t$ with the reparameterisation trick: 
\begin{equation}
\min_D\mathcal{L}_{\mathrm{{Gen}}},\text{ where }\mathcal{L}_{\mathrm{{Gen}}} =\mathbb{E}_{\mathbf{h}_0, \bm{\epsilon}, t}\left[\left\|\bm{\epsilon} -D\left(\sqrt{\bar{a}_t} \mathbf{h}_0 +\sqrt{\left(1-\bar{a}_t\right)} \bm{\epsilon}, t\right)\right\|_2^2\right]. \label{eq:latent_appendix}
\end{equation}

\begin{table}[h!]
\centering
\resizebox{0.6\linewidth}{!}{\begin{tabular}{c|c|c|c|c|c}
\toprule
\textbf{Hyperparams} & \textbf{} & \textbf{AUROC} & \textbf{AUPR} & \textbf{FPR} & \textbf{ID ACC} \\
\midrule
\multirow{4}{*}{$\beta_1$} & 0.00001 & 99.43 & 99.59 & 1.77 & 68.48 \\
          & \textbf{0.0001}  & 99.46 & 99.62 & 1.78 & 68.49 \\
          & 0.001   & 99.52 & 99.66 & 1.44 & 68.11 \\
          & 0.01    & 99.46 & 99.60 & 1.99 & 67.71 \\
\midrule
\multirow{3}{*}{$\beta_T$} & 0.005   & 85.91 & 88.79 & 44.77 & 71.04 \\
          & \textbf{0.02}    & 99.46 & 99.62 & 1.78 & 68.49 \\
          & 0.1     & 82.02 & 86.79 & 55.27 & 68.40 \\
\midrule
\multirow{6}{*}{$T$}       & 400     & 96.59 & 97.84 & 18.23 & 69.43 \\
          & 500     & 95.15 & 96.38 & 21.32 & 68.99 \\
          & \textbf{600}     & 99.46 & 99.62 & 1.78 & 68.49 \\
          & 700     & 98.80 & 99.27 & 4.07 & 68.09 \\
          & 800     & 92.02 & 95.68 & 66.94 & 68.85 \\
          & 1000    & 17.35 & 45.22 & 99.57 & 71.20 \\
\bottomrule
\end{tabular}}
\caption{Performance comparison for different hyperparameters for Diffusion model on Twitch dataset. Default values are highlighted in \textbf{Bold}.}
\label{Appendix:hyperparam_dif}
\end{table}
We set the diffusion model parameter $\beta$ to be a sequence of linearly increasing constants from $\beta_1=10^{-4}$ to $\beta_T=0.02$ as presented in \citep{DDPM,sd}.
A hyperparameter sensitivity experiment on the Twitch dataset is provided in Table \ref{Appendix:hyperparam_dif}. Generally, a larger (smaller) $\beta$ adds or removes more (less) noise at each step. A larger $T$ increases noise corruption, making recovery harder but with more output variation, while a smaller $T$ reduces noise corruption, making recovery easier but limiting variation. $\beta_1$ typically does not affect performance, while $\beta_T$ is more sensitive, reflecting higher/lower corruption at the end of the timestep. The value of $T$ also impacts performance, with non-default values either limiting or excessively diversifying the synthetic samples.

\subsection{Computational Cost} \label{Appendix:computational_cost}
In this section, we provide the computational cost of GOLD against SOTA baselines. GOLD outperforms the baselines with a rough trade-off of 2x training time and memory usage.
\begin{table}[!h]
    \centering
    \caption{\textbf{Computation cost (one 32GB (32768MiB) NVIDIA V100 GPU) and OOD detection performance of GOLD (Non-OOD Exposed) against both \textbf{Non- and Real-OOD exposed} SOTA baselines.} The `Train' column is the training convergence time in seconds. The `Test' column is the inference time in seconds. The `Mem.' column is the maximum memory usage in Mebibytes (MiB). The `FPR95' column is the OOD detection performance in \%, the lower the better. \textbf{The inference time of these methods is the same with the same backbone GNN.}}
    \resizebox{1\linewidth}{!}{
    \begin{tabular}{c|cccc|cccc|cccc|cccc|cccc}
        \toprule
        &\multicolumn{4}{c|}{\fontfamily{qcr}\selectfont Twitch}&\multicolumn{4}{c|}{\fontfamily{qcr}\selectfont Cora-F}&\multicolumn{4}{c|}{\fontfamily{qcr}\selectfont Amazon-F}&\multicolumn{4}{c|}{\fontfamily{qcr}\selectfont Coauthor-F}&\multicolumn{4}{c}{\fontfamily{qcr}\selectfont Arxiv}\\
        &Train&Test&Mem.&FPR95$(\downarrow)$&Train&Test&Mem.&FPR95$(\downarrow)$&Train&Test&Mem.&FPR95$(\downarrow)$&Train&Test&Mem.&FPR95$(\downarrow)$&Train&Test&Mem.&FPR95$(\downarrow)$\\
        \midrule
        \textsc{GNNSafe} (Non) 
        & 2.41 & 0.08 & 667 & 76.24 
        & 4.40 & 0.03 & 465 & 38.92 
        & 13.51 & 0.04 & 665 & 0.31
        & 57.80 & 0.35 &  1523 & 0.51
        & 85.23 & 0.40 &  3370 & 87.01\\     
        
        \midrule
        \textsc{GNNSafe++} (Real)
        & 4.74 & 0.09 & 667 &33.57
        & 5.32 & 0.03 & 465 & 27.73
        & 18.40 & 0.05 & 665 & 0.13
        & 67.83 & 0.36 & 1523 & 0.09
        & 132.36 & 0.40 & 3370 & 77.43\\
        \midrule
        GOLD w/ VAE (Non)
        & 2.78 & 0.09 & 1427 & 3.03
        & 3.91 & 0.04 & 1081 & 23.60
        & 12.52 & 0.05 & 1319 & 0.15
        & 55.65 & 0.35 & 2439 & 0.23
        & 80.77 & 0.45 & 9039 & 81.95\\
        \midrule
        GOLD w/ LDM (Non)
        & 8.96 & 0.10 & 1452 & 1.71
        & 5.93 & 0.04 & 1083 & 14.51
        & 39.04 & 0.07 & 1347 & 0.11
        & 89.74 & 0.37 & 2515 & 0.01
        & 244.95 & 0.47 & 10579 & 80.35\\
        \bottomrule
        \end{tabular}
            }
\end{table}

\subsection{Ablation with Additional Backbone}\label{Appendix: Backbone}
We provide the following experiments with two additional backbones: GAT~\citep{GAT} and MixHop~\citep{MixHop}. We compare these architectures against GNNSafe and NodeSafe and their OOD-exposed variants. To ensure a fair comparison, we maintain the same configuration as the original GCN implementation, with a hidden dimension of 64, two layers, 8 attention heads for GAT, and two hops for MixHop. The results shown in Table~\ref{tab:backbone}, demonstrate that GOLD outperforms other methods across the evaluated backbones.

\begin{table}[h!]
\centering
\caption{Ablation of different backbones}

\resizebox{1\linewidth}{!}{
\begin{tabular}{c|c|c|c|c|c|c|c}
\toprule
\textbf{Dataset} & \textbf{Backbone} & \textbf{Metrics} & \textbf{\textsc{GNNSafe}} & \textbf{\textsc{GNNSafe++}} & \textbf{\textsc{NodeSafe}} & \textbf{\textsc{NodeSafe++}} & \textbf{GOLD} \\ \midrule
\multirow{6}{*}{{\fontfamily{qcr}\selectfont Twitch}}           
    & \multirow{3}{*}{MixHop} & AUROC & \textcolor{purple}{72.08} & 95.07 & 57.91 & \underline{95.08}  & \textcolor{teal}{\textbf{96.94}}  \\ \cline{3-8} 
        &   & FPR95 & \textcolor{purple}{73.70} & 33.46 & 93.76 & \underline{30.71}  & \textcolor{teal}{\textbf{17.98}}  \\ \cline{3-8} 
        &   & ID Acc & 69.66 & 66.04 & 70.09 & 70.56 & 67.58  \\ \cline{2-8}
    & \multirow{3}{*}{GAT}  & AUROC  & \textcolor{purple}{83.08} & \underline{97.51}  & 54.78 & 95.07                & \textcolor{teal}{\textbf{98.64}}  \\ \cline{3-8} 
    &   & FPR95 & \textcolor{purple}{50.46}  & \underline{20.43}  & 93.24 & 30.71                & \textcolor{teal}{\textbf{1.42}}   \\ \cline{3-8} 
    &  & ID Acc & 68.21 & 68.54 & 68.40 & 70.56 & 67.32 \\ \midrule
\multirow{6}{*}{{\fontfamily{qcr}\selectfont Cora}} 
    & \multirow{3}{*}{MixHop}  & AUROC  & \textcolor{purple}{88.65}  & 91.33 & 82.60 & \textbf{92.79}  & \textcolor{teal}{\underline{91.42}}  \\ \cline{3-8} 
    &         & FPR95  & \textcolor{purple}{59.08}  & 44.59 & 60.22 & \underline{38.63}  & \textcolor{teal}{\textbf{25.09}}  \\ \cline{3-8} 
    &        & ID Acc  & 79.52  & 80.66 & 82.16 & 81.45 & 80.67  \\ \cline{2-8}
    & \multirow{3}{*}{GAT} & AUROC & \textcolor{purple}{91.62}  & \underline{92.50} & 85.55 & 92.32                & \textcolor{teal}{\textbf{94.66}}  \\ \cline{3-8} 
    &                      & FPR95 & \textcolor{purple}{33.81}  & \underline{33.44} & 55.20 & 34.93                & \textcolor{teal}{\textbf{19.63}}  \\ \cline{3-8} 
    &                      & ID Acc & 79.44 & 79.52 & 81.06 & 80.23 & 78.40 \\ 
    \bottomrule
\end{tabular}
}
\label{tab:backbone}
\end{table}

\end{document}